\UseRawInputEncoding
\pdfoutput=1

\documentclass{article}

\usepackage[utf8]{inputenc} % allow utf-8 input
\usepackage[T1]{fontenc}    % use 8-bit T1 fonts
\usepackage{hyperref}       % hyperlinks
\usepackage{url}            % simple URL typesetting
\usepackage{booktabs}       % professional-quality tables
\usepackage{nicefrac}       % compact symbols for 1/2, etc.
\usepackage{microtype}      % microtypography
\usepackage{lipsum}		% Can be removed after putting your text content
\usepackage{graphicx}
\usepackage{doi}
\usepackage{xcolor}         % colors
\usepackage{amsmath,amsfonts,amssymb,amsthm,bm}  
\usepackage{algorithm}
\usepackage{algorithmic}
\usepackage{wrapfig}
\usepackage{caption}
\usepackage{subcaption}
\usepackage{enumitem,kantlipsum}
\usepackage{xcolor,colortbl}
\usepackage{makecell}
\usepackage{bbding}
\usepackage{pifont}
\usepackage{wasysym}
\usepackage{wrapfig}
\usepackage{amssymb}
\usepackage{booktabs}% http://ctan.org/pkg/booktabs
\newcommand{\tabitem}{~~\llap{\textbullet}~~}
\usepackage{tikz}
\newcommand\diag[4]{%
  \multicolumn{1}{p{#2}|}{\hskip-\tabcolsep
  $\vcenter{\begin{tikzpicture}[baseline=0,anchor=south west,inner sep=#1]
  \path[use as bounding box] (0,0) rectangle (#2+2\tabcolsep,\baselineskip);
  \node[minimum width={#2+2\tabcolsep},minimum height=\baselineskip+\extrarowheight] (box) {};
  \draw (box.north west) -- (box.south east);
  \node[anchor=south west] at (box.south west) {#3};
  \node[anchor=north east] at (box.north east) {#4};
 \end{tikzpicture}}$\hskip-\tabcolsep}}

\newcommand{\xmark}{\ding{55}}%

\definecolor{Gray}{gray}{0.85}
\definecolor{LightCyan}{rgb}{0.88,1,1}

\newtheorem{assumption}{Assumption}

\newtheorem{theorem}{Theorem}
\newtheorem{lemma}{Lemma}
\newtheorem{remark}{Remark}
\newtheorem{definition}{Definition}
\newtheorem{corollary}{Corollary}
\usepackage{tikz}

\newcommand{\hide}[1]{} %To hide large blocks of code without using % symbols

\newcommand\Ebb{\ensuremath{\mathbb{E}}}

\newcommand\Rbb{\ensuremath{{\mathbb{R}}}}

\newcommand\Zbb{\ensuremath{{\mathbb{Z}}}}
\newcommand\Oc{\ensuremath{\mathcal{O}}}

\newcommand\Sc{\ensuremath{\mathcal{S}}}

\newcommand\Nc{\ensuremath{{\mathcal{N}}}}

\usepackage{tikz}

\newcommand{\red}[1]{{\color{red}{#1}}}

\newcommand{\specialcell}[2][c]{%
  \begin{tabular}[#1]{@{}c@{}}#2\end{tabular}}
\usepackage{hyperref}
\usepackage{url}

\usepackage{arxiv}

\title{$z$-SignFedAvg: A Unified  Stochastic Sign-based Compression for Federated Learning}

 \author{ Zhiwei Tang \\
	\texttt{zhiweitang1@link.cuhk.edu.cn} \\
	\And
	Yanmeng Wang\\
	\texttt{yanmengwang@link.cuhk.edu.cn} \\
    \And 
    Tsung-Hui Chang\\
	\texttt{changtsunghui@cuhk.edu.cn} \\
    \AND
    The Chinese University of Hong Kong, Shenzhen
}

\renewcommand{\headeright}{}
\renewcommand{\undertitle}{}
\renewcommand{\shorttitle}{}

\hypersetup{
pdftitle={A template for the arxiv style},
pdfsubject={q-bio.NC, q-bio.QM},
pdfauthor={David S.~Hippocampus, Elias D.~Striatum},
pdfkeywords={First keyword, Second keyword, More},
}

\begin{document}
\maketitle

\begin{abstract}
    Federated Learning (FL) is a promising privacy-preserving distributed learning paradigm but suffers from high communication cost when training large-scale machine learning models.  Sign-based methods,  such as SignSGD \cite{bernstein2018signsgd},  have been proposed as a biased gradient compression technique for reducing the communication cost. However,  sign-based algorithms could diverge under heterogeneous data, which thus motivated the development of advanced techniques, such as the error-feedback method and stochastic sign-based compression, to fix this issue.
    Nevertheless, these methods still suffer from slower convergence rates. Besides, none of them allows multiple local SGD updates like FedAvg \cite{mcmahan2017communication}.  In this paper,  we propose a novel noisy perturbation scheme with a general symmetric noise distribution for sign-based compression, which not only allows one to flexibly control the tradeoff between gradient bias and convergence performance, but also provides a unified viewpoint to existing stochastic sign-based methods.  More importantly,   the unified noisy perturbation scheme enables the development of the very first sign-based FedAvg algorithm ($z$-SignFedAvg) to accelerate the convergence. Theoretically,  we show that $z$-SignFedAvg achieves a faster convergence rate than existing sign-based methods and,  under the uniformly distributed noise, can enjoy the same convergence rate as its uncompressed counterpart. 
    Extensive experiments are conducted to demonstrate that the $z$-SignFedAvg can achieve competitive empirical performance on real datasets  and 
    outperforms existing schemes. 
\end{abstract}

\section{Introduction}
\label{sec:intro}
We consider the Federated Learning (FL) network 
with one parameter server and $n$ clients \cite{mcmahan2017communication,li2020federated}, with the focus on solving 
the following distributed learning problem

\begin{align}
    \label{p:problem}
	\min_{x\in\Rbb^d} f(x) = \frac{1}{n}\sum_{i=1}^n f_i(x),
\end{align} where $f_i(\cdot)$ is the local objective function for the $i$-th client, for $i=1,\ldots,n$. Throughout this paper, we assume that each $f_i$ is smooth and possibly non-convex. The local objective functions are generated from the local dataset owned by each client. 
%Such a setting is common in Federated Learning (FL) \cite{mcmahan2017communication,li2020federated}, and a seminal algorithm is federated averaging (FedAvg) \cite{mcmahan2017communication,konevcny2016federated}. FedAvg considers local multiple SGD updates with periodic communication, which can signficantly reduce the communicatio cost in FL. 
When designing distributed algorithms to solve \eqref{p:problem}, a crucial aspect is the communication efficiency since a massive number of clients need to transmit their local gradients to the server frequently \cite{li2020federated}. 
As one of the most popular FL algorithms, the federated averaging (FedAvg) algorithm \cite{mcmahan2017communication,konevcny2016federated} considers  multiple local  SGD updates with periodic communications to reduce the communication cost.
Another way is to compress the local gradients before sending them to the server \cite{li2020federated,alistarh2017qsgd,reisizadeh2020fedpaq}. Among the existing compression methods, a simple yet elegant technique is to take the sign of each coordinate of the local gradients, which requires only one bit for transmitting each coordinate. For any $x\in\Rbb$, we define the sign operator as: $\text{Sign}(x) = 1 \text{ if }x\geq 0$ and $-1$ otherwise. 

% Recently, optimization algorithms with sign-based compression have attracted much attention as they enjoy a great communication efficiency while still achieving similar empirical performance to algorithms without any compression \cite{bernstein2018signsgd,karimireddy2019error,safaryan2021stochastic}. However, for distributed optimization, especially the scenario with heterogeneous data, i.e., $f_i\neq f_j$ for every $i\neq j$, a naive application of sign-based algorithm cannot guarantee convergence \cite{karimireddy2019error,chen2020distributed,safaryan2021stochastic}. 

% A few works have tried to fix this problem for sign-based compression. There are mainly two approaches in the existing literature. The first one is the stochastic sign-based method, which proposes to bring stochasticity into the sign operation, and representative papers adopting this method are \cite{jin2020stochastic,safaryan2021stochastic,chen2020distributed}. The second one is Error-Feedback (EF) method, and the corresponding works include \cite{karimireddy2019error,vogels2019powersgd,tang2019doublesqueeze}. However, there are still some gaps in existing works. First, the theoretical convergence speed of existing algorithms is worse than uncompressed algorithms like \cite{SGD,yu2019parallel}. Second, none of them has shown to be able to deal with periodic communication, which is widely used in FL for reducing the frequency of communication. This work aims at closing these gaps for sign-based algorithms.

It has been shown recently that optimization algorithms with the sign-based compression can enjoy a great communication efficiency while still achieving comparable empirical performance as uncompressed algorithms \cite{bernstein2018signsgd,karimireddy2019error,safaryan2021stochastic}. However, for distributed learning, especially the scenarios with heterogeneous data, i.e., $f_i\neq f_j$ for every $i\neq j$, a naive application of the sign-based algorithm may end up with divergence \cite{karimireddy2019error,chen2020distributed,safaryan2021stochastic}.

\textbf{A counterexample for sign-based distributed gradient descent. }
% Even when there is no gradient noise, distributed signed gradient descent cannot converge to the optimum. 
 Consider the one-dimensional problem with two clients: $\min_{x\in \Rbb}\ (x-A)^2 + (x+A)^2$,
where $A>0$ is some constant. 
For any $x\in[-A,A]$, the averaged sign gradient at $x$ is $\text{Sign}(x-A)+\text{Sign}(x+A)=0,$ i.e., the algorithm never moves. Similar examples are also discussed by \cite{chen2020distributed,safaryan2021stochastic}. The fundamental reason for this undesirable result is the uncontrollable bias brought by the sign-based compression.

There are mainly two approaches to fixing this issue in the existing literature. The first one is the stochastic sign-based method, which introduces stochasticity into the sign operation \cite{jin2020stochastic,safaryan2021stochastic,chen2020distributed}, and the second one is the Error-Feedback (EF) method \cite{karimireddy2019error,vogels2019powersgd,tang2019doublesqueeze}. However, these works are still unsatisfactory. Specifically, on one hand, { both the theoretical convergence rates and empirical performance} of these algorithms are still worse than uncompressed algorithms like \cite{SGD,yu2019parallel}. On the other hand, none of them allows the clients to have multiple local SGD updates within one communication round like the FedAvg, { which thereby are less communication efficient.}
This work aims at addressing these issues and closing the gaps for sign-based methods.

\textbf{Main contributions.}    
Our contributions are summarized as follows.

\begingroup
\renewcommand\labelenumi{(\theenumi)}
\begin{enumerate}
    \item {\bfseries A unified family of stochastic sign operators.}  We show an intriguing fact: The bias brought by the sign-based compression can be flexibly controlled by injecting a proper amount of random noise before the sign operation. In particular, our analysis is based on a novel noisy perturbation scheme with a general symmetric noise distribution, which also provides a unified framework to understand existing stochastic sign-based methods including \cite{jin2020stochastic,safaryan2021stochastic,chen2020distributed}.
    \item {\bfseries The first sign-based FedAvg algorithm. }  In contrast to the existing sign-based methods which do not allow multiple local SGD updates within one communication round, based on the proposed stochastic sign-based compression, we design a novel family of sign-based federated averaging algorithms ($z$-SignFedAvg) that can achieve the best of both worlds: high communication efficiency {and} fast convergence rate.
    \item {\bfseries New theoretical convergence rate analyses. } By leveraging the asymptotic unbiasedness property of the stochastic sign-based compression, we derive a series of theoretical results for $z$-SignFedAvg and demonstrate its improved convergence rates over the existing sign-based methods. In particular, we show that by injecting a sufficiently large uniform noise, $z$-SignFedAvg can have a matching convergence rate with the uncompressed algorithms.
    % \item {\bfseries Application to differentially-private federated learning (DP-FL). }  We remark that adding random noise to the local gradients is also a common practice for privacy protection, especially in DP-FL \cite{geyer2017differentially,agarwal2021skellam,agarwal2018cpsgd}. With this observation, we propose a differentially-private variant of $z$-SignFedAvg, termed DP-SignFedAvg, and empirically show that it can even perform comparably as the uncompressed DP-FedAvg \cite{geyer2017differentially,kairouz2021distributed} under the same privacy budget.  
%    {\color{red} our proposed $z$-SignFedAvg  significantly improve the communication efficiency for the DP-FedAvg \cite{geyer2017differentially,kairouz2021distributed} without degrading its performance too much.
%    (We did not show this in Fig 6 actually)}
\end{enumerate}
\endgroup

% In summary,
% this paper contributes to the field of Federated Learning on both aspects of optimization and privacy protection. From the optimization perspective, we provide a novel theoretical analysis for a family of sign-based algorithms based on the asymptotical unbiasedness of the sign operator combined with proper additive noise. , and also leads to a sign-based algorithm that can also work well with periodic aggregation both theoretically and empirically. Another contribution is w.r.t Differential Privacy. In this work, we prove that taking sign operation after applying the Gaussian Mechanism can amplify the corresponding privacy guarantee. With this result, we propose a differentially private sign-based algorithm that can achieve similar empirical performance to the uncompressed algorithm under different levels of privacy budget.

\textbf{Organization. }
In Section \ref{sec:sign}, the proposed general noisy perturbation scheme for the sign-based compression and its key property, i.e., asymptotic unbiasedness, are presented. Inspired by this result, the main algorithms are devised in Section \ref{sec:algo} together with their convergence analyses under different noise distribution parameters.
% Section \ref{sec:private} mainly focuses on the privacy aspect, where we formally state the privacy amplification effect of sign operation. 
We evaluate our proposed algorithms on real datasets and benchmarks with existing sign-based methods in Section \ref{sec:exp}. Finally, conclusions are drawn in Section \ref{sec:conclusion}.

\textbf{Notations. } For any $x\in\Rbb^d$, we denote $x(j)$ as the $j$-th element of the vector $x$. We define the $\ell_p$-norm for $p\geq 1$ as $\|x\|_p=(\sum_{j=1}^d|x(j)|^p)^{\frac{1}{p}}$. We denote that $\|\cdot\|=\|\cdot\|_2$, and $\|x\|_\infty = \max_{j\in\{1,...,d\}}|x(j)|$. For any function $f(x)$, we denote $f^{(k)}(x)$ as its $k$-th derivative, and for a vector $x=[x(1),...,x(d)]^\top\in \Rbb^d$, we define $\text{Sign}(x) = [\text{Sign}(x(1)),...,\text{Sign}(x(d))]^\top$.

\subsection{Related works}
\label{sec:related}
%  This work focuses on the scenario with heterogeneous data and considers using sign-based compression only for the uplink communication. It is worth mentioning that \cite{tang2019doublesqueeze,jin2020stochastic,chen2020distributed} also compress for the downlink communication.

\textbf{Stochastic sign-based method. } Our proposed algorithm belongs to this category. Among the existing works \cite{safaryan2021stochastic,jin2020stochastic,chen2020distributed}, the setting considered by \cite{safaryan2021stochastic} is closest to ours since the latter two consider gradient compression not only in the uplink but also in the downlink. Despite of this difference and the use of different convergence metrics, the algorithms therein achieve the same convergence rate $O(\tau^{-\frac{1}{4}})$, where $\tau$ is the total number of gradient queries to the local objective function.  Compared to existing works, our proposed $z$-SignFedAvg requires a slightly stronger assumption on the minibatch gradient noise, but achieves a faster convergence rate $O(\tau^{-\frac{1}{3}})$ or even $O(\tau^{-\frac{1}{2}})$, with the  standard squared $\ell_2$-norm of gradients as the convergence metric. 

\textbf{Error-Feedback method.}
The error-feedback (EF) method is first proposed by \cite{seide20141} and later theoretically justified by \cite{karimireddy2019error}. Then, \cite{vogels2019powersgd,tang2019doublesqueeze,tang20211} further extended this EF method into distributed and adaptive gradient schemes.  The key idea of the EF-based methods is to show that the sign operator scaled by the gradient norm is a contractive compressor, and the error induced by the contractive compressor can be compensated. However, such EF-based methods cannot deal with partial client participation otherwise the error residuals cannot be correctly tracked. Besides, the EF-based methods have a convergence rate $\Oc(\tau^{-\frac{1}{2}}+d^2\tau^{-1})$, where $d$ is the dimension of the gradients, and therefore is not competitive for high-dimension problems. 

{
\textbf{Unbiased quantization method.} Apart from the sign-based gradient compression, another popular way of compression is the unbiased stochastic quantization method adopted by \cite{alistarh2017qsgd,reisizadeh2020fedpaq,haddadpour2021federated}. A key assumption made by this category of methods
is that the quantization error is bounded by the norm of the input, which however does not hold for sign-based compression, and therefore the existing convergence results therein do not apply to sign-based methods. 
% However, unlike sign-based compression which only needs two alphabets for each coordinate, the unbiased quantization method requires at least three alphabets, i.e., $-1,1, 0$. 
Besides, as shown in \cite{alistarh2017qsgd,reisizadeh2020fedpaq}, these methods usually have degraded convergence speed when fewer quantization bits are used.
} 

%is $\Oc(\tau^{-\frac{1}{2}}+{\tau^{-1}}/{\delta^2})$, where $\delta$ is the parameter of contractive compressor. In the worst case, the $\delta$ of the sign operator multiplying with one norm is  ${1}/{d}$, where $d$ is the dimension of the gradients. Therefore, the convergence rate  becomes $\Oc(\tau^{-\frac{1}{2}}+d^2\tau^{-1})$, which could become very bad especially for high-dimension optimization problem. 

As mentioned, some of the existing sign-based methods like \cite{chen2020distributed,safaryan2021stochastic} do not adopt the standard squared $\ell_2$-norm of gradients as the metric for the convergence rate analysis. Thus, it is tricky to make a fair comparison between them and the proposed $z$-SignFedAvg. In Appendix \ref{app:discuss}, we provide a detailed discussion and summarize the convergence rates of some representative algorithms in Table \ref{table:rw}. 

% \textbf{Differential Private Federated Learning.} Recent works have discovered that the gradients transmitted from the clients to the server in FL could reveal the clients' privacy \cite{geiping2020inverting}. One way to remedy this is to consider the DP-FL, where the clients' privacy is protected through the lens of Differential Privacy (DP) \cite{dwork2014algorithmic}. In particular, to fulfill the user-level DP \cite{geyer2017differentially}, a common practice is to perturb the gradients by adding some random noise, e.g., the Gaussian noise  \cite{abadi2016deep,geyer2017differentially}. Some DP-FL algorithms such as \cite{agarwal2018cpsgd,kairouz2021distributed} also considered gradient compression and perturbed the compressed gradients by discrete noise . In contrast to these works, our proposed $z$-SignFedAvg adds (continuous valued) noise before the sign compression and therefore can benefit communication efficiency as well as users' privacy in a more straightforward manner. 

%Some works like  also consider the communication efficiency of DP-FL and propose first to compress the transmitted data and then perturb it with discrete noise. 
% One drawback of this scheme is that the privacy guarantee does not depend on the number of bits used, which is not intuitive because we expect it to be more private with fewer bits transmitted. 
%Contrary to existing works, we consider to add noise before compression, because now the injected random noise can help to get the best of both worlds: fix sign-based compression 
%{ and} protect the privacy of local data.

\section{Sign operator with symmetric and zero-mean noise}
\label{sec:sign}
In this section, we introduce a general noisy perturbation scheme for the sign-based compression and analyze the asymptotic unbiasedness of compressed gradients. The results serve as the foundation for the proposed algorithms in subsequent sections.  

% For any $x\in\Rbb$ and $\sigma>0$, we can derive that 
% % \begin{align}
% % 	\Ebb[\text{Sign}(x+\sigma\xi)]&=\text{Sign}(x)\int_{-\frac{|x|}{\sigma}}^{+\infty}p(t)dt +(-\text{Sign}(x))\int_{-\infty}^{-\frac{|x|}{\sigma}}p(t)dt\\
% % 	&= \text{Sign}(x)\left(\int_{-\frac{|x|}{\sigma}}^{+\infty}p(t)dt -\int_{-\infty}^{-\frac{|x|}{\sigma}}p(t)dt\right)\\
% % 	&=\text{Sign}(x)\left(\int_{-\frac{|x|}{\sigma}}^{+\infty}p(t)dt -\int_{\frac{|x|}{\sigma}}^{+\infty}p(t)dt\right)\\
% % 	&=\text{Sign}(x)\int_{-\frac{|x|}{\sigma}}^{\frac{|x|}{\sigma}}p(t)dt=2\int_{0}^{\frac{x}{\sigma}}p(t)dt.\\
% % \end{align}

% \begin{align}
% 	\Ebb[\text{Sign}(x+\sigma\xi)]
% 	&=\text{Sign}(x)\int_{-\frac{|x|}{\sigma}}^{\frac{|x|}{\sigma}}p(t)dt=2\int_{0}^{\frac{x}{\sigma}}p(t)dt.
% \end{align}

\textbf{Key observation.}  Let $\xi$ be a random variable that is symmetric, zero-mean and has the p.d.f $p(t)$. If $p(0)\neq 0$ and $p(t)$ is continuous and uniformly bounded on $(-\infty,+\infty)$, then it holds that
\begin{align}
\label{p:exp_converge}
&\lim_{\sigma\to +\infty}\frac{\sigma}{2p(0)}\Ebb[\text{Sign}(x+\sigma\xi)]\lim_{\sigma\to +\infty}	\frac{\sigma}{p(0)}\int_{0}^{\frac{x}{\sigma}}p(t)dt=x.
\end{align}
In other words, the perturbed sign operator is  an asymptotically unbiased estimator of the input $x$ when $\sigma\to\infty$. Furthermore, assume that $p(t)$ is uniformly bounded on $(-\infty,+\infty)$ and differentiable for an arbitrary order. Then, with the Taylor's expansion, we can have
$\frac{\sigma}{p(0)}\int_{0}^{\frac{x}{\sigma}}p(t)dt=x + \frac{1}{p(0)}\sum_{k=1}^{+\infty}\frac{p^{(k)}(0)x^{k+1}}{(k+1)!\sigma^k} = x+ \sum_{k=1}^{+\infty}p^{(k)}(0)\Oc\left(\sigma^{-k}\right).$ Therefore, suppose that $K$ is the largest integer such that $p^{(1)}(0)=0,...,p^{(K)}(0)=0$. The LHS of \eqref{p:exp_converge} will converge to $x$ with the order $\Oc(\sigma^{-(K+1)})$.
This observation motivates us to propose the following family of noise distribution parameterized by a positive integer $z\in\Zbb_+$.

% \begin{align}
% 	\int_{0}^{+\infty} e^{-t^k}dt = \int_{0}^{+\infty} e^{-y}dy^{\frac{1}{k}} = \int_{0}^{+\infty}\frac{1}{k}y^{\frac{1}{k}-1} e^{-y}dy=\frac{1}{k}\Gamma(\frac{1}{k}).
% \end{align}
\begin{definition}[$z$-distribution] A random variable $\xi_z$ is said to follow the $z$-distribution if its p.d.f is 
	\begin{align}
		\label{p:pdf}
		p_z(t)= \frac{1}{2\eta_z}e^{-\frac{t^{2z}}{2}},
	\end{align}where $\eta_z={2^\frac{1}{2z}\Gamma\left(1+\frac{1}{2z}\right)}$ and $\Gamma(z)=\int_{0}^{+\infty}t^{z-1}e^{-t}dt$ is the Gamma function. 
\end{definition}
It can be verified $p_z(t)$ in \eqref{p:pdf} is a valid p.d.f. When $z=1$, it corresponds to the standard Gaussian distribution. In addition, one can also show that $p_z(t)$ converges to the p.d.f of the uniform random variable on the interval $[-1,1]$ when $z\to+\infty$ (see Lemma \ref{lm:converge2uni} in Appendix \ref{app:formal}). This $z$-distribution has a nice property that can be leveraged to bound the bias caused by the sign-based compression, as stated in the following lemma.

% \begin{lemma}
% \label{lm:phi} For any $x\in \Rbb$
% 	\begin{align}
% 		|x|- \frac{|x|^{2z+1}}{2(2z+1)}	\leq |\Psi_z(x)|\leq |x| ,\text{ where }\Psi_z(x) \stackrel{\rm{def.}}{=} \int_0^x e^{-\frac{t^{2z}}{2}}dt.
% 	\end{align}
% \end{lemma}

% Similar to the vector form of Sign operator, for a vector $x\in\Rbb^d$, we define that
% \begin{align}
% 	\Psi_z(x) = [\Psi_z(x(1)),...,\Psi_z(x(d))]^\top.
% \end{align}

% With Lemma \ref{lm:phi}, we can derive the following important  corollary that bounds the bias caused by sign operation.

\begin{lemma}
	\label{lm:phi_lm} For any $x\in \Rbb^d$ and $\sigma>0$,
		\begin{align}\label{p:col_phi_1}
			\left\|\eta_z\sigma\Ebb\left[\rm{Sign}(x+\sigma \xi_z)\right]-x\right\|^2\leq 	\frac{\|x\|_{4z+2}^{4z+2}}{4(2z+1)^2\sigma^{4z}},
		\end{align}where $\xi_z(1),...,\xi_z(d)$ follow the i.i.d. $z$-distribution.
	\end{lemma}

\begin{remark}\label{rm:limit}
    One can see that the RHS of \eqref{p:col_phi_1}  involves the term $\left({\|x\|_{4z+2}}/{\sigma}\right)^{4z}$. Thus, as long as $\sigma > \|x\|_\infty$, the LHS of \eqref{p:col_phi_1} converges to zero when $z\to+\infty$. 
    Since Lemma \ref{lm:converge2uni} implies that 
    $\xi_\infty$ follows the i.i.d uniform distribution on $[-1,1]$, we obtain $\sigma\Ebb\left[\rm{Sign}(x+\sigma \xi_\infty)\right]=x$ as long as $\sigma> \|x\|_\infty$. It is interesting to remark that the stochastic sign operators proposed in  \cite{jin2020stochastic,safaryan2021stochastic} are exactly the sign operator injected by the uniform noise, and \cite{chen2020distributed} also considered the use of a symmetric noise for gradient perturbation. Thus, sign-based compression with the $z$-distribution offers a unified perspective to understand the relationship among the existing stochastic sign-based methods. 
    
    %In the meanwhile, Lemma \ref{lm:converge2uni} shows that $p_z(t)$ will converge to the uniform distribution at $[-1,1]$. In fact, if $\xi_\infty$ is a vector whose elements follow uniform distribution at $[-1,1]$ i.i.d and $\sigma> \|x\|_\infty$, it is easy to verify that $\sigma\Ebb\left[\rm{Sign}(x+\sigma \xi_\infty)\right]=x.$ We remark that the sign operator with sufficient amount of uniform noise is in fact equivalent to the stochastic sign operators proposed in  \cite{jin2020stochastic,safaryan2021stochastic}.
\end{remark}

\section{$z$-SignFedAvg Algorithm}
\label{sec:algo}

In this section, based on the analysis in Section \ref{sec:sign},  we propose the following sign-based FedAvg algorithm, termed as $z$-SignFedAvg. {While FedAvg-type algorithms with gradient compression are also presented in \cite{haddadpour2021federated}, they require unbiased compression and are not applicable to sign-based methods.} The details of $z$-SignFedAvg are presented in Algorithm \ref{alg:SignFedAvg}. A prominent difference between the proposed $z$-SignFedAvg and the existing sign-based methods lies in that the clients are allowed to perform multiple SGD updates per communication round ($E>1$) before applying the stochastic sign-based compression. Like the FedAvg algorithm, it is anticipated that $z$-SignFedAvg can greatly benefit from this and has a significantly reduced communication cost. 

Note that in practice we only consider $z=1$ and $z=+\infty$ for the $z$-SignFedAvg since they correspond to the Gaussian distribution and uniform distribution, respectively. Nevertheless, we are interested in the convergence properties of $z$-SignFedAvg for a general positive integer $z$ as it provides better insights on the role of $z$ for the convergence rate.

\begin{algorithm}[htbp]
\small
	\begin{algorithmic}[1]
	  \REQUIRE Total communication rounds $T$, number of local steps $E$, number of clients $n$, clients stepsize $\gamma$, server stepsize $\eta$, noise coefficient $\sigma$, parameter  of noise distribution $z$.
	  \STATE Initialize $x_0$.  
	\FOR{$t=1$ to $T$}

	\STATE \textbf{On Clients:}
	\FOR{$i=1$ to $n$}
	\STATE $x_{t-1,0}^i=x_{t-1}$
	\FOR{$s=1$ to $E$}
	\STATE $g_{t-1,s}^{i}=g_i(x_{t-1,s-1}^i)$, where $g_i(\cdot)$ is the minibatch gradient oracle of the $i$-th client.
	\STATE $x_{t-1,s}^i =x_{t-1,s-1}^i - \gamma g_{t-1,s}^{i}$.
	\ENDFOR
 \STATE Sample $\xi_z\in\Rbb^d$ from the distribution $p_z(t)$ i.i.d.
	\STATE $\Delta_{t-1}^{i} = \text{Sign}\left(\frac{x_{t-1}-x_{t-1,E}^i}{\gamma}+\sigma \xi_z\right)$.
	\STATE Send $\Delta_{t-1}^{i}$ to the server.
	\ENDFOR

	\STATE \textbf{On Server:}

      \STATE  $x_t = x_{t-1}- \eta\gamma \frac{1}{n}\sum_{i=1}^n  \Delta_{t-1}^{i}.$ 
	\STATE Broadcast $x_t$ to the clients.
	\ENDFOR
	\end{algorithmic}
	\caption{$z$-SignFedAvg (or $z$-SignSGD when $E=1$)}
	\label{alg:SignFedAvg}
	\end{algorithm}

 We first state some standard assumptions for problem \eqref{p:problem}.
\begin{assumption} \label{asp:common}We assume that each $f_i(x)$ has the following properties:
	\begin{enumerate}
		\item[A.1] The minibatch gradient is unbiased and has bounded variance, i.e., $ \Ebb[g_i(x)]=\nabla f_i(x)$ and $\Ebb[\|g_i(x)-\nabla f_i(x)\|^2_2]\leq \zeta^2$.
		
		\item[A.2] Each $f_i$ is smooth, i.e., for any $x,y\in\Rbb^d$, there exists some non-negative constants $L_1,\ldots,L_d$, such that
			$f(y) - f(x)
		\leq \langle \nabla f(x), y - x \rangle + \frac{ \sum_{j=1}^d L_j \left(y(j) - x(j) \right)^2}{2}.$
		
		\item[A.3]$f$ is lower bounded, i.e., there exists some constant $f^*$ such that
			$f(x)\geq f^*,\forall x\in\Rbb^d.$
		\item[A.4]  There exists a constant $G\geq0$ such that $\|\nabla f_i(x)\|\leq G$, $\forall i=1,...,n$,  and $x\in\Rbb^d$.
	\end{enumerate}
	\end{assumption}

	Assumption A.2 is a more fine-grained assumption on the function smoothness than the commonly used one and is also used by \cite{bernstein2018signsgd,safaryan2021stochastic}. 
	For the convergence rate analysis, we consider two cases, namely,  the case with $z<+\infty$ and the case of $z=\infty$.

\subsection{Case 1: $z<+\infty$}
\label{sec:zlessinf}

% \begin{algorithm}[H]
% 	\begin{algorithmic}[1]
% 	  \REQUIRE Total communication rounds $T$, Number of local steps $E$, Number of clients $n$, Clients stepsize $\gamma$, Server stepsize $\eta$, Noise coefficient $\sigma$.
% 	  \STATE Initialize $x_0$ and for $i=1,...,n.$  
% 	\FOR{$t=1$ to $T$}

% 	\STATE \textbf{On Clients:}
% 	\FOR{$i=1$ to $n$}
% 	\STATE $x_{t-1,0}^i=x_{t-1}$
% 	\FOR{$s=1$ to $E$}
% 	\STATE $g_{t-1,s}^{i}=\text{minibatch gradient at }x_{t-1,s-1}^i$.
% 	\STATE $x_{t-1,s}^i =x_{t-1,s-1}^i - \gamma g_{t-1,s}^{i}$.
% 	\ENDFOR
% 	\STATE $\Delta_{t-1}^{i} = \text{Sign}\left(\frac{x_{t-1}-x_{t-1,E}^i}{\gamma}+\Nc(0,\sigma^2I)\right)$
% 	\STATE Send $\Delta_{t-1}^{i}$ to the server.
% 	\ENDFOR

% 	\STATE \textbf{On Server:}

%       \STATE  $x_t = x_{t-1}- \eta\gamma \frac{1}{n}\sum_{i=1}^n  \Delta_{t-1}^{i}.$ 
	
% 	\ENDFOR
% 	\RETURN $x_T$. 
% 	\end{algorithmic}
% 	\caption{Sign-FedAvg}
% 	\label{alg:Sign-FedAvg}
% 	\end{algorithm}

As we can see from Lemma \ref{lm:phi_lm}, there always exists some gradient bias when $z<+\infty$. In order to bound it, we further assume that a higher order moment of the minibatch gradient noise is bounded. 

% We first study the case with $z<+\infty$. As we can see from Lemma \ref{lm:phi_lm}, there will always be some bias in the transmitted data. In order to bound the bias, we need to assume that a higher order moment of the gradient noise is bounded. We denote $\bar x_{t,s} = \frac{1}{n}\sum_{i=1}^n x_{t,s}^i$ in the following results.

\begin{assumption}\label{asp:z-moment} There exists a constant $Q_z\geq0$ such that for any $x\in \Rbb^d$, we have
	\begin{align}
		\Ebb[\|g_i(x)-\nabla f_i(x)\|_{4z+2}^{4z+2}]\leq Q_z.
	\end{align}
	\end{assumption}

\begin{theorem}
	\label{thm:sign-fedavg_z}
Suppose that Assumption \ref{asp:common} and \ref{asp:z-moment} hold. Denote $\bar x_{t,s} = \frac{1}{n}\sum_{i=1}^n x_{t,s}^i$ and $L_{\max}=\max_{j}L_j$. Then, for
  $\eta = \eta_z\sigma$, $\gamma\leq \frac{1}{L_{\max}}$ and $z<+\infty$ in Algorithm \ref{alg:SignFedAvg}, we have
	% \begin{align}
	% 	\Ebb[\frac{1}{TE}\sum_{t=1}^{T}\sum_{s=1}^E] \|\nabla f(\bar x_{t-1,s-1})\|^2 &\leq \underbrace{ \frac{2\Ebb[f(x_0) - f^*]}{TE\gamma} + \frac{\gamma\xi^2L}{n} + \frac{(E-1)(2E-1)\gamma^2 L^2G^2}{3}}_{\text{standard terms in federated averaging}}\\
	% 	& + \underbrace{\frac{4\sqrt{2} GE^{2}(G^6+Q)^{\frac{1}{2}}}{3\sigma^2} + \frac{8\gamma LE^5 (G^6+Q)}{9\sigma^4}}_{\text{bias terms}}+ \underbrace{\frac{2\gamma \sigma^2\pi L}{En}}_{\text{variance term}}.
	% \end{align}
	
	\begin{subequations}\label{main theorem1}
	\begin{align}
		&\Ebb\left[\frac{1}{TE}\sum_{t=1}^{T}\sum_{s=1}^E \|\nabla f(\bar x_{t-1,s-1})\|^2\right]\notag \\&\leq \underbrace{\frac{2\Ebb[f(x_0) - f^*]}{TE\gamma} + \frac{\gamma\zeta^2L_{\max}}{n} + \frac{4\gamma^2 (E-1)E L_{\max}^2(\zeta^2+G^2)}{3}}_{\text{\rm (a) Standard terms in FedAvg}}\\
		& \label{p:thm1_sig}+\underbrace{ \frac{ 2^{2z+1} E^{2z}\sqrt{Q_z+G^{4z+2}}G}{\sqrt 2(2z+1)\sigma^{2z}} + \frac{\gamma 2^{4z} E^{4z+1}(Q_z+G^{4z+2})L_{\max}}{2(2z+1)^2\sigma^{4z}}}_{\text{\rm (b) Bias terms}}\\ 
		&\label{p:thm1_sig2}+\underbrace{\frac{4\eta_z^2\gamma \sigma^2\sum_{j=1}^d L_j}{En}}_{\text{ \rm (c) Variance term}}.
	\end{align}
	\end{subequations}
\end{theorem}

\textbf{When is the bound non-trivial? }  Since we assume that the $\ell_2$-norm of gradient is bounded by $G$, all the terms in the RHS of \eqref{main theorem1} should be no larger than $G^2$. For example, to have the first term in \eqref{p:thm1_sig} less than $G^2$, one 
requires $\sigma$ to be greater than ${ {2^{1+\frac{1}{4z}}E\left(Q_z/G+G^{4z}\right)^{\frac{1}{4z}}}/{(2z+1)^{\frac{1}{2z}}}}
$.

\textbf{Bias-variance trade-off.} An interesting observation from Theorem 1 is that there exists a trade-off between the bias and variance terms. One can see that the terms in \eqref{p:thm1_sig} is caused by the gradient bias of the sign operation (see \eqref{p:col_phi_1}) and is an infinitesimal of $\sigma$ with $\Oc\left({\sigma^{-2z}}\right)$, while the term in \eqref{p:thm1_sig2} is due to the injected noise and is in the order of $\Oc\left({\gamma\sigma^{2}}\right)$.
 Specifically, the first term in \eqref{p:thm1_sig} only depends on the noise scale $\sigma$ and mostly affects the final objective. Meanwhile, the variance term in \eqref{p:thm1_sig2} mainly affects the convergence speed because a smaller stepsize is required for it to diminish.

Theoretically, we can choose an iteration-dependent noise scale $\sigma$ so as to make the algorithm converge to a stationary solution. To see this, let us denote $\tau=TE$ as the total number of gradient queries per client, and present the following corollary.

% \begin{corollary}
	% Denote $\tau=TE$,
	% if we choose $\gamma=\min\{n^{\frac{1}{3}}\tau^{-\frac{2}{3}},\frac{1}{L}\}$ and $\sigma=(n\tau)^{\frac{1}{6}}$ in Theorem \ref{thm:sign-fedavg}, we have

% 	\begin{align}
% 		\Ebb[\frac{1}{\tau}\sum_{t=1}^{T}\sum_{s=1}^E] \|\nabla f(\bar x_{t-1,s-1})\|^2 &\leq \frac{2\Ebb[f(x_0) - f^*]}{(n\tau)^{\frac{1}{3}}} + \frac{\xi^2L}{(n\tau)^{\frac{2}{3}}} + \frac{(E-1)(2E-1)n^{\frac{2}{3}}L^2G^2}{3\tau^{\frac{4}{3}}}\\
% 		& + \frac{4\sqrt{2} GE^{2}(G^6+Q)^{\frac{1}{2}}}{3(n\tau)^{\frac{1}{3}}} + \frac{8 LE^5 (G^6+Q)}{9n^{\frac{1}{3}}\tau^{\frac{4}{3}}}+ \frac{2\pi L}{E(n\tau)^{\frac{1}{3}}}.
% 	\end{align}

% 	If $E\leq \sqrt{\frac{\tau}{n}}$, we have 
% 	\begin{align}
% 		\Ebb[\frac{1}{\tau}\sum_{t=1}^{T}\sum_{s=1}^E] \|\nabla f(\bar x_{t-1,s-1})\|^2 &\leq \frac{2\Ebb[f(x_0) - f^*]}{(n\tau)^{\frac{1}{3}}} + \frac{\xi^2L}{(n\tau)^{\frac{2}{3}}} + \frac{2L^2G^2}{3(n\tau)^{\frac{1}{3}}}\\
% 		& + \frac{4\sqrt{2} GE^{2}(G^6+Q)^{\frac{1}{2}}}{3(n\tau)^{\frac{1}{3}}} + \frac{8 LE^5 (G^6+Q)}{9n^{\frac{1}{3}}\tau^{\frac{4}{3}}}+ \frac{2\pi L}{E(n\tau)^{\frac{1}{3}}},
% 	\end{align} i.e., we need $(n\tau)^\frac{1}{2}$ communication rounds to achieve a linear speedup.
% \end{corollary}

\begin{corollary}[Informal]
	\label{col:thm1_informal}
	Let $\sigma=(n\tau)^{\frac{1}{4z+2}}$ and $\gamma=\min\{n^{\frac{z}{2z+1}}\tau^{-\frac{z+1}{2z+1}},{L^{-1}_{\max}}\}$ in Theorem \ref{thm:sign-fedavg_z}, and let $E\leq n^{-\frac{3z}{4z+2}}\tau^{\frac{z+2}{4z+2}}$. We have
	\begin{align}
		\Ebb\left[\frac{1}{\tau}\sum_{t=1}^{T}\sum_{s=1}^E \|\nabla f(\bar x_{t-1,s-1})\|^2\right] &=\Oc\left((n\tau)^{-\frac{z}{2z+1}}\right).
	\end{align}
% 	\begin{align}
% 		\Ebb[\frac{1}{\tau}\sum_{t=1}^{T}\sum_{s=1}^E] \|\nabla f(\bar x_{t-1,s-1})\|^2 &\leq {\frac{2\Ebb[f(x_0) - f^*]}{(n\tau)^{\frac{z}{2z+1}}} + \frac{\zeta^2L_{\max}}{(n\tau)^{\frac{z+1}{2z+1}}} + \frac{2 L_{\max}^2G^2}{3(n\tau)^{\frac{z}{2z+1}}}}\\
% 		& +{ \frac{ 2^{2z} E^{2z}\sqrt{Q_z+G^{4z+2}}G}{(2z+1)(n\tau)^{\frac{z}{2z+1}}} + \frac{ 2^{4z} E^{4z+1}(Q_z+G^{4z+2})L_{\max}}{2(2z+1)^2n^{\frac{z}{2z+1}}\tau^{\frac{3z+1}{2z+1}}}}\\ 
% 		&+{\frac{2^{\frac{1}{z}}(\Gamma(\frac{1}{2z}))^2\sum_{j=1}^d L_j}{z^2E(n\tau)^{\frac{z}{2z+1}}}},
% 	\end{align} 
\end{corollary}

\textbf{Achieveing linear speedup. }
From Corollary \ref{col:thm1_informal}, we can see that the $z$-SignFedAvg needs $(n\tau)^{\frac{3z}{4z+2}}$ communication rounds to achieve a linear-speedup convergence rate. Particularly, when $z=1$, the corresponding convergence rate is $\Oc((n\tau)^{-\frac{1}{3}})$ and the required communication rounds is $(n\tau)^{\frac{1}{2}}$. To the best of our knowledge, the previous works have never shown the sign-based method can achieve a linear-speedup convergence rate.

%such linear speedup analysis under sign-based compression is not known in previous works. 

% \begin{remark}
%     Corollary \ref{col:thm1} show that linear speedup can also be achieved  for sign-based distributed optimization algorithmm.
% \end{remark}

\textbf{Relationship to \cite{chen2020distributed}.}
The work \cite{chen2020distributed} also considered the use of a symmetric and zero-mean noise for the sign-based compression and proved that the algorithm has a convergence rate $\Oc(\tau^{-\frac{1}{4}})$. However, their results have three differences from our $z$-SignFedAvg and Theorem \ref{thm:sign-fedavg_z}. First, \cite{chen2020distributed} considered   gradient compression both in the uplink and downlink communications. In addition, the convergence metric they used is not the standard squared $\ell_2$-norm of gradients and is hard to interpret. Second, their analysis is rooted in the median-based algorithm, whereas we judiciously exploit the property of the sign operation and hence provide a general analysis framework for the stochastic sign-based methods. Last but not the least, unlike our $z$-SignFedAvg, \cite{chen2020distributed} cannot allow multiple local  SGD updates. 

% Despite of the difference, we remark that the convergence rate  $\Oc(\tau^{-\frac{1}{4}})$ in their result can be understanded in a more clear way using our framework. Specifically, from \eqref{p:asp_unbiased_symmetric} we have seen that for a general symmetric and zero-mean noise, the bias term is $\Oc(\frac{1}{\sigma})$, while the order of bias term is $\Oc(\frac{1}{\sigma^{2z}})$ for $z$-distribution. Therefore, it can be proved that the order of convergence rate with arbitrary symmetric and zero-mean noise can be obtained by setting $z=\frac{1}{2}$ in  Corollary \ref{col:thm1}, which recovers the $\Oc(\tau^{-\frac{1}{4}})$ convergence rate.

\subsection{Case 2: $z=+\infty$}
\label{sec:zinf}
When $z=+\infty$, the injected noise $\xi_z$ in the $z$-SignFedAvg is uniformly distributed on $[-1,1]$. 
From Remark \ref{rm:limit}, we have learned that the gradient bias can vanish as long as the noise scale $\sigma$ is sufficiently large. To quantify this threshold, we need the following assumption which is a limit form of Assumption \ref{asp:z-moment}.
\begin{assumption}\label{asp:inf} There exists a constant $Q_\infty\geq0$ such that for any $x\in \Rbb^d$, with probability 1,
	\begin{align}
		\|g_i(x)-\nabla f_i(x)\|_\infty\leq Q_\infty.
	\end{align}
	\end{assumption}

\begin{theorem}(Informal)
	\label{thm:sign-fedavg_inf_informal}
Suppose that Assumption \ref{asp:common} and  \ref{asp:inf} hold. For $\gamma=\min\{n^{\frac{1}{2}}\tau^{-\frac{1}{2}},{L^{-1}_{\max}}\}$, $\eta=\sigma$, $z=+\infty$, $E\leq n^{-\frac{3}{4}}\tau^{\frac{1}{4}}$ and $\sigma>E(G+Q_\infty)$  in Algorithm \ref{alg:SignFedAvg} we have
\begin{align}
		\Ebb\left[\frac{1}{\tau}\sum_{t=1}^{T}\sum_{s=1}^E \|\nabla f(\bar x_{t-1,s-1})\|^2\right] &=\Oc\left((n\tau)^{-\frac{1}{2}}\right).\label{thm2 bound2}
	\end{align}
\noindent However, if $\sigma\leq E(G+Q_\infty)$, there exists a problem instance for which Algorithm \ref{alg:SignFedAvg} cannot converge.
\end{theorem}
% As we can see, the bound in Theorem \ref{thm:sign-fedavg_inf} is almost the same as the algorithm \cite{yu2019parallel} without any compression.

\begin{remark}
Note that Theorem \ref{thm:sign-fedavg_inf_informal} implies that $\infty$-SignFedAvg has a matching convergence rate as the uncompressed FedAvg. The reason why $\infty$-SignFedAvg cannot converge when $\sigma\leq E(G+Q_\infty)$ is simply that the uniform noise has a finite support and cannot always change the sign of gradients. For example, %if $\xi_\infty$ follows uniform distribution on $[-1,1]$, and now 
if $\sigma<A$ for some $A>0$, then we have $	\text{\rm Sign}(x+\sigma \xi_\infty)=\text{\rm Sign}(x)$ for any $x\geq A$.
\end{remark}

% Similarly, the algorithm will converge if we select the stepsize properly.
% \begin{corollary}
% 	\label{col:thm2}
% 	If we choose $\gamma=\min\{n^{\frac{1}{2}}\tau^{-\frac{1}{2}},\frac{1}{L_{\max}}\}$ in Theorem \ref{thm:sign-fedavg_inf}, we have

% 	\begin{align}
% 		\Ebb[\frac{1}{\tau}\sum_{t=1}^{T}\sum_{s=1}^E] \|\nabla f(\bar x_{t-1,s-1})\|^2 &\leq {\frac{2\Ebb[f(x_0) - f^*]}{(n\tau)^{\frac{1}{2}}} + \frac{\zeta^2L_{\max}}{(n\tau)^{\frac{1}{2}}} + \frac{(E-1)(2E-1)n L_{\max}^2G^2}{3\tau}}+{\frac{4\sigma^2\sum_{j=1}^d L_j}{E(n\tau)^{\frac{1}{2}}}}.
% 	\end{align}
% 	Furthermore, if $E\leq n^{-\frac{3}{4}}\tau^{\frac{1}{4}}$, the upper bound above will converge as $\Oc\left((n\tau)^{-\frac{1}{2}}\right)$, which recovers the convergence result of uncompressed algorithm \cite{yu2019parallel}.
% \end{corollary}

\textbf{Relationship to \cite{jin2020stochastic,safaryan2021stochastic}. }
As mentioned in Remark \ref{rm:limit}, both the stochastic sign operators in \cite{jin2020stochastic,safaryan2021stochastic} are equivalent to the sign operator injected by the uniform noise. 
Nevertheless, there are still two distinctions when compared with our $\infty$-SignFedAvg. First, while \cite{safaryan2021stochastic} shows their algorithm has a $\Oc(\tau^{-\frac{1}{4}})$ convergence rate, it is based on the $\ell_2$-norm of gradients and cannot imply the same rate as that in \eqref{thm2 bound2} (see Appendix \ref{app:discuss}). Second, although \cite{safaryan2021stochastic} does not need Assumption \ref{asp:inf}, it relies on an input-dependent noise scale which, unfortunately, often slows the algorithm convergence in practice especially when the problem dimension is large. 
%In particular, \cite{jin2020stochastic} also consider downlink compression and hence its convergence results are not directly comparable to the Case 2. 

%\cite{safaryan2021stochastic} adopts an input-dependent noise scale and proves $\Oc(\tau^{-\frac{1}{4}})$ convergence rate with $\ell_2$ norm of gradient as the metric. We remark that this rate is usually worse than the rate $\Oc(\tau^{-\frac{1}{2}})$ with squared $\ell_2$ norm as the metric.  Such input-dependent noise scale make it possible to prove convergence without the bounded support of gradient noise assumed in this work. But there are two disadvantages for this choice. First, it can not be extended to federated averaging algorithm. Second, it often leads to slow convergence in practice when the problem dimension is very high. 

\begin{table}[htpb]
		\caption{Comparison of Case 1 and Case 2.}
	\label{table:compare}
    \centering
	\begin{tabular}{|c| c| c| c|} 
	 \hline
	 \makecell{\bfseries Case} &\makecell{\bfseries Convergence\\ \bfseries rate}& \makecell{\bfseries Threshold \\\bfseries  on $\sigma$} & \makecell{\bfseries Assumption on\\\bfseries gradient noise} \\ 
	 \hline\hline 
	\makecell{$z<+\infty$}&   \makecell{ $\Oc(\tau^{-\frac{z}{2z+1}})$} &\makecell{ ${ {\widetilde \Oc\left(\left(\frac{Q_z}{G}+G^{4z}\right)^{\frac{1}{4z}}\right)}}
$}
& \makecell{ Assumption 2} \\ \hline
	\makecell{ $z=+\infty$ } &  \makecell{$\Oc(\tau^{-\frac{1}{2}})$} & \makecell{$\widetilde \Oc(Q_\infty+G)$} & 
	 \makecell{ Assumption 3}
	  \\ \hline
	\end{tabular}
	\end{table}

More theoretical results and proofs are relegated to Appendix \ref{app:formal} and \ref{app:proof}. {Below, we have two more remarks.}

%\red{
%\subsection{Discussion}
%
{ 
\begin{remark}
\textbf{(Bounded minibatch gradient noise)} 
While both Assumption \ref{asp:z-moment} and \ref{asp:inf} are slightly stronger than the commonly used second-order condition on the minibatch gradient noise, they are still justifiable since unbounded minibatch gradient noise is rarely to happen in practice.
\end{remark}

\begin{remark}\label{sec:discussion}
\textbf{(Minibatch gradient noise works as noise perturbation)} 
When the minibatch gradient is used as the input of the sign operator in \eqref{p:exp_converge}, the minibatch gradient noise itself may function as the perturbation noise. In particular, as shown in \cite{chen2020understanding} the minibatch gradient noise approximately follows a symmetric distribution.
Therefore, in practice, one may not need to inject as large noises as suggested by Theorem \ref{thm:sign-fedavg_inf_informal} since the minibatch gradient noise can also help mitigate the bias due to sign-based compression.
This also explains why a small noise scale is sufficient for $z$-SignFedAvg to achieve good performance in the experiment section. 

%
%An interesting 

%As shown in \cite{chen2020understanding} the minibatch gradient noise usually follows an approximately symmetric distribution. Therefore, as is similarly discussed in \cite{chen2020distributed}, from \eqref{p:exp_converge} we can see that the minibatch gradient noise itself may also help to mitigate the bias in sign-based compression. In this case, the optimal noise scale for real applications can be much smaller than the theoretically suggested one, which is also confirmed in the following experiment section. However, we do not consider this in previous theoretical results, because, it is generally unrealistic to place any symmetric assumption on the distribution of minibatch gradient noise.
\end{remark}
}

% Comparing to the \cite{safaryan2021stochastic}, our results explicitly observe and make use of the unbiasedness of sign operator with uniform distribution and hence have nearly the same theoretical results with uncompressed algorithm \cite{SGD,yu2019parallel}, though with an 

\subsection{Comparison of Case 1 and Case 2} We summarize the results of Case 1 and Case 2 in Table \ref{table:compare}, where $\widetilde O(\cdot)$ hides some constants that do not affect the comparison. Especially, we can see that when the mini-batch gradient noise has a long tail such that $Q_z/G \ll Q_\infty^{4z}$, Case 1 requires a less amount of noise than Case 2 for guaranteeing convergence. Despite of the difference in theory, we will see in Section \ref{sec:exp} that $z$-SignFedAvg under Case 1 and Case 2 have almost the same behavior in practice.

\subsection{Implication on  Differentially Private Federated Learning (DP-FL)} Beyond the convergence issue, adding Gaussian noise to the local gradients is also a common practice for privacy protection, especially in DP-FL \cite{geyer2017differentially,agarwal2021skellam,agarwal2018cpsgd}. With this observation, it is straightforward to propose a differentially-private variant of $z$-SignFedAvg, which we term DP-SignFedAvg. More details and 
comparison results between DP-SignFedAvg and the uncompressed DP-FedAvg \cite{geyer2017differentially,kairouz2021distributed} under different privacy budgets are given in Appendix \ref{app:DP}.
\begin{figure}[htbp]
    % \begin{minipage}[t]{1\linewidth}
        \centering
        \begin{subfigure}[b]{0.27\textwidth}
        \centering
        \includegraphics[width=\textwidth]{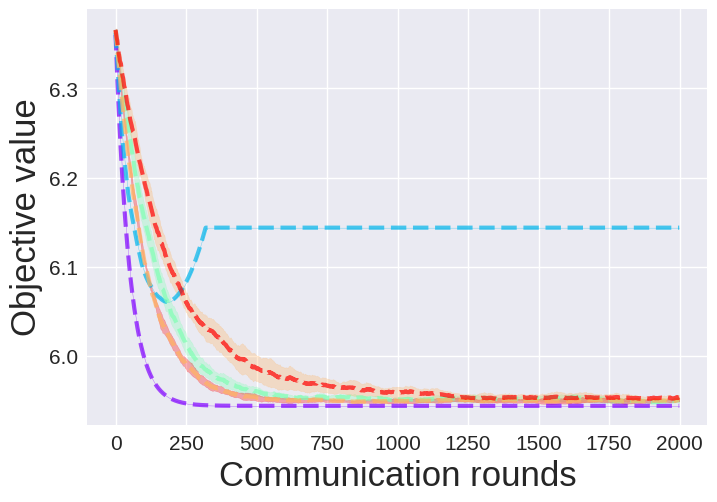}
        \caption{$d=10$}
    \end{subfigure}
    \hfill
        \begin{subfigure}[b]{0.27\textwidth}
        \centering
        \includegraphics[width=\textwidth]{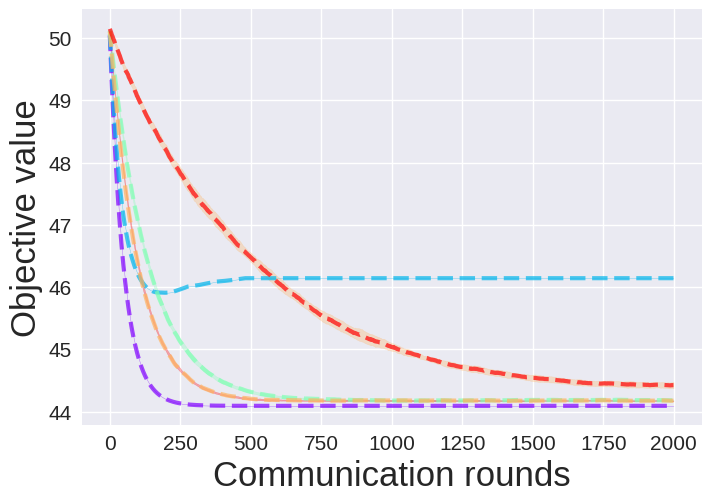}
        \caption{$d=100$}
        \label{}
    \end{subfigure}
    \hfill
        \begin{subfigure}[b]{0.27\textwidth}
        \centering
        \includegraphics[width=\textwidth]{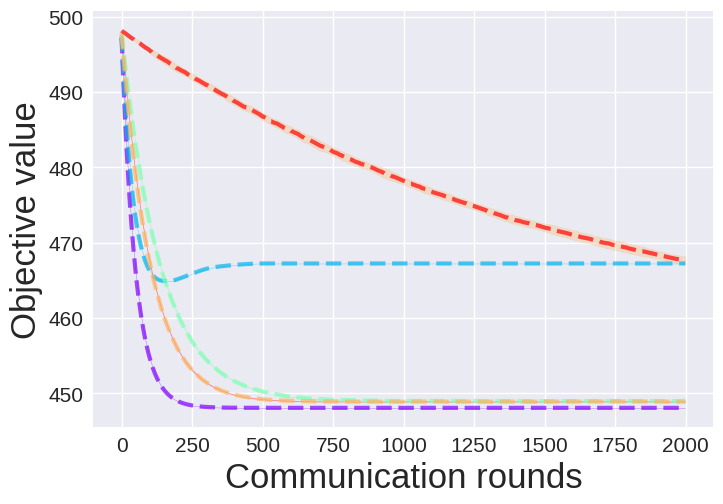}
        \caption{$d=1000$}
        \label{}
    \end{subfigure}
    \begin{subfigure}[][-80pt][b]{0.15\textwidth}
        \centering
        \includegraphics[width=\textwidth]{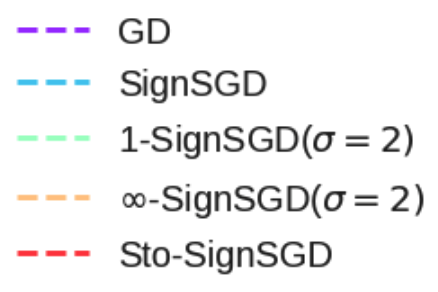}
    \end{subfigure}

    \caption{Performance of tested algorithms under different problem dimension.}
        \label{fig:Sign-SGD-Simple}
    % 	\end{minipage}
    % 	\begin{minipage}[t]{0.15\textwidth}
    % \centering
    % \includegraphics[width= \textwidth ]{exp1_legend.png}
    % \end{minipage}
    \end{figure}

\section{Experiments}
\label{sec:exp}

In this section, we present the experiment results on both synthetic and real problems, and all the figures in this section are obtained by 10 independent runs and are visualized in the form of mean$\pm$std.

\textbf{Noise scale as a hyperparameter. } Although we explicitly characterize how the performance of $z$-SignFedAvg depends on the noise scale $\sigma$ in the previous section, we treat $\sigma$ as a tunable hyperparameter in the experiments.{ This is because, on one hand, the theoretical lower bound for $\sigma$ are difficult to compute since it is impossible to access the moment condition of the minibatch gradient noise. On the other hand, as we have discussed in Remark \ref{sec:discussion}, owing to the presence of the minibatch gradient noise, we can use a much smaller noise scale than the theoretical one in practice.

Aside from the experiments presented in this section, we also compare our algorithm to another popular family of unbiased stochastic compressed FL algorithms, namely, the QSGD in \cite{alistarh2017qsgd} and FedPAQ in \cite{reisizadeh2020fedpaq}. For detailed results, we refer readers to Appendix \ref{app:QSGD}.
}

%Because, on one hand, it usually impossible to compute the moment and support of the gradient noise in reality. One the other hand, since the theoretical results above only provide a worst-case guarantee, for some real problems, the optimal noise scales selected from grid search can be much smaller than the choice suggested by theory.

\subsection{A simple consensus problem}
\label{sec:exp1}
In this section, we verify our theoretical results in Section \ref{sec:algo} by considering the simple consensus problem with 10 clients:
	$\min_{x\in \Rbb^d} \frac{1}{2}\sum_{i=1}^{10}\|x-y_i\|^2,$ where $y_1,...,y_{10}\in \Rbb^d$ are generated using i.i.d standard Gaussian distribution, and $d$ is the problem dimension. We implemented the following algorithms: GD (Gradient descent), Sto-SignSGD \cite{safaryan2021stochastic}, SignSGD (Algorithm \ref{alg:SignFedAvg} with $z=1$, $E=1$ and $\sigma=0$), $1$-SignSGD (Algorithm \ref{alg:SignFedAvg} with $z=1$ and $E=1$.), $\infty$-SignSGD (Algorithm \ref{alg:SignFedAvg} with $z=+\infty$ and $E=1$).  For all the algorithms, we considered the full gradient (no minibatch SGD), and used the same stepsize $0.01$ and initialization by a zero vector.

\textbf{Results. }As we can see from Figure \ref{fig:Sign-SGD-Simple}, the vanilla SignSGD  fails to converge to the optimal solution whereas the others can. Besides, $1$-SignSGD and $\infty$-SignSGD have roughly the same convergence speed which is slightly slower than the uncompressed GD. It is also observed that the input-dependent noise scale adopted by \cite{safaryan2021stochastic} could slow the convergence when the problem dimension is high, as discussed in Section \ref{sec:zinf}.

\begin{figure}[htpb]
	\centering
	\begin{subfigure}[b]{0.4\textwidth}
	\centering
	\includegraphics[width=\textwidth]{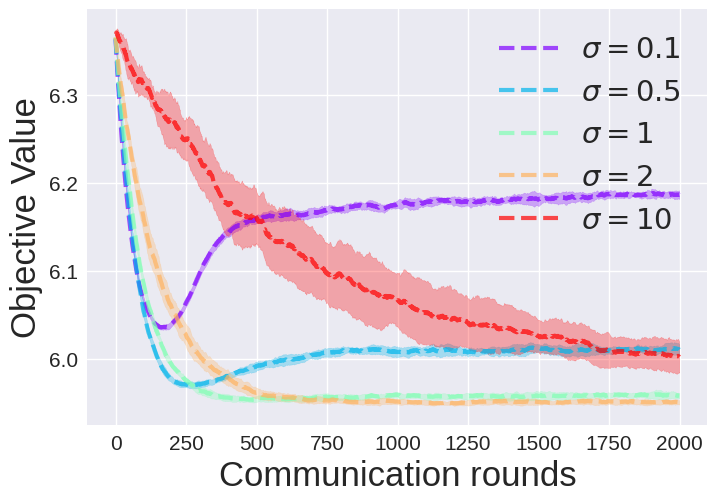}
 	\caption{$1$-SignSGD}
	\label{}
\end{subfigure}
	\begin{subfigure}[b]{0.4\textwidth}
	\centering
	\includegraphics[width=\textwidth]{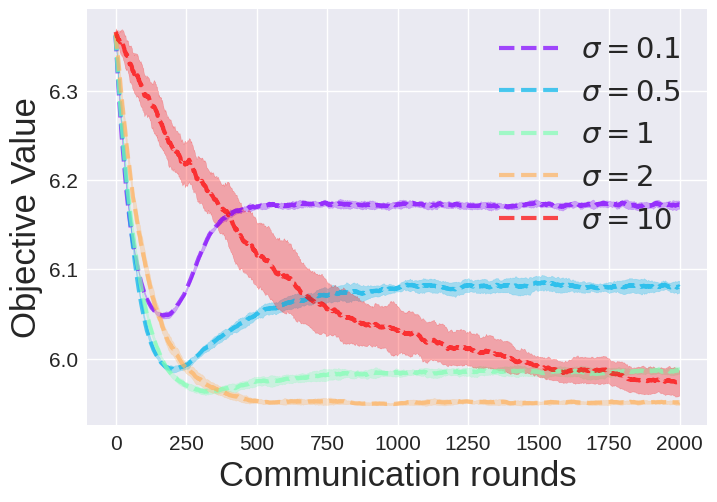}
	\caption{$\infty$-SignSGD}
	\label{}
\end{subfigure}
\caption{$z$-SignSGD under various noise scales.}
	\label{fig:Sign-SGD-Simplen_noise}
\end{figure}
Figure \ref{fig:Sign-SGD-Simplen_noise} displays the results of $1$-SignSGD and $\infty$-SignSGD with various noise scales.
We can see that there is a clear bias-variance trade-off for different noise scales and it corroborates our analysis after Theorem \ref{thm:sign-fedavg_z}. It is also worth mentioning that the best choice of $\sigma$ for Algorithm \ref{alg:SignFedAvg} shown in Figure \ref{fig:Sign-SGD-Simplen_noise} is much smaller than the one predicted by the theorems.
% \begin{figure}[htbp]
% 	\centering
% 	\includegraphics[width=12cm]{Sign-SGD-Simple.png}
% 	\caption{A simple experiment to verify Theorem \ref{thm:signsgd}}
% 	\label{fig:Sign-SGD-Simple}
% \end{figure}

\begin{figure}[htbp]
	\centering
	\begin{subfigure}[b]{0.26\textwidth}
	\centering
	\includegraphics[width=\textwidth]{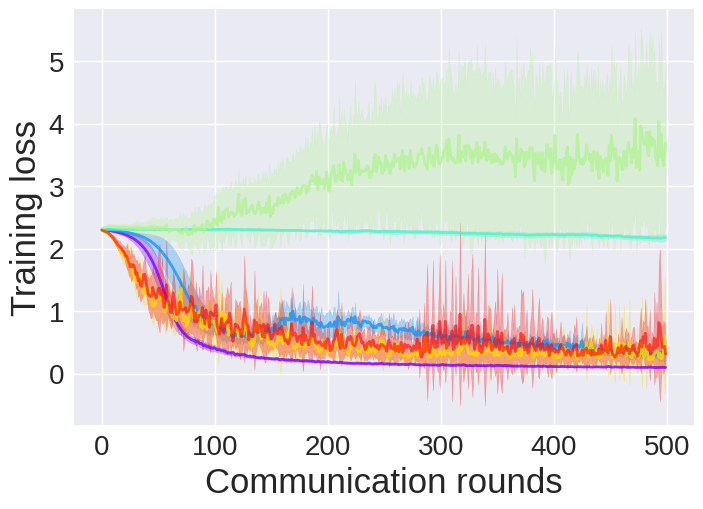}
	\caption{Training Loss}
	\label{fig:exp2_train}
\end{subfigure}
	\begin{subfigure}[b]{0.27\textwidth}
	\centering
	\includegraphics[width=\textwidth]{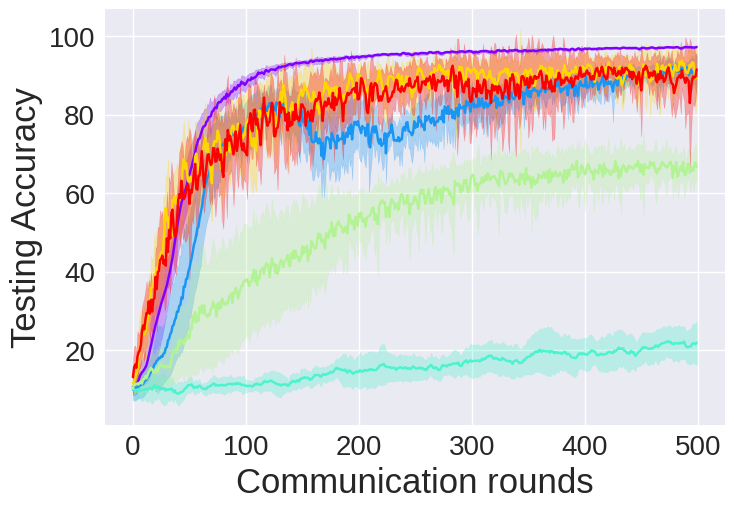}
	\caption{Test Accuracy}
	\label{fig:exp2_test}
\end{subfigure}
\begin{subfigure}[b]{0.27\textwidth}
	\centering
	\includegraphics[width=\textwidth]{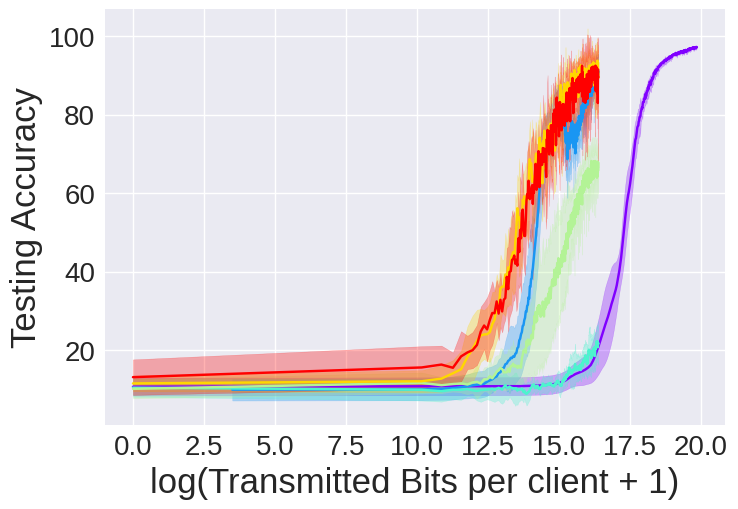}
	\caption{Test Accuracy w.r.t bits}
	\label{fig:exp2_bit}
\end{subfigure}
\begin{subfigure}[][-80pt][b]{0.15\textwidth}
	\centering
	\includegraphics[width=\textwidth]{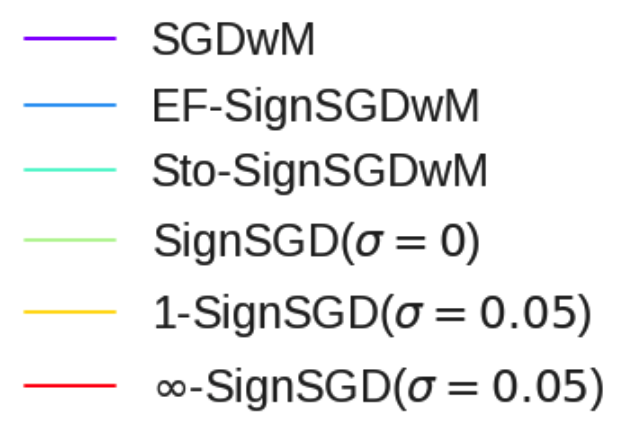}
\end{subfigure}
\caption{Performance of various SignSGD algorithms on non-i.i.d MNIST.}
\label{fig:exp2}
\end{figure}

\subsection{$z$-SignSGD on Non-i.i.d MNIST}
\label{sec:exp2}
In this section, we consider an extremely non-i.i.d setting with the MNIST dataset \cite{deng2012mnist}. Specifically, we split the dataset into 10 parts based on the labels and each client has the data of one digit only. A simple two-layer convolutional neural network (CNN) from Pytorch tutorial \cite{paszke2017automatic}  was used. The following algorithms were implemented: SGDwM (Distributed SGD \cite{SGD} with momentum), EF-SignSGDwM (Distributed SignSGD with error-feedback and momentum \cite{karimireddy2019error,vogels2019powersgd}), and Sto-SignSGDwM (Sto-SignSGD with momentum \cite{safaryan2021stochastic}). 
%Note that some baseline algorithms have an additional hyperparameter for the momentum. 
For each of the algorithms, we selected its best hyperparameters, including the stepsize, momentum coefficient and the noise scale, via grid search (see Appendix \ref{app:exp2}). 

%For more details like network architecture,  hyperparameters for all the tested algorthms and the performance of $1$-SignSGD and $\infty$-SignSGD under different noise scales, we refer the readers to Appendix \ref{app:exp2}.

\textbf{Results.} One can observe from Figure \ref{fig:exp2_train}-\ref{fig:exp2_test} that again the vanilla SignSGD does not converge well. The proposed $1$-SignSGD and $\infty$-SignSGD clearly outperform the existing 
EF-SignSGDwM and Sto-SignSGDw, and perform closely to the uncompressed SGDwM. The reason for the slow convergence of Sto-SignSGDw is that the injected noise is too large due to the input-dependent noise scale. 
Figure \ref{fig:exp2_bit} further displays the testing accuracy of all methods versus the accumulated number of bits transmitted from the clients to the server. One can see that the proposed algorithms achieve the state-of-the-art performance on this task. More results for $1$-SignSGD and $\infty$-SignSGD under different noise scales are presented in Appendix \ref{app:exp2}.

%, we can observe that $1$-SignSGD and $\infty$-SignSGD have roughly the same performance which  outperform other sign-based algorithms and is slightly inferior to the uncompressed algorithm. Our theory is, again, verified by comparing the performance of noiseless SignSGD and our proposed algortihms. If we compare the performance with respect to the total number of transmitted bits, our algorithms achieve the state-of-the-art performance on this task as we can see in Figure \ref{fig:exp2_bit}. 

%the performance of $1$-SignSGD and $\infty$-SignSGD under different noise scales, we refer the readers to Appendix \ref{app:exp2}.

\subsection{$z$-SignFedAvg on EMNIST and CIFAR-10}
\label{sec:exp3}
In this section, we evaluate the performance of our proposed $z$-SignFedAvg on two classical datasets: EMNIST\cite{cohen2017emnist} and CIFAR-10 \cite{krizhevsky2010convolutional}. 
In particular, the proposed $z$-SignFedAvg with $z=1$ and $z=\infty$ are benchmarked against the uncompressed FedAvg \cite{mcmahan2017communication,yu2019parallel}. Since $1$-SignFedAvg and $\infty$-SignFedAvg behave similarly, we only report the results of $1$-SignFedAvg in this section and relegate the others to Appendix \ref{app:exp3}. For EMNIST, we use the same 2-layer CNN as the one in Section \ref{sec:exp2}. For CIFAR-10, we used the ResNet18 \cite{he2016deep} with group normalization \cite{wu2018group}.

{\bfseries Settings. } For both the experiments on EMNIST and CIFAR-10, we followed a setting similar to \cite{reddi2020adaptive}. We also considered the scenario with partial client participation. For the EMNIST dataset, there are 3579 clients in total and 100 clients were uniformly sampled in each communication round to upload their compressed gradients.
For the CIFAR-10 dataset, the training samples are partitioned among 100 clients, and each client has an associated multinomial distribution over labels drawn from a symmetric Dirichlet distribution with parameter 1. In each communication round, 10 out of 100 clients were uniformly sampled. 
The same noise scales for $1$-SignFedAvg and $\infty$-SignFedAvg were used: $\sigma=0.01$ for EMNIST and $\sigma=0.0005$ for CIFAR-10.  
More details about the hyperparameters are referred to Appendix \ref{app:exp3}.

\begin{figure}[htbp]
	\centering
\begin{subfigure}[b]{0.27\textwidth}
	\centering
	\includegraphics[width=\textwidth]{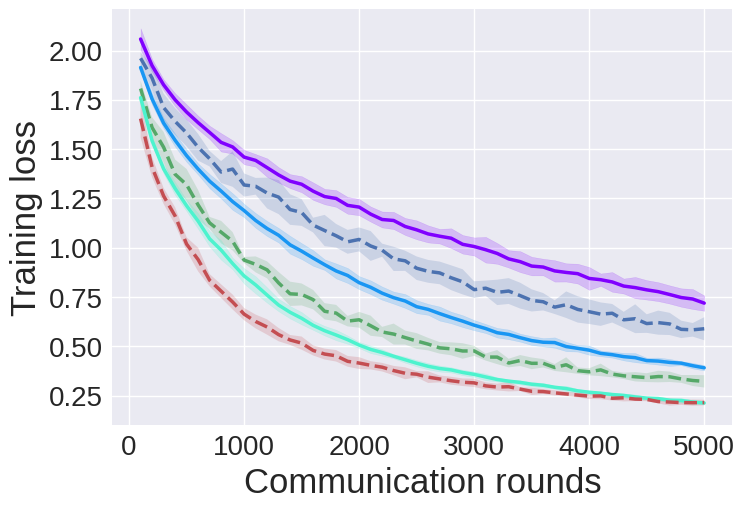}
	\caption{Training Loss}
	\label{}
\end{subfigure}
	\begin{subfigure}[b]{0.26\textwidth}
	\centering
	\includegraphics[width=\textwidth]{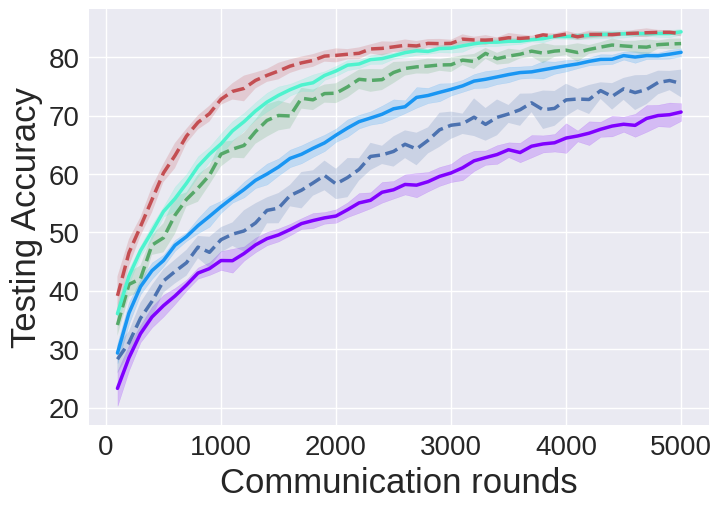}
	\caption{Test Accuracy}
	\label{}
\end{subfigure}
\begin{subfigure}[b]{0.26\textwidth}
	\centering
	\includegraphics[width=\textwidth]{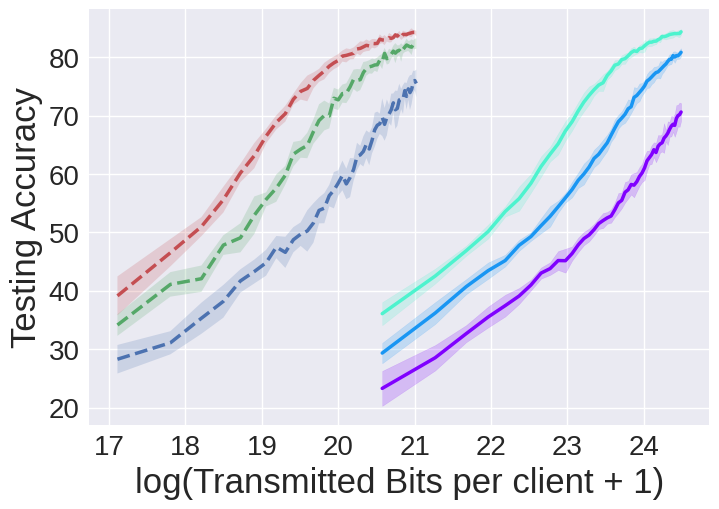}
	\caption{Test Accuracy w.r.t bits}
	\label{}
\end{subfigure}
\begin{subfigure}[][-80pt][b]{0.15\textwidth}
	\centering
	\includegraphics[width=\textwidth]{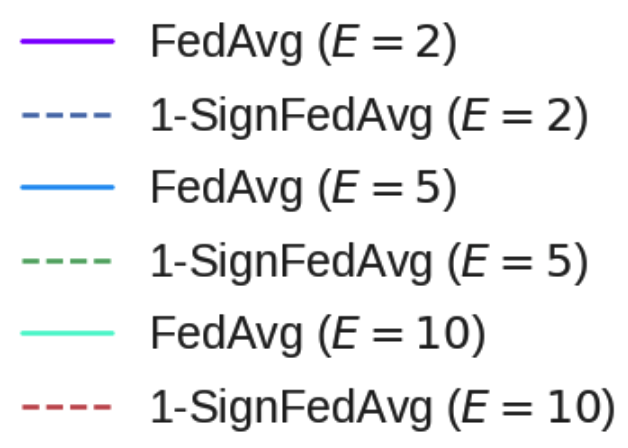}
\end{subfigure}
\vspace{-0.2cm}
\caption{Performance of FedAvg and $1$-SignFedAvg on the CIFAR-10 dataset. }
\label{fig:exp3-cifar}
\end{figure}

{\bfseries Results. } 
%The adopted network architecture and hyperparameters for the tested algorithms are in Appendix \ref{app:exp3}. 
%We use the same noise scale $\sigma$ for both $1$-SignFedAvg and $\infty$-SignFedAvg: $\sigma=0.01$ for EMNIST dataset and $\sigma=0.005$ for CIFAR-10 dataset. 
We can see from  Figure \ref{fig:exp3-cifar} that both uncompressed FedAvg and $1$-SignFedAvg can benefit from multiple local SGD steps. More surprisingly, $1$-SignFedAvg can even outperform the uncompressed FedAvg. This is probably because the EMNIST dataset is less heterogeneous than the one we used in Section \ref{sec:exp2}. The results on the performance of $1$-SignFedAvg and $\infty$-SignFedAvg under various choices of noise scales are relegated to Appendix \ref{app:exp3}, which are also consistent with our theoretical claims in Section \ref{sec:algo}. 
\begin{figure}[htbp]
	\centering
	\begin{subfigure}[b]{0.27\textwidth}
	\centering
	\includegraphics[width=\textwidth]{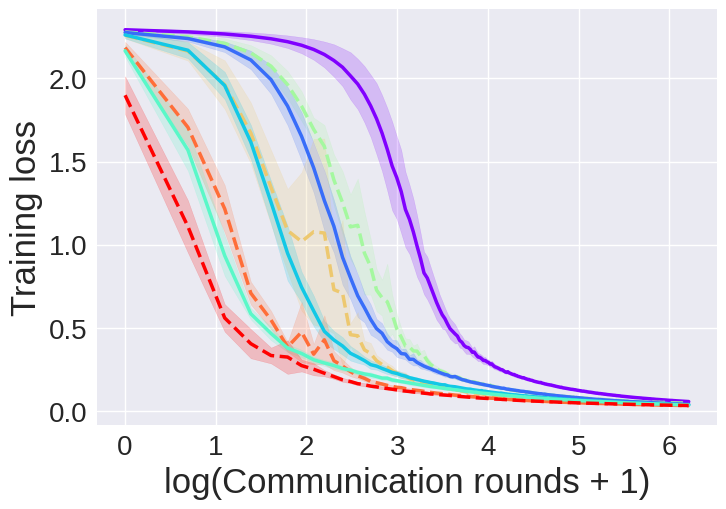}
	\caption{Training Loss}
	\label{}
\end{subfigure}
	\begin{subfigure}[b]{0.27\textwidth}
	\centering
	\includegraphics[width=\textwidth]{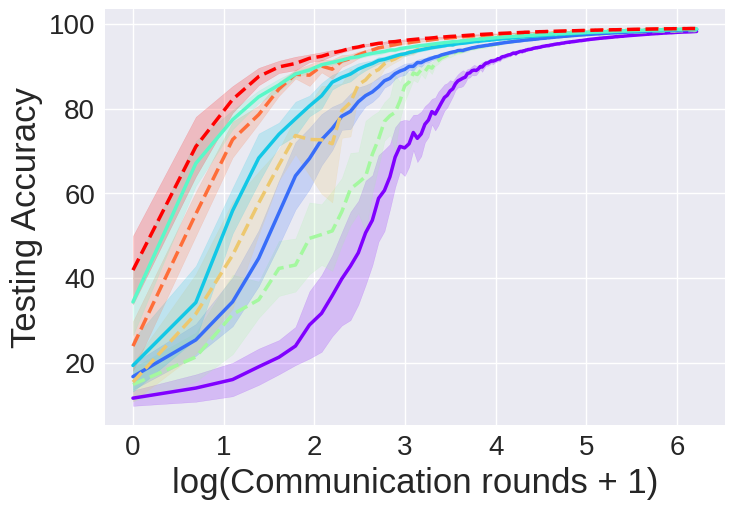}
	\caption{Test Accuracy}
	\label{}
\end{subfigure}
\begin{subfigure}[b]{0.27\textwidth}
	\centering
	\includegraphics[width=\textwidth]{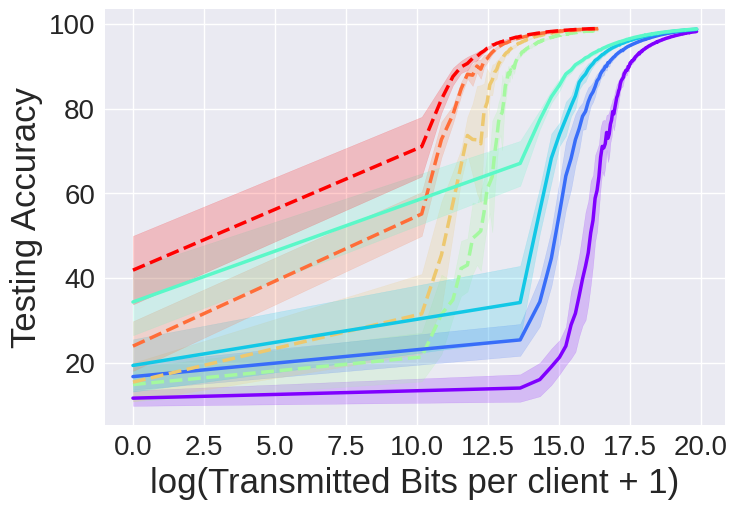}
	\caption{Test Accuracy w.r.t bits}
	\label{}
\end{subfigure}
\begin{subfigure}[][-50pt][b]{0.15\textwidth}
	\centering
	\includegraphics[width=\textwidth]{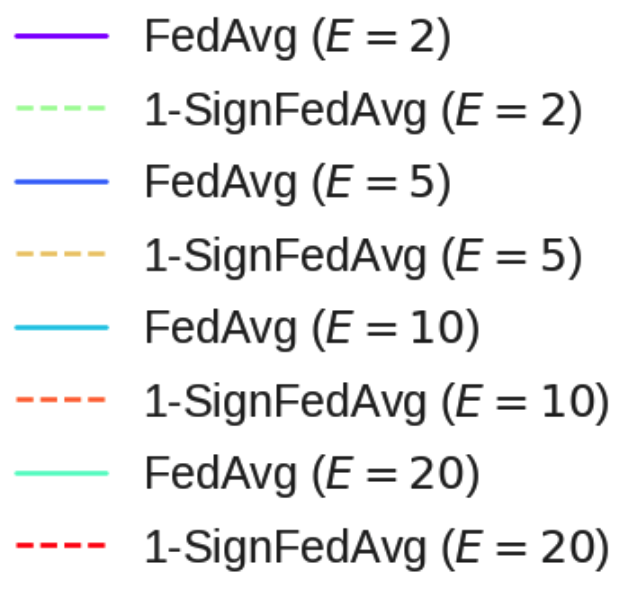}
\end{subfigure}
\vspace{-0.2cm}
\caption{Performance of FedAvg and  $1$-SignFedAvg on the EMNIST dataset. }
\label{fig:exp3-emnist}
\end{figure}

\subsection{Plateau Criterion for tuning the noise scale}\label{sec:plateau}
From previous experiments, we have learned that the noise scale $\sigma$ has to be properly chosen for the algorithm to perform well. However, it could be time-consuming to select the optimal noise scale via grid search. Therefore, here we introduce a simple yet useful strategy that can tune the noise scale adaptively during the training process. Figure \ref{fig:Sign-SGD-Simplen_noise} indicates that the noise scale should plays a similar role as the stepsize when training a neural network: Small noise scale leads to fast convergence at the beginning, while large noise scale guarantees a better final performance. This suggests that we should use an increasing noise scale during the optimization process.  We can also see this from Corollary \ref{col:thm1_informal} because that the noise scale $\sigma$ is proportional to $\tau$. Besides, it has been shown that the gradients of neural network tend to be sparser during the training process \cite{karimireddy2019error}. Therefore, as studied in \cite{isik2022learning}, from the rate-distortion theoretic aspect, the noise scale should be increasing as the compression becomes more aggressive. Motivated by all of these insights, we propose the following Plateau criterion for adapting the noise scale.

\textbf{Plateau criterion.} We denote a few parameters $\sigma_{\text{bound}}\geq\sigma_{\text{init}}>0$,  $\kappa\in \Zbb_+$, $\beta>0$. We first start Algorithm \ref{alg:SignFedAvg} with a small noise scale $\sigma_{\text{init}}$, i.e., $\sigma = \sigma_{\text{init}}$, and then update the noise scale via $\sigma = \beta \sigma$, where $\beta\in[1.5,2]$, whenever the objective function stops improving for $\kappa$ communication rounds. We stop updating $\sigma$ if it has already been greater than a relatively large number $\sigma_{\text{bound}}$. %Finally, we can set $\beta\in[1.5,2]$ that controls the trend of noise scale.

\begin{figure}[htbp]
	\centering
\begin{subfigure}[b]{0.27\textwidth}
	\centering
	\includegraphics[width=\textwidth]{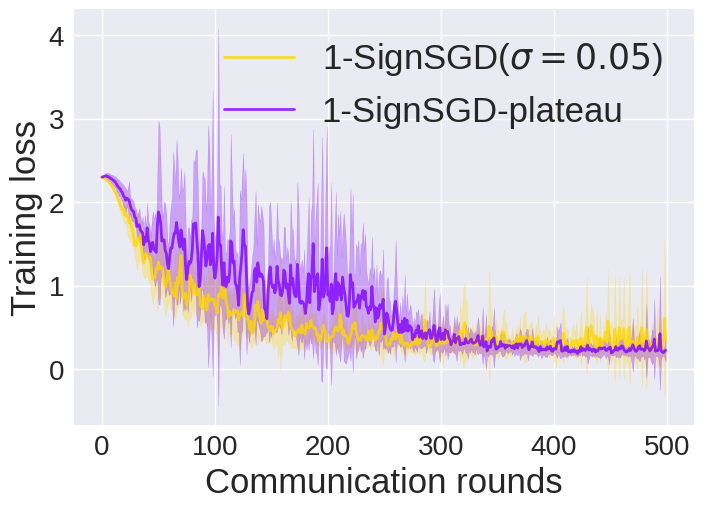}
	\caption{Non-i.i.d MNIST}
	\label{}
\end{subfigure}
	\begin{subfigure}[b]{0.27\textwidth}
	\centering
	\includegraphics[width=\textwidth]{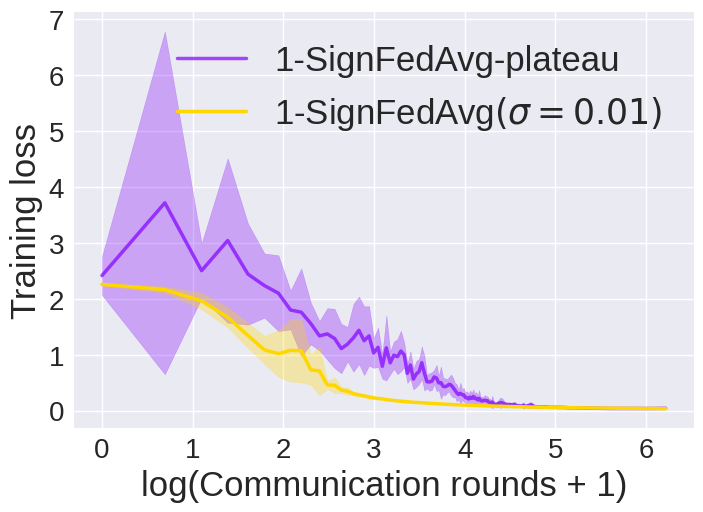}
	\caption{EMNIST}
	\label{}
\end{subfigure}
\begin{subfigure}[b]{0.28\textwidth}
	\centering
	\includegraphics[width=\textwidth]{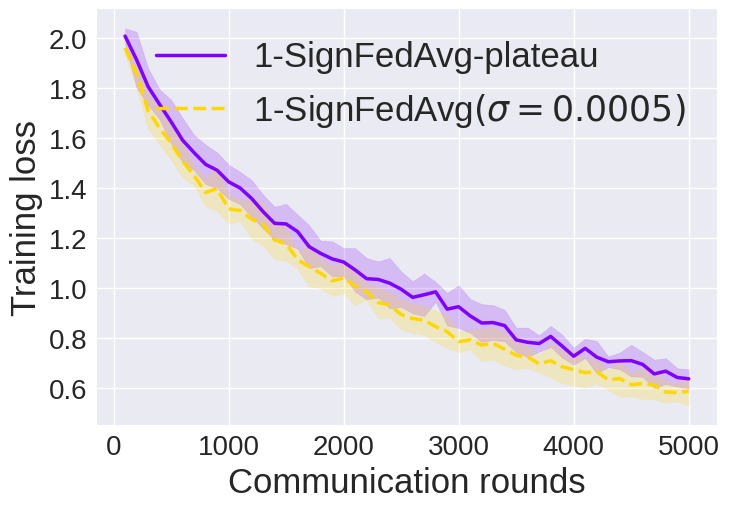}
	\caption{CIFAR-10}
	\label{}
\end{subfigure}

\caption{{ Evaluating the efficacy of Plateau criterion  on three different datasets.} }
\label{fig:exp-plateau}
\end{figure}

\textbf{Results.} We demonstrate the efficacy of the Plateau criterion by comparing the performance of $1$-SignSGD/$1$-SignFedAvg with the optimal noise scale found in previous experiments and the ones with Plateau criterion. Figure \ref{fig:exp-plateau} shows the results under the three different settings used in Section \ref{sec:exp2} and \ref{sec:exp3}. We can see that,  the Plateau criterion could results in a  slower convergence speed than the optimal noise scale in the middle phase of optimization, because it requires some time for the algorithm to adapt to a suitable noise scale. But eventually it can lead to the same objective value obtained by using the optimal noise scale. For more details like the hyperparameters for Plateau criterion and the evolution of noise scale, we refer readers to Appendix \ref{app:plateau}.

\section{Conclusion}
\label{sec:conclusion}
In this work, we have proposed the $z$-SignFedAvg: a FedAvg-type algorithm with the stochastic sign-based compression. Thanks to the novel noisy perturbation scheme in Section \ref{sec:sign}, the proposed $z$-SignFedAvg provides a unified viewpoint to the existing sign-based methods as well as a general framework for convergence rate analysis.
Through both theoretical analyses and empirical experiments, we have shown that the $z$-SignFedAvg can perform nearly the same, sometimes even better, than the uncompressed FedAvg and enjoy a significant reduction in the number of bits transmitted from clients to the server. As a final remark, the stochastic sign-based compression proposed in this work can be of independent interest and can be conveniently combined with other adaptive FL algorithms or gradient sparsification techniques such as those in \cite{karimireddy2020scaffold,reddi2020adaptive,basu2019qsparse}, to further improve the communication efficiency.

\nocite{*}

\bibliography{template}
\bibliographystyle{plain}

%%%%%%%%%%%%%%%%%%%%%%%%%%%%%%%%%%%%%%%%%%%%%%%%%%%%%%%%%%%%%%%%%%%%%%%%%%%%%%%
%%%%%%%%%%%%%%%%%%%%%%%%%%%%%%%%%%%%%%%%%%%%%%%%%%%%%%%%%%%%%%%%%%%%%%%%%%%%%%%
% APPENDIX
%%%%%%%%%%%%%%%%%%%%%%%%%%%%%%%%%%%%%%%%%%%%%%%%%%%%%%%%%%%%%%%%%%%%%%%%%%%%%%%
%%%%%%%%%%%%%%%%%%%%%%%%%%%%%%%%%%%%%%%%%%%%%%%%%%%%%%%%%%%%%%%%%%%%%%%%%%%%%%%
\newpage
\appendix
\section*{Appendix}
\renewcommand{\thesubsection}{\Alph{subsection}}
\subsection{Comparison with existing stochastic sign-based methods}
\label{app:discuss}

Table \ref{table:rw} summarizes the key features of a few representative stochastic sign-based methods and the proposed $z$-SignFedAvg, including the convergence rate, metric used in the convergence rate analysis, extra assumptions other than A.1-A.3 in Assumption \ref{asp:common}, and whether the algorithm can achieve linear speedup and allow multiple local SGD steps.

For communication complexity, we focus on the uplink communication cost, i.e., the number of bits transmitted from the clients to the server in each communication round. We assume that all the uncompressed algorithms use 32 bits to represent a single float number as it is the most common setting in Tensorflow \cite{abadi2016tensorflow} and Pytorch \cite{paszke2017automatic}. 

While most of the existing methods use the squared $\ell_2$-norm of gradients as the convergence metric, the work \cite{safaryan2021stochastic} adopts the $\ell_2$-norm of gradients. The work \cite{chen2020distributed} uses a convergence metric mixed with squared $\ell_2$-norm and $\ell_1$-norm of gradients due to the compression in both  uplink and downlink .  
%or some of the mixed criterion. 

%In this table, we review the algorithms in these works on the convergence rate along with the used convergence metrics, communication complexity, assumptions required, and also whether they allow local computation. Particularly, \cite{chen2020distributed,safaryan2021stochastic} adopt convergence metrics other than the commonly used average squared $\ell_2$ norm of gradients. Due to the additional step of downlink compression with majority vote, \cite{chen2020distributed} use a convergence metric mixed with $\ell_2$ norm  and $\ell_1$ norm, while \cite{safaryan2021stochastic} use $\ell_2$ norm instead of squared $\ell_2$ norm. 

%For communication complexity, we only compare the unlink communication, and report the used bits per communication. We assume that all the algorithms need 32 bits to represent a single float number as this is the most common setting in Tensorflow \cite{abadi2016tensorflow} and Pytorch \cite{paszke2017automatic}. 

\begin{table}[htbp]
	\small
	\centering
	\begin{tabular}{|c| c| c| c|c|c|} 
	 \hline
	 \makecell{\bfseries Algorithm} &\makecell{\bfseries Convergence\\\bfseries rate / metric }& \makecell{\bfseries Num. of bits per\\\bfseries commun. round} & \makecell{\bfseries Extra\\\bfseries Assumptions?} & \makecell{\bfseries Can achieve\\\bfseries linear speedup?}  & \makecell{\bfseries Can allow\\ \bfseries multiple \\ \bfseries local steps?} \\ [0.5ex] 
	 \hline\hline
	 
	 \makecell{\cite{SGD}}&  \makecell{$\Oc(\tau^{-\frac{1}{2}})$  \\  squared $\ell_2$} & $32d$ & No & {\checkmark}  & {\red \xmark} \\ \hline
	 \makecell{\cite{mcmahan2017communication}\\
	 \cite{yu2019parallel}}  & \makecell{ $\Oc(\tau^{-\frac{1}{2}})$ \\  squared $\ell_2$} & $32d$& \makecell[l]{
		\tabitem Bounded gradient
	 }&{\checkmark}  & {\checkmark} \\\hline
	 \makecell{
	 \cite{karimireddy2019error}}  &  \makecell{$\Oc(\tau^{-\frac{1}{2}}+ d^2\tau^{-1})$ \\  squared $\ell_2$} & $d+32$& \makecell[l]{\tabitem Bounded gradient}&{\red \xmark} & {\red \xmark}\\\hline
	 \makecell{
	 \cite{safaryan2021stochastic}}  &  \makecell{$\Oc(\tau^{-\frac{1}{4}})$\\ $\ell_2$}&$d$& No &{\red \xmark} & {\red \xmark} \\ [1ex]
	 \hline
	 \makecell{
	 \cite{jin2020stochastic}}  &  \makecell{$\Oc(\tau^{-\frac{1}{4}})$ \\ squared $\ell_2$}&$d$& \makecell[l]{\tabitem Bounded gradient\\ \tabitem  n is an odd number} &{\red \xmark} & {\red \xmark} \\ [1ex]
	 \hline
	 \makecell{
	 \cite{chen2020distributed}}  & \makecell{$\Oc(\tau^{-\frac{1}{4}})$  \\ mixed }&$d$ & \makecell[l]{\tabitem Bounded gradient\\ \tabitem  n is an odd number}& {\red \xmark}& {\red \xmark}\\ \hline
	 	 \makecell{
	{\cite{alistarh2017qsgd}}}  & \makecell{{$\Oc(\tau^{-\frac{1}{2}})$}  \\ {squared $\ell_2$}} & {$\approx sd + 32$} & {No} & {\checkmark}& {\red \xmark}\\ \hline
	 	 \makecell{
	 {\cite{haddadpour2021federated}}} & \makecell{{$\Oc(\tau^{-\frac{1}{2}})$}  \\ {squared $\ell_2$} }&$ {\approx sd + 32}$ & \makecell[l]{\tabitem {Bounded gradient}\\ \ \ \ \ \ {dissimilarity}}& {\checkmark}& {\checkmark}\\ \hline
	%  \makecell{N-MedianSGD\\
	%  \cite{chen2020distributed}}  & $\Oc(\tau^{-\frac{1}{3}})$&$32d$& \makecell[l]{\tabitem Bounded gradient\\ \tabitem  n is an odd number}&{\red \xmark} & {\red \xmark} \\ [1ex] 
	 \hline
	 \specialcell{$1$-SignFedAvg (ALG. \ref{alg:SignFedAvg})\\
	 %G-SignSGDwM (ALG. \ref{alg:Sign-SGD-wM})\\
	 \bfseries This work } &\makecell{$\Oc(\tau^{-\frac{1}{3}})$  \\ squared $\ell_2$ }&$d$  & \makecell[l]{\tabitem Bounded gradient\\ \tabitem Bounded   $6$th moment\\ \ \ \ \ of gradient noise } & {\checkmark} & {\checkmark}\\ \hline
	 \specialcell{$\infty$-SignFedAvg (ALG. \ref{alg:SignFedAvg})\\
	 \bfseries This work} & \makecell{$\Oc(\tau^{-\frac{1}{2}})$ \\ squared $\ell_2$ }&$d$  & \makecell[l]{\tabitem Bounded gradient\\ \tabitem Bounded support \\ \ \ \ \ of  gradient noise} & {\checkmark} & {\checkmark}\\ \hline
	\end{tabular}
	\caption{Summary of representative stochastic sign-based methods.}
	\label{table:rw}
	\end{table}
	
	Among the works in Table \ref{table:rw}, the setting considered by \cite{safaryan2021stochastic} is closest to ours. \cite{safaryan2021stochastic} proposed an algorithm that can achieve the convergence rate $\Oc(\tau^{-\frac{1}{4}})$ with the $\ell_2$-norm of gradients as the metric. We remark that this is inferior to the convergence rate $\Oc(\tau^{-\frac{1}{2}})$ with the squared $\ell_2$-norm as the metric. To illustrate this point, we denote a series of vector as $\{\alpha_1,...,\alpha_\tau,...\}$ with $\alpha_i\in\Rbb^d$. If now
    \begin{align}
        \frac{1}{\tau}\sum_{i=1}^\tau \|\alpha_i\| = \Oc(\tau^{-\frac{1}{4}}),
    \end{align}in the worst case, we can only guarantee that
    \begin{align}
    \label{p:less}
        \frac{1}{\tau}\sum_{i=1}^\tau \|\alpha_i\|^2 \leq \tau \left(\frac{1}{\tau}\sum_{i=1}^\tau \|\alpha_i\|\right)^2= \Oc(\tau^{\frac{1}{2}}).
    \end{align}  
    As a simple example, the equality in \eqref{p:less} holds if and only if there is exactly one non-zero term in $\{\alpha_1,...,\alpha_\tau\}$.
    
    On the contrary, if it holds that \begin{align}
        \frac{1}{\tau}\sum_{i=1}^\tau \|\alpha_i\|^2 = \Oc(\tau^{-\frac{1}{2}}),
    \end{align} then we have 
    \begin{align}
        \frac{1}{\tau}\sum_{i=1}^\tau \|\alpha_i\| \leq \sqrt{\frac{1}{\tau}\sum_{i=1}^\tau \|\alpha_i\|^2}= \Oc(\tau^{-\frac{1}{4}}).
    \end{align}
    Thus, the convergence results in \cite{safaryan2021stochastic} cannot imply the rate in Theorem \ref{thm:sign-fedavg_inf_informal}.
   Besides, the algorithm in \cite{safaryan2021stochastic} is equivalent to our Algorithm \ref{alg:SignFedAvg} with $z=\infty$, $E=1$ and $\sigma=\|g_{t-1,s}^i\|$.
   This input-dependent noise scale is linearly increasing w.r.t the problem dimension and is too conservative for practical applications. From Figure \ref{fig:Sign-SGD-Simple} and Figure \ref{fig:exp2}, we have already seen that this input-dependent noise scale could result in an extremely slow convergence for high-dimensional problems.

   {
   Except for the previous sign-based compression methods, 
   another type of compressed FL algorithms, such as \cite{alistarh2017qsgd} and \cite{haddadpour2021federated}, adopt a unified unbiased compressor $Q(\cdot)$ that satisfies $\Ebb[\|Q(x)-x\|^2]\leq C\|x\|^2$ for some constant $C>0$. We remark that such property is not fulfilled by any of the existing sign-based compressors. Thus, the theoretical results therein cannot be applied to sign-based methods.    
   A specific example of such unbiased compressor is described below.
   
   \begin{definition}[Unbiased quantizer]
    For any variable $x\in \Rbb^d$, the unbiased quantizer $Q(\cdot):\Rbb^d\to \Rbb^d$ is defined as below
    \begin{align}
    \label{p:quantizer}
        Q(x) = \|x\|_2 \cdot \begin{bmatrix}
\rm{Sign}(x_1)\xi(x_1,s)  \\
\rm{Sign}(x_2)\xi(x_2,s)  \\
\vdots\\
\rm{Sign}(x_d)\xi(x_d,s)  
\end{bmatrix}
    \end{align}

where $\xi(x_i,s)$ is a random variable taking on value
$\frac{l+1}{s}$ with probability $\frac{|x_i|}{\|x\|_2} s -l$
and $\frac{l}{s}$otherwise. Here,
the tuning parameter $s$ corresponds to the number of
quantization levels and $l \in [0, s)$ is an integer such that
$\frac{|x_i|}{\|x\|_2} \in [l
/s, l+1/s).$
\end{definition}

In Table \ref{table:rw}, we assume both \cite{alistarh2017qsgd} and \cite{haddadpour2021federated} adopt the quantizer in \eqref{p:quantizer}. Generally speaking, this type of unbiased quantization usually requires much more bits than sign-based compression to obtain a good performance, which is also verified empirically in Appendix \ref{app:QSGD}. It is also worthwhile to mention that the FedPAQ in \cite{reisizadeh2020fedpaq} and the FedCOM in \cite{haddadpour2021federated} are equivalent in algorithm, but only the latter one considers the heterogeneous scenario theoretically.

    }

 \subsection{Detailed theoretical results}   
 \label{app:formal} 
 We first state the result on the limit of $z$-distribution.
 
\begin{lemma}
	\label{lm:converge2uni}
	The $z$-distribution weakly converges to uniform distribution on $[-1,1]$ when $z\to+\infty$.
\end{lemma}

The following corollary is the formal version of Corollary \ref{col:thm1_informal}.
\begin{corollary}[Formal version of Corollary \ref{col:thm1_informal}]
	\label{col:thm1}
	For $\gamma=\min\{n^{\frac{z}{2z+1}}\tau^{-\frac{z+1}{2z+1}},\frac{1}{L_{\max}}\}$ and $\sigma=(n\tau)^{\frac{1}{4z+2}}$ in Theorem \ref{thm:sign-fedavg_z}, we have
	\begin{align}
		\Ebb\left[\frac{1}{\tau}\sum_{t=1}^{T}\sum_{s=1}^E \|\nabla f(\bar x_{t-1,s-1})\|^2\right] \leq & {\frac{2\Ebb[f(x_0) - f^*]}{(n\tau)^{\frac{z}{2z+1}}} + \frac{\zeta^2L_{\max}}{(n\tau)^{\frac{z+1}{2z+1}}} + \frac{4(E-1)En^{\frac{2z}{2z+1}} L_{\max}^2\left(\zeta^2+G^2\right)}{3\tau^{\frac{2z+2}{2z+1}}}} \notag \\
		& +{ \frac{ 2^{2z+1} E^{2z}\sqrt{Q_z+G^{4z+2}}G}{\sqrt 2(2z+1)(n\tau)^{\frac{z}{2z+1}}} + \frac{ 2^{4z} E^{4z+1}(Q_z+G^{4z+2})L_{\max}}{2(2z+1)^2n^{\frac{z}{2z+1}}\tau^{\frac{3z+1}{2z+1}}}} \notag \\ 
		&+{\frac{4\eta_z^2\sum_{j=1}^d L_j}{E(n\tau)^{\frac{z}{2z+1}}}}.
	\end{align}
	Furthermore, if  $E\leq n^{-\frac{3z}{4z+2}}\tau^{\frac{z+2}{4z+2}}$, the upper bound above becomes
	\begin{align}
		\Ebb\left[\frac{1}{\tau}\sum_{t=1}^{T}\sum_{s=1}^E \|\nabla f(\bar x_{t-1,s-1})\|^2\right] \leq & {\frac{2\Ebb[f(x_0) - f^*]}{(n\tau)^{\frac{z}{2z+1}}} + \frac{\zeta^2L_{\max}}{(n\tau)^{\frac{z+1}{2z+1}}} + \frac{4 L_{\max}^2\left(\zeta^2+G^2\right)}{3(n\tau)^{\frac{z}{2z+1}}}} \notag \\
		& +{ \frac{ 2^{2z+1} E^{2z}\sqrt{Q_z+G^{4z+2}}G}{\sqrt 2(2z+1)(n\tau)^{\frac{z}{2z+1}}} + \frac{ 2^{4z} E^{4z+1}(Q_z+G^{4z+2})L_{\max}}{2(2z+1)^2n^{\frac{z}{2z+1}}\tau^{\frac{3z+1}{2z+1}}}} \notag \\ 
		&+{\frac{4\eta_z^2\sum_{j=1}^d L_j}{E(n\tau)^{\frac{z}{2z+1}}}}.
	\end{align}.
% 	\begin{align}
% 		\Ebb[\frac{1}{\tau}\sum_{t=1}^{T}\sum_{s=1}^E] \|\nabla f(\bar x_{t-1,s-1})\|^2 &\leq {\frac{2\Ebb[f(x_0) - f^*]}{(n\tau)^{\frac{z}{2z+1}}} + \frac{\zeta^2L_{\max}}{(n\tau)^{\frac{z+1}{2z+1}}} + \frac{2 L_{\max}^2G^2}{3(n\tau)^{\frac{z}{2z+1}}}}\\
% 		& +{ \frac{ 2^{2z} E^{2z}\sqrt{Q_z+G^{4z+2}}G}{(2z+1)(n\tau)^{\frac{z}{2z+1}}} + \frac{ 2^{4z} E^{4z+1}(Q_z+G^{4z+2})L_{\max}}{2(2z+1)^2n^{\frac{z}{2z+1}}\tau^{\frac{3z+1}{2z+1}}}}\\ 
% 		&+{\frac{2^{\frac{1}{z}}(\Gamma(\frac{1}{2z}))^2\sum_{j=1}^d L_j}{z^2E(n\tau)^{\frac{z}{2z+1}}}},
% 	\end{align} 
\end{corollary}

The formal version of Theorem  \ref{thm:sign-fedavg_inf_informal} is given below.

\begin{theorem}[Formal version of Theorem \ref{thm:sign-fedavg_inf_informal}]
	\label{thm:sign-fedavg_inf}
Suppose that Assumption \ref{asp:common} and  \ref{asp:inf} hold. For   $\gamma \leq \frac{1}{L_{\max}}$, $\eta=\sigma$, $z=+\infty$ and $\sigma>E(G+Q_\infty)$ in Algorithm \ref{alg:SignFedAvg}, we have
	\begin{align}
		\Ebb\left[\frac{1}{TE}\sum_{t=1}^{T}\sum_{s=1}^E \|\nabla f(\bar x_{t-1,s-1})\|^2\right] \leq & \underbrace{\frac{2\Ebb[f(x_0) - f^*]}{TE\gamma} + \frac{\gamma\zeta^2L_{\max}}{n} + \frac{4\gamma^2 (E-1)E L_{\max}^2(\zeta^2+G^2)}{3}}_{\text{Standard terms in FedAvg}} \notag \\ & +\underbrace{\frac{4\gamma\sigma^2\sum_{j=1}^d L_j }{En}}_{\text{Variance term}}.
	\end{align}
	Otherwise, if $\sigma\leq E(G+Q_\infty)$, there exists a problem instance for which the algorithm cannot converge. If we further choose $\gamma=\min\{n^{\frac{1}{2}}\tau^{-\frac{1}{2}},\frac{1}{L_{\max}}\}$, we have
	\begin{align}
	\label{p:18}
		\Ebb\left[\frac{1}{\tau}\sum_{t=1}^{T}\sum_{s=1}^E \|\nabla f(\bar x_{t-1,s-1})\|^2\right] \leq & {\frac{2\Ebb[f(x_0) - f^*]}{(n\tau)^{\frac{1}{2}}} + \frac{\zeta^2L_{\max}}{(n\tau)^{\frac{1}{2}}} + \frac{4(E-1)E n L_{\max}^2\left(\zeta^2+G^2\right)}{3\tau}} \notag  \\ & +{\frac{4\sigma^2\sum_{j=1}^d L_j}{E(n\tau)^{\frac{1}{2}}}}.
	\end{align}
	Furthermore, if $E\leq n^{-\frac{3}{4}}\tau^{\frac{1}{4}}$, the upper bound above becomes
	\begin{align}
		\Ebb\left[\frac{1}{\tau}\sum_{t=1}^{T}\sum_{s=1}^E \|\nabla f(\bar x_{t-1,s-1})\|^2\right] \leq & {\frac{2\Ebb[f(x_0) - f^*]}{(n\tau)^{\frac{1}{2}}} + \frac{\zeta^2L_{\max}}{(n\tau)^{\frac{1}{2}}} + \frac{4 L_{\max}^2\left(\zeta^2+G^2\right)}{3(n\tau)^{\frac{1}{2}}}} \notag \\ & +{\frac{4\sigma^2\sum_{j=1}^d L_j}{E(n\tau)^{\frac{1}{2}}}},
	\end{align}which recovers the convergence result of the uncompressed FedAvg algorithm \cite{yu2019parallel}.
\end{theorem}
% As we can see, the bound in Theorem \ref{thm:sign-fedavg_inf} is almost the same as the algorithm \cite{yu2019parallel} without any compression.
In particular, since the third term in the RHS of \eqref{p:18} is $\Oc(E^2n\tau^{-1})$, hence when $E\leq n^{-\frac{3}{4}}\tau^{\frac{1}{4}}$, this term becomes $\Oc((n\tau)^{-\frac 12})$.

\subsection{Proofs}
\label{app:proof}

\subsubsection{Proof of Lemma \ref{lm:phi_lm}}

We first state a useful inequality on the c.d.f of the $z$-distribution:
 \begin{lemma}
\label{lm:phi} For any $x\in \Rbb$, it holds that
	\begin{align}
		|x|- \frac{|x|^{2z+1}}{2(2z+1)}	\leq |\Psi_z(x)|\leq |x| ,\end{align} where $$\Psi_z(x) \stackrel{\rm{def.}}{=} \int_0^x e^{-\frac{t^{2z}}{2}}dt.$$
	
 \end{lemma}
 
 Similar to the sign operator, for any vector  $x=[x(1),...,x(d)]^\top\in \Rbb^d$, we define $$\Psi_z\left({x}\right) = [\Psi_z(x(1)),...,\Psi_z(x(d))]^\top.$$ 
 
 With the presence of Lemma \ref{lm:phi}, we have
 \begin{align}
			\left\|\eta_z\sigma\Ebb\left[\rm{Sign}(x+\sigma \xi_z)\right]-x\right\|^2&= \left\|x-\sigma\Psi_z\left(\frac{x}{\sigma}\right)\right\|^2  =\sum_{j=1}^d \left(x(j)-\sigma\Psi_z\left(\frac{x(j)}{\sigma}\right)\right)^2 \notag \\ \label{p:col_phi_2} & \leq \sum_{j=1}^d \frac{\left(x(j)\right)^{4z+2}}{4(2z+1)^2\sigma^{4z}} = 	\frac{\|x\|_{4z+2}^{4z+2}}{4(2z+1)^2\sigma^{4z}}.
		\end{align}

\begin{proof}[Proof of Lemma \ref{lm:phi}]
	Without loss of generality, we consider $x\geq 0$.
	First, 
	\begin{align}
		\int_{0}^x e^{-\frac{t^{2z}}{2}}dt\leq \int_{0}^x 1dt\leq x.
	\end{align}

% 	The Taylor's expansion of $e^x$ is $1+x+\frac{x^2}{2}+\sum_{k=3}^{+\infty}\frac{x^k}{k!}$. Hence we have 
% 	\begin{align}
% 		\int_{0}^x e^{-\frac{t^{2z}}{2}}dt &= \int_{0}^x \left(1-\frac{t^{2z}}{2}+\frac{t^{4z}}{8}+\sum_{k=3}^{+\infty}\frac{(-1)^{k}t^{2zk}}{2^k k!}\right) dt\\
% 		&=x-\frac{x^{2z+1}}{2(2z+1)}+\frac{x^{4z+1}}{8(4z+1)}+\sum_{k=3}^{+\infty}\frac{(-1)^{k}x^{2zk+1}}{2^k k!(2zk+1)}.
% 	\end{align}

	Now we define $F(x) \stackrel{\text{def.}}= \int_{0}^x e^{-\frac{t^{2z}}{2}}dt - x + \frac{x^{2z+1}}{2(2z+1)}$. Note that $F(0)=0$.
	Then, it suffices to show $F(x)\geq 0 $ by
	\begin{align}
	\label{p:lm4_u1}
	    F'(x) = e^{-\frac{x^{2z}}{2}}-1+\frac{x^{2z}}{2}\geq 0.
	\end{align}
	It is true since the inequality $e^{-t}-1+t\geq 0 $ for any $t\geq0$.
\end{proof}

\subsubsection{Proof of Lemma \ref{lm:converge2uni}}

	Now we denote the p.d.f of the uniform distribution as \begin{align}
		p_{\infty}(x) = 
	\begin{cases}
	  \frac{1}{2} & |x|\leq 1, \\
	  0 & |x|>1.
	\end{cases}
	\end{align}

	Without loss of generality, for any $x>1$ and $z\in\Zbb_+$, we have 
	\begin{align}
		\left\lvert\int_{-\infty}^{x}\frac{1}{2\eta_z}e^{-\frac{t^{2z}}{2}}dt - \int_{-\infty}^{x}p_{\infty}(t)dt\right\rvert 
		&=\left\lvert\int_{0}^{x}\left(\frac{1}{2\eta_z}e^{-\frac{t^{2z}}{2}}-p_{\infty}(t)\right)dt\right\rvert \notag \\
		&\leq \int_{0}^{1}\left\lvert\frac{1}{2\eta_z}e^{-\frac{t^{2z}}{2}}-\frac{1}{2}\right\rvert dt+\int_{1}^{x}\frac{1}{2\eta_z}e^{-\frac{t^{2z}}{2}}dt.
	\end{align}
	
	For any $0<\epsilon<\min\{1,x-1\}$, we have
	\begin{align}
		\int_{0}^{1}\left\lvert\frac{1}{2\eta_z}e^{-\frac{t^{2z}}{2}}-\frac{1}{2}\right\rvert dt &=\int_{0}^{1-\epsilon}\left\lvert\frac{1}{2\eta_z}e^{-\frac{t^{2z}}{2}}-\frac{1}{2}\right\rvert dt + \int_{1-\epsilon}^{1}\left\lvert\frac{1}{2\eta_z}e^{-\frac{t^{2z}}{2}}-\frac{1}{2}\right\rvert dt \notag \\
		&\leq \left\lvert\frac{1}{2\eta_z}e^{-\frac{(1-\epsilon)^{2z}}{2}}-\frac{1}{2}\right\rvert+\epsilon.
	\end{align}

	Since $\lim_{z\to\infty}\frac{1}{2\eta_z}=\lim_{z\to\infty}\frac{z}{2^\frac{1}{2z}\Gamma (\frac{1}{2z})}=\frac{1}{2}$ and $\lim_{z\to\infty}e^{-\frac{(1-\epsilon)^{2z}}{2}}=1$, there exists an integer $Z_1>0$ such that if $z>Z_1$, we have $$\left\lvert\frac{1}{2\eta_z}e^{-\frac{(1-\epsilon)^{2z}}{2}}-\frac{1}{2}\right\rvert\leq \epsilon.$$

	Similarly, we have 
	\begin{align}
		\int_{1}^{x}\frac{1}{2\eta_z}e^{-\frac{t^{2z}}{2}}dt &= \int_{1}^{1+\epsilon}\frac{1}{2\eta_z}e^{-\frac{t^{2z}}{2}}dt+\int_{1+\epsilon}^{x}\frac{1}{2\eta_z}e^{-\frac{t^{2z}}{2}}dt \notag \\
		&\leq \epsilon + \frac{1}{2\eta_z}e^{-\frac{(1+\epsilon)^{2z}}{2}}(x-1-\epsilon).
	\end{align}
	Since $\lim_{z\to\infty} e^{-\frac{(1+\epsilon)^{2z}}{2}}=0$, there exists an integer $Z_2>0$ such that if $z>Z_2$, we have 
	\begin{align}
		\int_{1}^{x}\frac{1}{2\eta_z}e^{-\frac{t^{2z}}{2}}dt\leq \epsilon.
	\end{align}

	In all, for any $0<\epsilon<\min\{1,x-1\}$, if $z$ is sufficiently large, we have 
	\begin{align}
		\left\lvert\int_{-\infty}^{x}\frac{1}{2\eta_z}e^{-\frac{t^{2z}}{2}}dt - \int_{-\infty}^{x}p_{\infty}(t)dt\right\rvert\leq 4\epsilon.
	\end{align}
	Taking $\epsilon\to 0$ and $z\to\infty$, we have 
	\begin{align}
		\lim_{z\to\infty} \left\lvert\int_{-\infty}^{x}\frac{1}{2\eta_z}e^{-\frac{t^{2z}}{2}}dt - \int_{-\infty}^{x}p_{\infty}(t)dt\right\rvert = 0.
	\end{align}

\subsubsection{Proof of Theorem \ref{thm:sign-fedavg_z}}

	We denote the  aggregated update $\bar x_{t} = \bar x_{t-1,E}.$
First, we state two technical lemmas:
\begin{lemma} Suppose that Assumption \ref{asp:common} and \ref{asp:z-moment} hold. For the $t$-th ($1\leq t\leq T$) communication round in Algorithm \ref{alg:SignFedAvg}, if
  $\eta = \eta_z\sigma$ and $z<+\infty$, we have
	\label{lm:thm2_lm1}
	\begin{align}
		\Ebb[f(x_t) - f(\bar x_t)] \leq & \frac{\gamma 2^{2z} E^{2z+1}\sqrt{Q_z+G^{4z+2}}G}{\sqrt 2(2z+1)\sigma^{2z}} + \frac{\gamma^2 2^{4z} E^{4z+2}(Q_z+G^{4z+2})L_{\max}}{4(2z+1)^2\sigma^{4z}} \notag \\ & +\frac{2\eta_z^2\gamma^2 \sigma^2\sum_{j=1}^d L_j}{n}.
	\end{align}
\end{lemma}

\begin{lemma} Suppose that Assumption \ref{asp:common} hold. For the $t$-th ($1\leq t\leq T$) communication round in Algorithm \ref{alg:SignFedAvg}, if $\gamma\leq \frac{1}{L_{\max}}$, we have
	\label{lm:thm2_lm2}
	\begin{align}
		\Ebb[f(\bar x_t)-f( x_{t-1})]\leq -\frac{\gamma}{2}\sum_{s=1}^E \|\nabla f(\bar x_{t-1,s-1})\|^2 + \frac{E\gamma^2\zeta^2L_{\max}}{2n} + \frac{2\gamma^3 (E-1)E^2 L_{\max}^2(\zeta^2+G^2)}{3}.
	\end{align}
\end{lemma}

By combining Lemma \ref{lm:thm2_lm1} and Lemma \ref{lm:thm2_lm2}, we have

\begin{align}\label{p:thm1_ineq}
	\Ebb[f(x_t) - f( x_{t-1})] &= \Ebb[f(x_t) - f(\bar x_t)]+E[f(\bar x_t)-f( x_{t-1})] \notag \\
	\leq & -\frac{\gamma}{2}\sum_{s=1}^E \|\nabla f(\bar x_{t-1,s-1})\|^2 + \frac{E\gamma^2\zeta^2L_{\max}}{2n} + \frac{2\gamma^3 (E-1)E^2 L_{\max}^2(\zeta^2+G^2)}{3} \notag \\
	& +\frac{\gamma 2^{2z} E^{2z+1}\sqrt{Q_z+G^{4z+2}}G}{\sqrt 2(2z+1)\sigma^{2z}} + \frac{\gamma^2 2^{4z} E^{4z+2}(Q_z+G^{4z+2})L_{\max}}{4(2z+1)^2\sigma^{4z}} \notag \\ & +\frac{2\eta_z^2\gamma^2 \sigma^2\sum_{j=1}^d L_j}{n}.
\end{align}

Rearranging the inequality \eqref{p:thm1_ineq}, we have
\begin{align}
	\frac{1}{E}\sum_{s=1}^E \|\nabla f(\bar x_{t-1,s-1})\|^2 \leq & \frac{2\Ebb[ f( x_{t-1})-f(x_t) ]}{E\gamma} + \frac{\gamma\zeta^2L_{\max}}{n} + \frac{4\gamma^2 (E-1)E L_{\max}^2(\zeta^2+G^2)}{3} \notag \\
	& + \frac{ 2^{2z+1} E^{2z}\sqrt{Q_z+G^{4z+2}}G}{\sqrt 2(2z+1)\sigma^{2z}} + \frac{\gamma 2^{4z} E^{4z+1}(Q_z+G^{4z+2})L_{\max}}{2(2z+1)^2\sigma^{4z}} \notag \\ 
	&+\frac{4\eta_z^2\gamma \sigma^2\sum_{j=1}^d L_j}{En}.
\end{align}

Finally, by a telescopic sum, we obtain
\begin{align}
	\Ebb\left[\frac{1}{TE}\sum_{t=1}^{T}\sum_{s=1}^E \|\nabla f(\bar x_{t-1,s-1})\|^2 \right] \leq & \frac{2\Ebb[f(x_0) - f^*]}{TE\gamma} + \frac{\gamma\zeta^2L_{\max}}{n} + \frac{4\gamma^2 (E-1)E L_{\max}^2(\zeta^2+G^2)}{3} \notag \\
	& + \frac{ 2^{2z+1} E^{2z}\sqrt{Q_z+G^{4z+2}}G}{\sqrt 2(2z+1)\sigma^{2z}} + \frac{\gamma 2^{4z} E^{4z+1}(Q_z+G^{4z+2})L_{\max}}{2(2z+1)^2\sigma^{4z}} \notag \\ 
	&+\frac{4\eta_z^2\gamma \sigma^2\sum_{j=1}^d L_j}{En}.
\end{align}

\begin{proof}[Proof of Lemma \ref{lm:thm2_lm1}]
% We first derive an accurate characterization of \eqref{p:asp_unbiased}  using Lemma \ref{lm:phi}.

% For any $x\in \Rbb^d$, we denote that $x=[x_1,...,x_d]^\top$, then we have,
% 	\begin{align}
% 		\left\|\sigma\Phi\left(\frac{x}{\sigma} \right)-  x\right\|^2
% 		= \sum_{j=1}^d \left(\sigma \Phi(\frac{x_j}{\sigma})-x_j\right)^2\leq \sum_{j=1}^d \left(\sigma \left(\frac{x_j}{\sigma}-\frac{x_j^3}{6\sigma^3}\right)-x_j\right)^2=\frac{\sum_{j=1}^d x_j^6}{36\sigma^4}=\frac{\|x\|_6^6}{36\sigma^4}.
% 	\end{align}
% 	Hence,
% 	\begin{align}
% 		\left\|\sigma\Phi\left(\frac{x}{\sigma} \right)-  x\right\|
% 		\leq\frac{\|x\|_6^3}{6\sigma^2}.
% 	\end{align} 

 First, we  know from function smoothness that
\begin{align}\label{p:lm4_term1}
	f(x_t) - f(\bar x_t)
	&\leq \langle \nabla f(\bar x_t), x_t - \bar x_t \rangle + \frac{ \sum_{j=1}^d L_j \left(x_t(j) - \bar x_t(j) \right)^2}{2}.
\end{align}

As can be seen from \eqref{p:lm4_term1}, we need to study the $x_t-\bar x_t$ in order to obtaining the upper bound for $f(x_t)-f(\bar x_t)$. Note that
\begin{align}
    x_t - \bar x_t &= \frac{\gamma }{n}\sum_{i=1}^n \left(\eta_z\sigma \text{Sign}\left(\sum_{s=1}^E g_{t,s}^i+\sigma \xi_z\right)	- \sum_{s=1}^E g_{t,s}^i\right).
\end{align}

For ease of presentation, we define that
\begin{align}
\label{p:def_Ac}
    \mathcal{A}_t^i \stackrel{\text{def.}}=  \eta_z\sigma \text{Sign}\left(\sum_{s=1}^E g_{t,s}^i+\sigma \xi_z\right).
\end{align}

By taking the expectation over the random vector $\xi_z$, 
% \begin{align}
% 	\Ebb_{\xi_z}[x_t - \bar x_t] &= \frac{\gamma }{n}\sum_{i=1}^n \left( \sigma\Psi_z\left(\frac{1}{\sigma}\sum_{s=1}^E g_{t,s}^i\right)	- \sum_{s=1}^E g_{t,s}^i\right).
% \end{align}
for any $j=1,...,d$, we have
\begin{subequations}\label{p:lm4:2}
\begin{align}
	\Ebb_{\xi_z}[\left(x_t(j) - \bar x_t(j) \right)^2]= & \frac{\gamma^2 }{n^2} \Ebb_{\xi_z}\left[\left(\sum_{i=1}^n \left(\mathcal{A}_t^i	- \sum_{s=1}^E g_{t,s}^i(j)\right)\right)^2\right]\\
	= &\frac{\gamma^2 }{n^2} \Ebb_{\xi_z}\left[\left(\sum_{i=1}^n \left(\mathcal{A}_t^i(j)	-\Ebb_{\xi_z}\left[\mathcal{A}_t^i(j)\right]+\Ebb_{\xi_z}\left[\mathcal{A}_t^i(j)\right]-\sum_{s=1}^E g_{t,s}^i(j)\right)\right)^2\right]\\
	\leq & \frac{\gamma^2 }{n^2} \Ebb_{\xi_z}\left[\left(\sum_{i=1}^n \left(\mathcal{A}_t^i(j)	-\Ebb_{\xi_z}\left[\mathcal{A}_t^i(j)\right]\right)\right)^2\right]\label{p:lm4:2_term1}\\ & + \frac{\gamma^2 }{n^2} \Ebb_{\xi_z}\left[\left(\sum_{i=1}^n \left(\Ebb_{\xi_z}\left[\mathcal{A}_t^i(j)\right]-\sum_{s=1}^E g_{t,s}^i(j)\right)\right)^2\right],
\end{align}
\end{subequations} where the last inequality is obtained because $\sum_{i=1}^n \left(\mathcal{A}_t^i(j)	-\Ebb_{\xi_z}\left[\mathcal{A}_t^i(j)\right]\right)$ is zero-mean and independent of $\sum_{i=1}^n \left(\Ebb_{\xi_z}\left[\mathcal{A}_t^i(j)\right]-\sum_{s=1}^E g_{t,s}^i(j)\right)$.

From \eqref{p:def_Ac} it is easy to check that 
$|\mathcal{A}_t^n(j)| \leq \eta_z^2\sigma^2.$ Hence, for the RHS of \eqref{p:lm4:2_term1}, we have 
\begin{align}\label{p:lm4_abc}
 \Ebb_{\xi_z}\left[\left(\sum_{i=1}^n \left(\mathcal{A}_t^i(j)	-\Ebb_{\xi_z}\left[\mathcal{A}_t^i(j)\right]\right)\right)^2\right] 
 &\overset{(a)}{=} \sum_{i=1}^n \Ebb_{\xi_z}\left[\left(\mathcal{A}_t^i(j)	-\Ebb_{\xi_z}\left[\mathcal{A}_t^i(j)\right]\right)^2\right] \notag \\ 
 &\leq 2 \sum_{i=1}^n\left( \Ebb_{\xi_z}\left[\left(\mathcal{A}_t^i(j)\right)^2\right]	+\left(\Ebb_{\xi_z}\left[\mathcal{A}_t^i(j)\right]\right)^2 \right) \notag  \\& \leq 4n\eta_z^2\sigma^2,
\end{align}
where equality (a) is true because $\mathcal{A}_t^1(j),...,\mathcal{A}_t^n(j)$ are independent to each other.  

Therefore, from \eqref{p:lm4:2} and \eqref{p:lm4_abc} we have

\begin{subequations}\label{p:comb0}
\begin{align}
	\Ebb_{\xi_z}\left[\sum_{j=1}^d L_j\left(x_t(j) - \bar x_t(j) \right)^2\right]  = & \sum_{j=1}^d L_j\Ebb_{\xi_z}\left[\left(x_t(j) - \bar x_t(j) \right)^2\right]\\ \leq & \frac{4\eta_z^2\gamma^2\sigma^2\sum_{j=1}^d L_j}{n} \notag \\ & + \frac{\gamma^2 }{n^2} \sum_{j=1}^d L_j \Ebb_{\xi_z}\left[\left(\sum_{i=1}^n \left(\Ebb_{\xi_z}\left[\mathcal{A}_t^i(j)\right]-\sum_{s=1}^E g_{t,s}^i(j)\right)\right)^2\right] \\
	\leq & \frac{4\eta_z^2\gamma^2\sigma^2\sum_{j=1}^d L_j}{n} \notag \\ & + \frac{\gamma^2 L_{\max}}{n^2}  \Ebb_{\xi_z}\left[\left\|\sum_{i=1}^n \left(\Ebb_{\xi_z}\left[\mathcal{A}_t^i\right]-\sum_{s=1}^E g_{t,s}^i\right)\right\|^2\right]\label{p:lm4:2_term2}.
\end{align}
\end{subequations}

To bound the RHS of \eqref{p:lm4:2_term2}, we have 
\begin{align}\label{p:comb1}
\Ebb_{\xi_z}\left[\left\|\sum_{i=1}^n \left(\Ebb_{\xi_z}\left[\mathcal{A}_t^i\right]-\sum_{s=1}^E g_{t,s}^i\right)\right\|^2\right] & \leq  n\sum_{i=1}^n \Ebb_{\xi_z}\left[\left\| \Ebb_{\xi_z}\left[\mathcal{A}_t^i\right]-\sum_{s=1}^E g_{t,s}^i \right\|^2\right] \notag \\ 
& \leq  \frac{n}{4(2z+1)^2\sigma^{4z}}\sum_{i=1}^n \left\|\sum_{s=1}^E g_{t,s}^i\right\|_{4z+2}^{4z+2},
\end{align} 
where the last inequality is due to  Lemma \ref{lm:phi_lm}.

Now we need to bound 
\begin{align*}
	\Ebb\left[ \left\|\sum_{s=1}^E g_{t,s}^i\right\|^{4z+2}_{4z+2}\right], 
\end{align*} 
where the expectation is taken over both $\xi_z$ and the minibatch gradient noise.
To this end, we need the following lemma about the $\ell_p$-norm.
\begin{lemma} \label{lm:lpnorm_col} For any $M\in \Zbb_+$, $p>1$ and $M$ vectors $x_1,...,x_M\in\Rbb^d$, we have 
    \begin{align}
        \left\|\sum_{i=1}^M x_i\right\|_p^p \leq M^{p-1}\sum_{i=1}^M \|x_i\|_p^p.
    \end{align}
\end{lemma}

As a direct application of Lemma \ref{lm:lpnorm_col}, we obtain 
\begin{align}
    \Ebb\left[ \left\|\sum_{s=1}^E g_{t,s}^i\right\|^{4z+2}_{4z+2}\right]
	&\leq \Ebb\left[E^{4z+1}\sum_{s=1}^E  \left\|g_{t,s}^i\right\|^{4z+2}_{4z+2}\right] =E^{4z+1}\sum_{s=1}^E  \Ebb\left[ \left\|g_{t,s}^i\right\|^{4z+2}_{4z+2}\right] \label{p:xxx}
\end{align}

Then we can bound the RHS of \eqref{p:xxx} as 
\begin{align}\label{p:comb3}
	\Ebb\left[ \left\|g_{t,s}^i\right\|^{4z+2}_{4z+2}\right]
	& = \Ebb\left[\left\| g_{t,s}^i- \nabla f_i(x_{t,s-1}^i)+\nabla f_i(x_{t,s-1}^i)\right\|^{4z+2}_{4z+2}\right] \notag \\
	&\overset{(a)}{\leq} \Ebb\left[2^{4z+1} \left\| g_{t,s}^i- \nabla f_ti(x_{t,s-1}^i)\right\|^{4z+2}_{4z+2}+2^{4z+1} \left\|\nabla f_i(x_{t,s-1}^i)\right\|^{4z+2}_{4z+2}\right] \notag \\
	&\overset{(b)}{\leq} 2^{4z+1} Q_z+2^{4z+1}\left\|\nabla f_i(x_{t,s-1}^i)\right\|^{4z+2}_2 \notag \\ 
	& \overset{(c)}{\leq} 2^{4z+1} (Q_z+G^{4z+2}), 
\end{align}
where inequality (a) follows 
Lemma \ref{lm:lpnorm_col}, inequality (b) is due to Assumption \ref{asp:z-moment}, and inequality (c) is due to A.4 of Assumption \ref{asp:common}.

Combing \eqref{p:comb0}, \eqref{p:comb1}, \eqref{p:xxx} and \eqref{p:comb3}, we have
\begin{align}
    \Ebb\left[\left\|\sum_{i=1}^n \left(\Ebb_{\xi_z}\left[\mathcal{A}_t^i\right]-\sum_{s=1}^E g_{t,s}^i\right)\right\|\right] &\leq \sqrt{\Ebb\left[\left\|\sum_{i=1}^n \left(\Ebb_{\xi_z}\left[\mathcal{A}_t^i\right]-\sum_{s=1}^E g_{t,s}^i\right)\right\|^2\right]} \notag \\
    &\leq \sqrt{\frac{  n^2 2^{4z} E^{4z+2}(Q_z+G^{4z+2})}{2(2z+1)^2\sigma^{4z}}} \notag \\
    &\leq \frac{  n 2^{2z} E^{2z+1}\sqrt{(Q_z+G^{4z+2})} }{\sqrt{2}(2z+1)\sigma^{2z}}
\end{align}
and
\begin{align}
	\Ebb\left[\sum_{j=1}^d L_j\left(x_t(j) - \bar x_t(j) \right)^2\right] \leq & \frac{4\eta_z^2\gamma^2\sigma^2\sum_{j=1}^d L_j}{n} + \frac{\gamma^2 L_{\max}}{n^2}  \Ebb\left[\left\|\sum_{i=1}^n \left(\Ebb_{\xi_z}\left[\mathcal{A}_t^i\right]-\sum_{s=1}^E g_{t,s}^i\right)\right\|^2\right] \notag \\ \leq & \frac{4\eta_z^2\gamma^2 \sigma^2\sum_{j=1}^d L_j}{n} + \frac{\gamma^2 2^{4z+1} E^{4z+2}(Q_z+G^{4z+2})L_{\max}}{4(2z+1)^2\sigma^{4z}}.
\end{align}

Hence, we have
\begin{align}
\Ebb\left[f(x_t) - f(\bar x_t)\right]
	\leq & \Ebb\left[\left\langle \nabla f(\bar x_t), \frac{\gamma }{n}\sum_{i=1}^n \left(\Ebb_{\xi_z}\left[\mathcal{A}_t^i\right]-\sum_{s=1}^E g_{t,s}^i\right)\right\rangle \right] \notag \\
	& +\Ebb\left[ \frac{ \sum_{j=1}^d L_j \left(x_t(j) - \bar x_t(j) \right)^2}{2}\right] \notag \\
	\leq &
	\|\nabla f(\bar x_t)\| \Ebb\left[\left\|\frac{\gamma }{n}\sum_{i=1}^n \left(\Ebb_{\xi_z}\left[\mathcal{A}_t^i\right]-\sum_{s=1}^E g_{t,s}^i\right)\right\|\right] \notag \\ 
	& +\Ebb\left[ \frac{ \sum_{j=1}^d L_j \left(x_t(j) - \bar x_t(j) \right)^2}{2}\right] \notag \\
	\leq & \frac{\gamma 2^{2z} E^{2z+1}\sqrt{Q_z+G^{4z+2}}G}{\sqrt 2(2z+1)\sigma^{2z}} + \frac{\gamma^2 2^{4z} E^{4z+2}(Q_z+G^{4z+2})L_{\max}}{4(2z+1)^2\sigma^{4z}} \notag \\ 
	& +\frac{2\eta_z^2\gamma^2 \sigma^2\sum_{j=1}^d L_j}{n}.
\end{align}
\end{proof}

\begin{proof}[Proof of Lemma \ref{lm:lpnorm_col}]
To prove this lemma, we need to use  a classical result on the monotonicity of $\ell_p$ norm:

\begin{lemma}\cite{kantorovich2016functional}
\label{lm:lpnorm}
    For any $x\in \Rbb^d$ and $1<r<p$, we have
\begin{align}
	\|x\|_p\leq \|x\|_r \leq d^{\frac{1}{r}-\frac{1}{p}}\|x\|_p.
\end{align}
\end{lemma}
Now from the definition of $\ell_p$ norm we have 
\begin{align}
     \left\|\sum_{i=1}^M x_i\right\|_p^p &= \sum_{j=1}^d \left(\sum_{i=1}^M x_i(j)\right)^p\leq \sum_{j=1}^d \left(\sum_{i=1}^M |x_i(j)|\right)^p \notag \\
     & = \sum_{j=1}^d  \|[x_1(j),...,x_M(j)]^\top\|_1^p \notag \\
     &\overset{(a)}{\leq} M^{p-1} \sum_{j=1}^d \|[x_1(j),...,x_M(j)]^\top\|_p^p \notag \\
     & = M^{p-1} \sum_{j=1}^d \sum_{i=1}^M \left(x_i(j)\right)^p \notag \\
     & = M^{p-1} \sum_{i=1}^M \left\| x_i\right\|_p^p,
\end{align} 
where inequality (a) is due to Lemma \ref{lm:lpnorm}.
\end{proof}

\begin{proof}[Proof of Lemma \ref{lm:thm2_lm2}] First we unroll the difference $f(\bar x_{t}) - f( x_{t-1})$ into a telescopic sum across $E$ local steps.
    \begin{subequations}\label{p:apple}
	\begin{align}
		f(\bar x_{t}) - f( x_{t-1}) &= f(\bar x_{t-1,E}) - f(\bar x_{t-1,0}) = \sum_{s=1}^E f(\bar x_{t-1,s}) - f(\bar x_{t-1,s-1})\\
		&\leq \sum_{s=1}^E \left(-\langle \nabla f(\bar x_{t-1,s-1}), \bar x_{t-1,s-1}-\bar x_{t-1,s}  \rangle + \frac{ L_{\max}}{2}\|\bar x_{t-1,s} - \bar x_{t-1,s-1} \|^2 \right)\\
		&=\sum_{s=1}^E \left(-\gamma\langle \nabla f(\bar x_{t-1,s-1}), \frac{1}{n}\sum_{i=1}^n g_{t-1,s}^i \rangle + \frac{ \gamma^2 L_{\max}}{2}\left\|\frac{1}{n}\sum_{i=1}^n g_{t-1,s}^i \right\|^2 \right),\label{p:asdqw}
	\end{align}
	\end{subequations} where the inequality is due to the smoothness assumption.
	Taking expectation over the minibatch gradient noise $g_{t-1,s}^1,...,g_{t-1,s}^n$, for the first terms in \eqref{p:asdqw}, we obtain
	\begin{subequations}\label{p:red}
	\begin{align}
		\Ebb\left[-\left\langle \nabla f(\bar x_{t-1,s-1}), \frac{1}{n}\sum_{i=1}^n g_{t-1,s}^i \right\rangle\right] =& -\left\langle \nabla f(\bar x_{t-1,s-1}), \frac{1}{n}\sum_{i=1}^n \nabla f_i(x_{t-1,s-1}^i)\right\rangle\\
		=& -\frac{1}{2}\left\|\nabla f(\bar x_{t-1,s-1})\right\|^2 -\frac{1}{2}\left\|\frac{1}{n}\sum_{i=1}^n \nabla f_i(x_{t-1,s-1}^i)\right\|^2 \\
		&+\frac{1}{2}\left\|\nabla f(\bar x_{t-1,s-1})-\frac{1}{n}\sum_{i=1}^n \nabla f_i(x_{t-1,s-1}^i)\right\|^2. \label{p:okoinoi}
	\end{align}
	\end{subequations}
	
	For the second terms in \eqref{p:asdqw}, we have
	\begin{align}\label{p:blue}
		\Ebb\left[\left\|\frac{1}{n}\sum_{i=1}^n g_{t-1,s}^i \right\|^2\right]& = \Ebb\left[\left\|\frac{1}{n}\sum_{i=1}^n g_{t-1,s}^i -\frac{1}{n}\sum_{i=1}^n \nabla f_i(x_{t-1,s-1}^i) +\frac{1}{n}\sum_{i=1}^n \nabla f_i(x_{t-1,s-1}^i) \right\|^2\right] \notag \\
		& \overset{(a)}{=} \Ebb\left[\left\|\frac{1}{n}\sum_{i=1}^n g_{t-1,s}^i -\frac{1}{n}\sum_{i=1}^n \nabla f_i(x_{t-1,s-1}^i)  \right\|^2\right] + \left\|\frac{1}{n}\sum_{i=1}^n \nabla f_i(x_{t-1,s-1}^i) \right\|^2 \notag \\
		& \overset{(b)}{=} \frac{1}{n^2}\sum_{i=1}^n\Ebb\left[\left\| g_{t-1,s}^i -\frac{1}{n}\sum_{i=1}^n \nabla f_i(x_{t-1,s-1}^i)  \right\|^2\right] + \left\|\frac{1}{n}\sum_{i=1}^n \nabla f_i(x_{t-1,s-1}^i) \right\|^2 \notag 
		\\& \overset{(c)}{\leq}   \frac{\zeta^2}{n}+\left\|\frac{1}{n}\sum_{i=1}^n \nabla f_i(x_{t-1,s-1}^i) \right\|^2,
	\end{align}
	where equalities (a) and (b) are true because the minibatch gradient noise is independent, and inequality (c) is due to A.1 of Assumption \ref{asp:common}.

	Notice that owing to the function smoothness, we have for arbitrary $x,y\in \Rbb^d$, 
\begin{align}
	f(y)\leq \langle \nabla f(x),y-x\rangle + \frac{L_{\max} }{2}\|y-x\|^2,
\end{align}which is equivalent to 
\begin{align}
	\| \nabla f(x)-\nabla f(y)\| \leq  L_{\max} \|y-x\|.
\end{align}

Now to bound the term in \eqref{p:okoinoi}, for every $s$, we have 
\begin{align}\label{p:oijoiw}
	&\left\|\nabla f(\bar x_{t-1,s-1})-\frac{1}{n}\sum_{i=1}^n \nabla f_i(x_{t-1,s-1}^i)\right\|^2 \notag \\
	&= \left\|\frac{1}{n}\sum_{i=1}^n \nabla f_i(\bar x_{t-1,s-1})-\frac{1}{n}\sum_{i=1}^n \nabla f_i(x_{t-1,s-1}^i)\right\|^2 \notag \\
	&\leq  \frac{L^2}{n} \sum_{i=1}^n\|\bar x_{t-1,s-1}-x_{t-1,s-1}^i\|^2 \notag \\
	&=  \frac{\gamma^2 L_{\max}^2}{n} \sum_{i=1}^n\left\|\sum_{q=1}^{s-1} \left(\frac{1}{n}\sum_{j=1}^n g_{t-1,q}^j -g_{t-1,q}^i\right)\right\|^2 \notag \\
	&\leq  \frac{(s-1)\gamma^2 L_{\max}^2}{n} \sum_{i=1}^n\sum_{q=1}^{s-1}\left\| \frac{1}{n}\sum_{j=1}^n g_{t-1,q}^j -g_{t-1,q}^i\right\|^2 \notag \\
	&\leq  \frac{2(s-1)\gamma^2 L_{\max}^2}{n} \sum_{i=1}^n\sum_{q=1}^{s-1}\left(\left\| \frac{1}{n}\sum_{j=1}^n g_{t-1,q}^j \right\|^2+\left\| g_{t-1,q}^i\right\|^2\right) \notag \\
		&\leq  \frac{2(s-1)\gamma^2 L_{\max}^2}{n} \sum_{i=1}^n\sum_{q=1}^{s-1}\left(\frac{1}{n}\sum_{j=1}^n\left\|  g_{t-1,q}^j \right\|^2+\left\| g_{t-1,q}^i\right\|^2\right).
\end{align}

For any $t=1,...,T$, $i=1,...,n$ and $q=1,...,s-1$, taking expectation over minibatch gradient noise, we have 
\begin{align}\label{p:ubuk}
    \Ebb\left[\left\|  g_{t-1,q}^j \right\|^2\right] &= \Ebb\left[\left\|  g_{t-1,q}^i-\nabla f_i(x_{t-1,q-1}^i) + \nabla f_i(x_{t-1,q-1}^i) \right\|^2\right] \notag \\
    &\leq \Ebb\left[\left\|  g_{t-1,q}^i-\nabla f_i(x_{t-1,q-1}^i) \right\|^2\right] + \left\| \nabla f_i(x_{t-1,q-1}^i) \right\|^2 \notag \\
    &\leq \zeta^2+G^2.
\end{align}

Substituting \eqref{p:ubuk} into \eqref{p:oijoiw}, we have 
\begin{align}\label{p:uncle}
	&\left\|\nabla f(\bar x_{t-1,s-1})-\frac{1}{n}\sum_{i=1}^n \nabla f_i(x_{t-1,s-1}^i)\right\|^2
	\leq  {4(s-1)^2\gamma^2 L_{\max}^2}(\zeta^2+G^2).
\end{align}

Further substituting \eqref{p:blue}, \eqref{p:red} and \eqref{p:uncle} into \eqref{p:apple} and by rearranging the terms, we obtain 
\begin{align}\label{p:taco}
\Ebb[f(\bar x_{t}) - f( x_{t-1})] 
	\leq & \sum_{s=1}^E \bigg(-\frac{\gamma}{2}\|\nabla f(\bar x_{t-1,s-1})\|^2 +\frac{\gamma^2L_{\max}-\gamma}{2}\left\|\frac{1}{n}\sum_{i=1}^n \nabla f_i(x_{t-1,s-1}^i)\right\|^2  \bigg) \notag \\
	&+ \sum_{s=1}^E\bigg( \frac{\gamma^2\zeta^2L_{\max}}{2n} +\frac{\gamma}{2}\|\nabla f(\bar x_{t-1,s-1})-\frac{1}{n}\sum_{i=1}^n \nabla f_i(x_{t-1,s-1}^i)\|^2 \bigg) \notag \\
	\overset{(a)}{\leq} & -\frac{\gamma}{2} \sum_{s=1}^E \|\nabla f(\bar x_{t-1,s-1})\|^2 +\frac{E\gamma^2\zeta^2L_{\max}}{2n} \notag \\
	& +\sum_{s=1}^E {2(s-1)^2\gamma^3 L_{\max}^2}(\zeta^2+G^2),
\end{align}	
where inequality (a) is by $(\gamma^2L_{\max}-\gamma)\leq 0$.

Note that 
\begin{align}
    \sum_{s=1}^E (s-1)^2 = \frac{(E-1)E(2E-1)}{6}\leq \frac{(E-1)E^2}{3}.
\end{align}
By applying it to \eqref{p:taco}, we finally have

\begin{align}
    \Ebb[f(\bar x_{t}) - f( x_{t-1})] \leq -\frac{\gamma}{2} \sum_{s=1}^E \|\nabla f(\bar x_{t-1,s-1})\|^2 +\frac{E\gamma^2\zeta^2L_{\max}}{2n} + \frac{2\gamma^3 (E-1)E^2 L_{\max}^2(\zeta^2+G^2)}{3}.
\end{align}

\end{proof}	

\subsubsection{Proof of Theorem \ref{thm:sign-fedavg_inf}}

We need a lemma similar to Lemma \ref{lm:thm2_lm1}.

\begin{lemma} Suppose that Assumption \ref{asp:common} and \ref{asp:inf} hold. For the $t$-th ($1\leq t\leq T$) communication round in Algorithm \ref{alg:SignFedAvg}, if
  $\eta = \sigma$ and $z=+\infty$, and $\sigma>E(G+Q_\infty)$, then
	\label{lm:thm2_lm3}
	\begin{align}
		\Ebb[f(x_t) - f(\bar x_t)]
			&\leq \frac{2\gamma^2\sigma^2\sum_{j=1}^d L_j }{n}.
	\end{align}
	
\end{lemma}

Following the similar idea as in the proof of Theorem \ref{thm:sign-fedavg_z}, we have 
\begin{align}
	&\Ebb[f(x_t) - f( x_{t-1})] = \Ebb[f(x_t) - f(\bar x_t)]+E[f(\bar x_t)-f( x_{t-1})] \notag \\
	\leq & -\frac{\gamma}{2}\sum_{s=1}^E \|\nabla f(\bar x_{t-1,s-1})\|^2 + \frac{E\gamma^2\zeta^2L_{\max}}{2n} \notag 
	\\ & + \frac{2\gamma^3 (E-1)E^2 L_{\max}^2(\zeta^2+G^2)}{3}+\frac{2\gamma^2\sigma^2\sum_{j=1}^d L_j }{n}.
\end{align}

Rearranging the terms, we have
\begin{align}
	\frac{1}{E}\sum_{s=1}^E \|\nabla f(\bar x_{t-1,s-1})\|^2 \leq & \frac{2\Ebb[ f( x_{t-1})-f(x_t) ]}{E\gamma} + \frac{\gamma\zeta^2L_{\max}}{n} \notag  \\ & + \frac{4\gamma^2 (E-1)E L_{\max}^2(\zeta^2+G^2)}{3} +\frac{4\gamma\sigma^2\sum_{j=1}^d L_j }{En}.
\end{align}

Form the telescopic sum, we obtain

\begin{align}
	\Ebb\left[\frac{1}{TE}\sum_{t=1}^{T}\sum_{s=1}^E \|\nabla f(\bar x_{t-1,s-1})\|^2 \right] \leq & \frac{2\Ebb[f(x_0) - f^*]}{TE\gamma} + \frac{\gamma\zeta^2L_{\max}}{n} \notag  \\ & +  \frac{4\gamma^2 (E-1)E L_{\max}^2(\zeta^2+G^2)}{3}+\frac{4\gamma\sigma^2\sum_{j=1}^d L_j }{En}.
\end{align}

Here we provide a simple example to show that when $\sigma<E(G+Q_\infty)$, the algorithm cannot converge.
Consider $E=1$, $Q_\infty=0$ and the problem  $$\min_{x\in\Rbb}(x-A)^2+(x+A)^2, $$ where $A>0$ is some positive number. If we choose the initial to be $x_0=\frac{A}{2}$. As one can see, the gradient at $x_0$ for the two parts of the objective function are $-A$ and $3A$, respectively. We denote that $\xi_\infty$ as the random noise following uniform distribution at $[-1,1]$. If now $\sigma<A$, we have \begin{align}
	\text{Sign}(-A+\sigma \xi_\infty)+\text{Sign}(3A+\sigma \xi_\infty)=0,
\end{align}i.e., this algorithm never update the variable.

\begin{proof}[Proof of Lemma \ref{lm:thm2_lm3}]
	We first note that, when $z=+\infty$, we have 
\begin{align}\label{p:psi_def}
    \Psi_{\infty}(x) = 
\begin{cases}
  x & x\in[-1,1], \\
  -1 & x<-1,\\
  1 & x>1.
\end{cases}
\end{align}

	Again, from the smoothness assumption (A.2 in Assumption \ref{asp:common}) we have,
	\begin{align}\label{p:smooth_again}
		f(x_t) - f(\bar x_t)
		&\leq \langle \nabla f(\bar x_t), x_t - \bar x_t \rangle + \frac{ \sum_{j=1}^d L_j \left(x_t(j) - \bar x_t(j) \right)^2}{2}.
	\end{align}

 Taking expectation over $\xi_\infty$,
\begin{align}\label{p:learn}
	\Ebb_{\xi_\infty}[x_t - \bar x_t] &= \Ebb_{\xi_\infty}\left[\frac{\gamma }{n}\sum_{i=1}^n \left(\sigma \text{Sign}\left(\sum_{s=1}^E g_{t,s}^i+\sigma \xi_\infty\right)	- \sum_{s=1}^E g_{t,s}^i\right)\right] \notag \\ 
	&\overset{(a)}{=} \frac{\gamma }{n}\sum_{i=1}^n \left(\sigma \Psi_{\infty}\left(\frac{\sum_{s=1}^E g_{t,s}^i}{\sigma}\right)	- \sum_{s=1}^E g_{t,s}^i\right)\notag \\
	&\overset{(b)}{=} \frac{\gamma }{n}\sum_{i=1}^n \left(\sum_{s=1}^E g_{t,s}^i	- \sum_{s=1}^E g_{t,s}^i\right)=0,
\end{align}
where equality (a) is because for any $x\in\Rbb^d$, $\Ebb_{\xi_\infty}[\text{Sign}(x+\sigma \xi_\infty)] = \Psi_{\infty}(x/\sigma)$, equality (b) is due to $\sigma>\|\sum_{s=1}^E g_{t,s}^i\|_\infty$ almost surely and the property of the function $\Psi_{\infty}(\cdot)$ in \eqref{p:psi_def}.

For ease of presentation, we define that
\begin{align}
\label{p:def_Bc}
    \mathcal{B}_t^i \stackrel{\text{def.}}= \sigma \text{Sign}\left(\sum_{s=1}^E g_{t,s}^i+\sigma \xi_\infty\right).
\end{align}

From \eqref{p:learn} we have learned that $\Ebb_{\xi_\infty}[\mathcal{B}_t^i]=\sum_{s=1}^E g_{t,s}^i$. Thus, for any $j=1,...,d$, we have
\begin{align}\label{p:rddd}
	\Ebb_{\xi_\infty}[\left(x_t(j) - \bar x_t(j) \right)^2] &\leq  \frac{\gamma ^2}{n^2}\Ebb_{\xi_\infty}\left[\left(\sum_{i=1}^n \left( \mathcal{B}_t^i(j)	- \Ebb_{\xi_\infty}\left[\mathcal{B}_t^i(j)\right]\right)\right)^2\right] \notag \\
	&=\frac{\gamma ^2}{n^2}\sum_{i=1}^n\Ebb_{\xi_\infty}\left[ \left( \mathcal{B}_t^i(j)	- \Ebb_{\xi_\infty}\left[\mathcal{B}_t^i(j)\right]\right)^2\right] \notag \\
	&\leq \frac{2\gamma ^2}{n^2}\sum_{i=1}^n\left(\Ebb_{\xi_\infty}\left[ \left( \mathcal{B}_t^i(j)	\right)^2\right]+\left( \Ebb_{\xi_\infty}\left[\mathcal{B}_t^i(j)\right]\right)^2\right) \notag \\
	&\leq \frac{4\gamma^2\sigma^2}{n}.
\end{align}

Finally, substituting \eqref{p:learn} and  \eqref{p:rddd} into \eqref{p:smooth_again}, and taking the expectation over both $\xi_\infty$ and the minibatch gradient noise, we have
\begin{align}
	\Ebb[f(x_t) - f(\bar x_t)]
		&\leq \Ebb[\langle \nabla f(\bar x_t), x_t - \bar x_t \rangle ]+ \Ebb\left[\frac{ \sum_{j=1}^d L_j \left(x_t(j) - \bar x_t(j) \right)^2}{2}\right] \notag \\
		&\leq \frac{2\gamma^2\sigma^2\sum_{j=1}^d L_j }{n}.
\end{align}

\end{proof}

\subsection{Experiment details}
\subsubsection{Details for the experiment in Section \ref{sec:exp2}}
\label{app:exp2}

In Table \ref{table:hp_MNIST}, we provide the tuned hyperparameters for all the tested algorithms on non-i.i.d MNIST. Specifically, we tuned the hyperparameters via grid search: $[0.1,0.05,0.01,0.005]$ for stepsize, $[0,0.3,0.5,0.7, 0.9]$ for the momentum coefficient, and $[0,0.02,0.05,0.01,0.03,0.05,0.1,0.3,0.5]$ for the noise scale.
\begin{table}[htpb]
	\centering
	\begin{tabular}{|c| c| c| c|} 
	 \hline
	 \makecell{\bfseries Algorithm} &{\bfseries Stepsize}& {\bfseries Momentum coefficient} & {\bfseries Noise scale} \\ 
	 \hline\hline
	SGDwM & 0.05 & 0.9 & \diag{.1cm}{1.5cm}{}{}  \\ \hline
	 EF-SignSGDwM  & 0.05 & 0.9& \diag{.1cm}{1.5cm}{}{} \\\hline
	 Sto-SignSGDwM  & 0.01& 0.9& \diag{.1cm}{1.5cm}{}{} \\\hline
	 SignSGD & 0.01&0& 0\\ 
	 \hline
	 $1$-SignSGD & 0.01&0& 0.05 \\ 
	 \hline
	 $\infty$-SignSGD  & 0.01&0& 0.05  \\ 
	 \hline
	\end{tabular}

	\caption{Hyperparameters used for FL on non-i.i.d MNIST.}
	\label{table:hp_MNIST}
	\end{table}

\begin{figure}[htbp]
	\centering
	\begin{subfigure}[b]{0.48\textwidth}
	\centering
	\includegraphics[width=\textwidth]{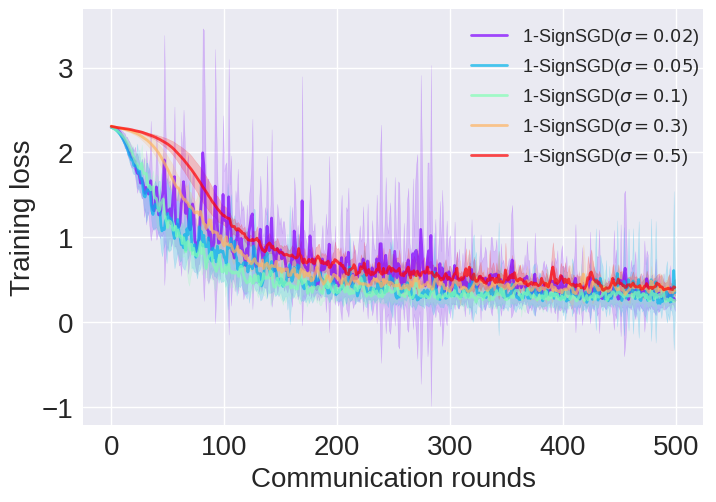}
	\caption{Training Loss:$1$-SignSGD}
	\label{fig:exp2_noise_tl_1}
\end{subfigure}
	\begin{subfigure}[b]{0.48\textwidth}
	\centering
	\includegraphics[width=\textwidth]{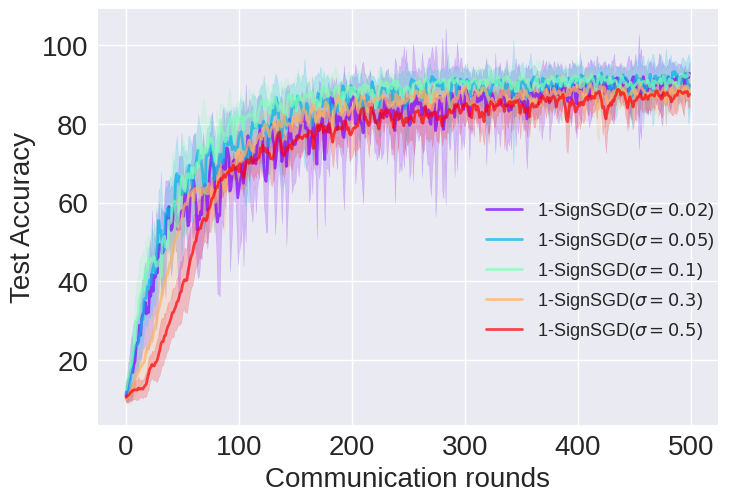}
	\caption{Test Accuracy:$1$-SignSGD}
	\label{fig:exp2_noise_acc_1}
\end{subfigure}
\begin{subfigure}[b]{0.48\textwidth}
	\centering
	\includegraphics[width=\textwidth]{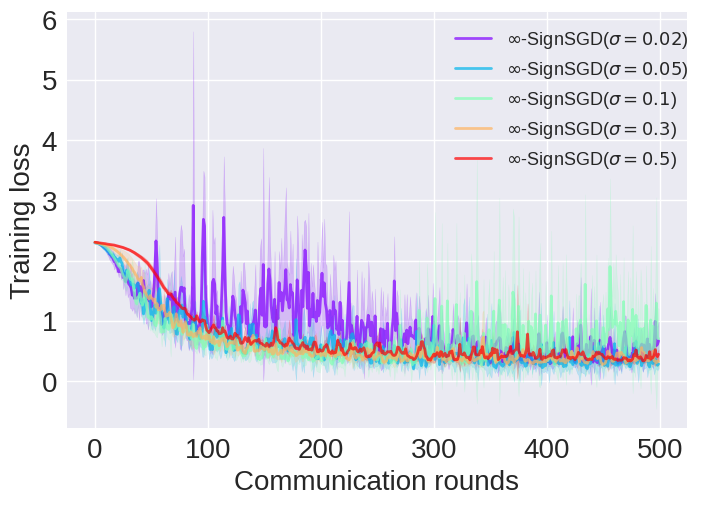}
	\caption{Training Loss:$\infty$-SignSGD}
	\label{fig:exp2_uninoise_tl_inf}
\end{subfigure}
	\begin{subfigure}[b]{0.48\textwidth}
	\centering
	\includegraphics[width=\textwidth]{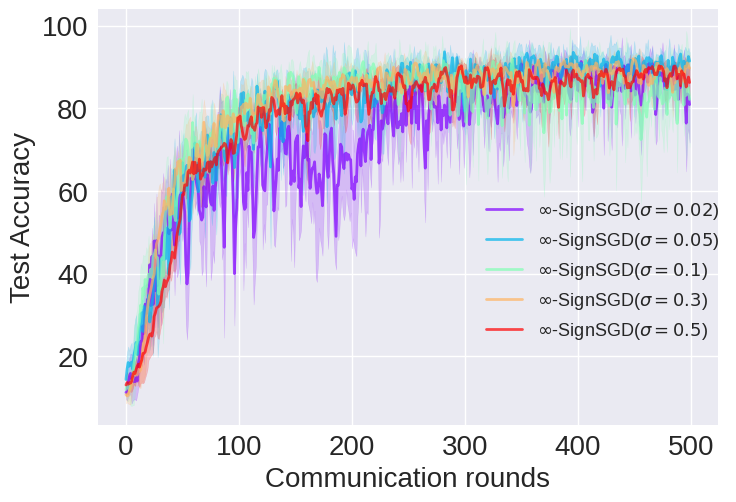}
	\caption{Test Accuracy:$\infty$-SignSGD}
	\label{fig:exp2_uninoise_acc_inf}
\end{subfigure}
\caption{$z$-SignFedAvg under different noise scales on non-i.i.d MNIST}
	\label{fig:Sign-SGD-Simplen_noise_MNIST}
\end{figure}

In Figure \ref{fig:Sign-SGD-Simplen_noise_MNIST}, we visualize the performance of $1$-SignSGD and $\infty$-SignSGD under different noise scales. As we can see, the results for $1$-SignSGD and $\infty$-SignSGD are almost the same, except that the $\infty$-SignSGD is slightly better than $1$-SignSGD when the noise scale is large.

% We can see that sometimes SignFedAvg can work, this is possibly because the minibatch gradient noise is symmetric and the variance is already enough.

\subsubsection{Details for the experiment in Section \ref{sec:exp3}}
\label{app:exp3}
We denote the noiseless case, i.e., Algorithm \ref{alg:SignFedAvg} with $\sigma=0$ as SignFedAvg.

\begin{figure}[htbp]
 \begin{minipage}[t]{1\linewidth}
\centering
\includegraphics[width= 5.5 in ]{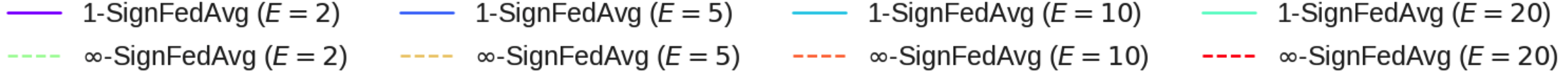}
\end{minipage}
	\centering
\begin{subfigure}[b]{0.32\textwidth}
	\centering
	\includegraphics[width=\textwidth]{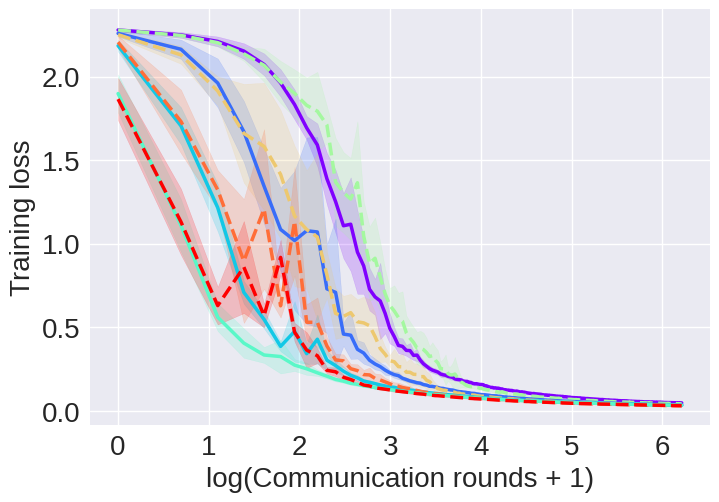}
	\caption{Training Loss}
	\label{}
\end{subfigure}
	\begin{subfigure}[b]{0.32\textwidth}
	\centering
	\includegraphics[width=\textwidth]{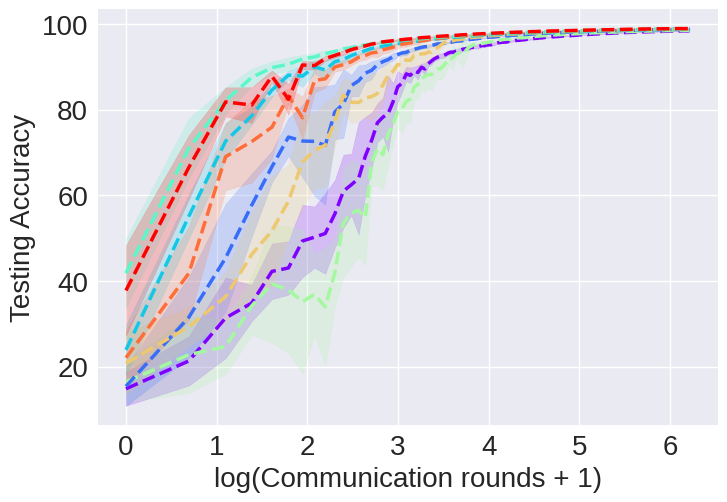}
	\caption{Test Accuracy}
	\label{}
\end{subfigure}
\begin{subfigure}[b]{0.32\textwidth}
	\centering
	\includegraphics[width=\textwidth]{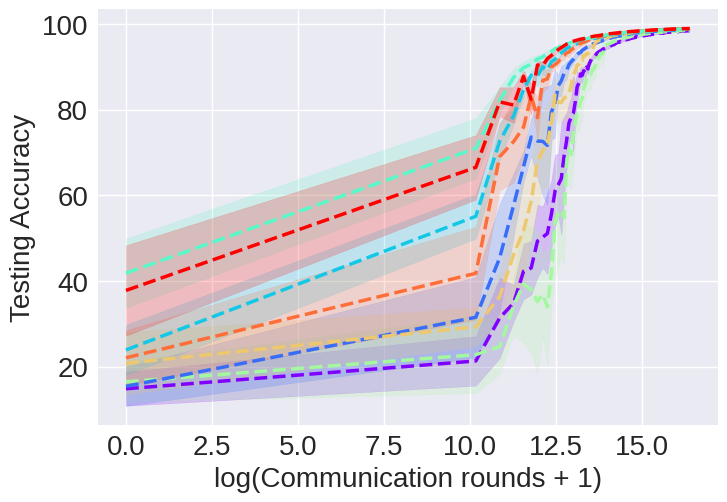}
	\caption{Test Accuracy w.r.t bits}
	\label{}
\end{subfigure}
\caption{Performance of $1$-SignFedAvg and $\infty$-SignFedAvg on EMNIST dataset. }
\label{fig:exp3-mnist_comp}
\end{figure}

\begin{figure}[htbp]
\begin{minipage}[t]{1\linewidth}
\centering
\includegraphics[width= 4.5 in ]{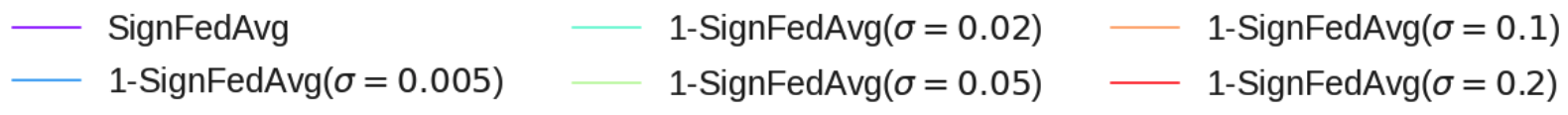}
\end{minipage}
	\centering
	\begin{subfigure}[b]{0.24\textwidth}
	\centering
	\includegraphics[width=\textwidth]{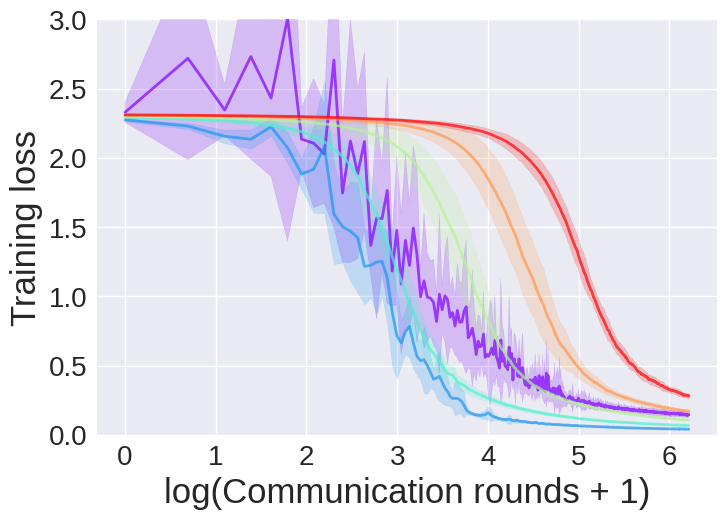}
	\caption{$E=2$}
	\label{}
\end{subfigure}
	\begin{subfigure}[b]{0.24\textwidth}
	\centering
	\includegraphics[width=\textwidth]{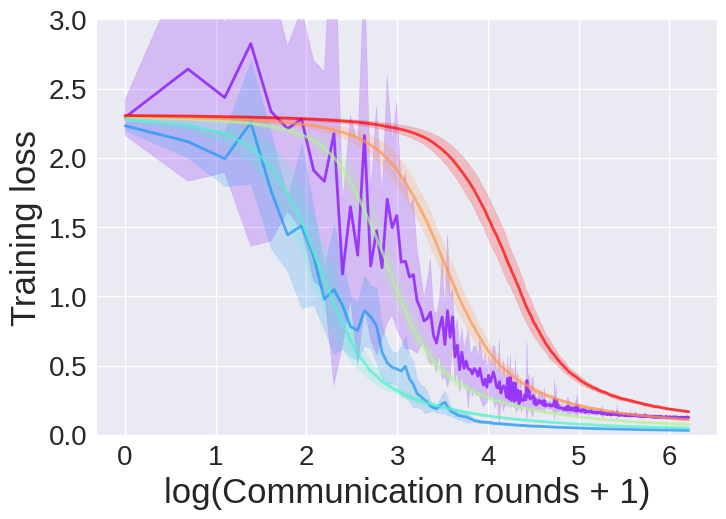}
	\caption{$E=5$}
	\label{}
\end{subfigure}
\begin{subfigure}[b]{0.24\textwidth}
	\centering
	\includegraphics[width=\textwidth]{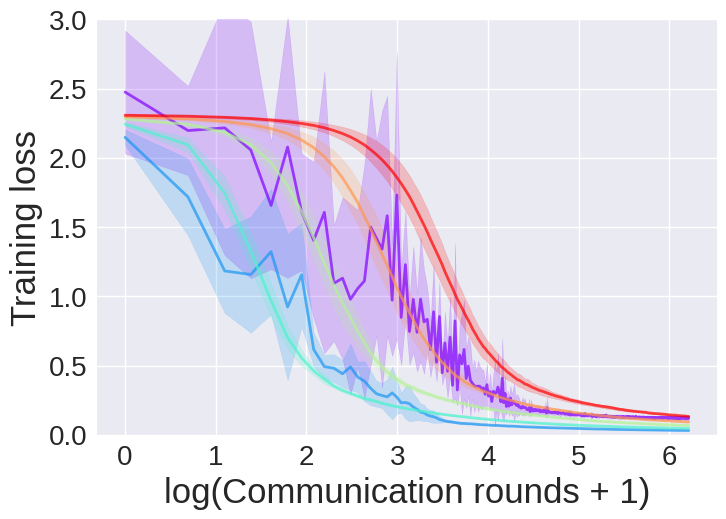}
	\caption{$E=10$}
	\label{}
\end{subfigure}
	\begin{subfigure}[b]{0.24\textwidth}
	\centering
	\includegraphics[width=\textwidth]{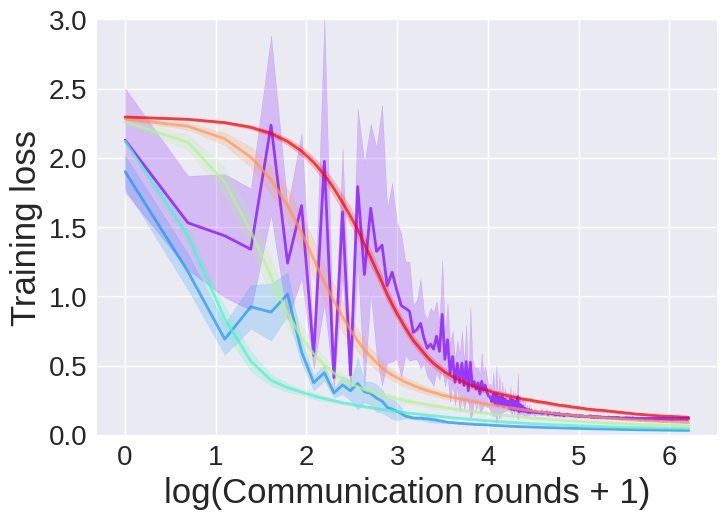}
	\caption{$E=20$}
	\label{}
\end{subfigure}
	\begin{subfigure}[b]{0.24\textwidth}
	\centering
	\includegraphics[width=\textwidth]{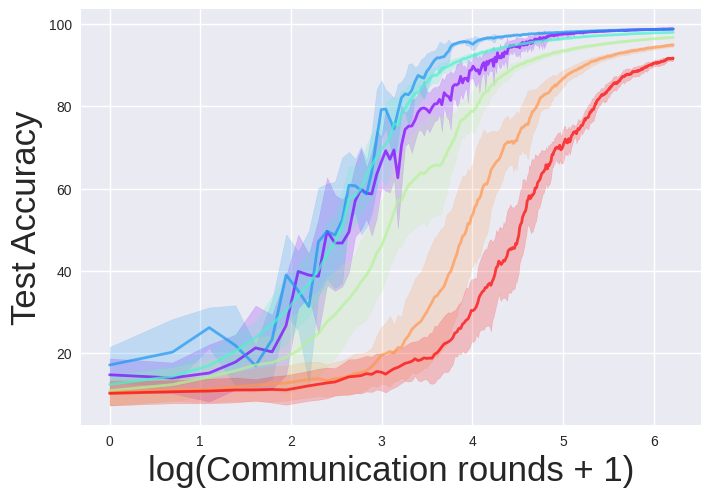}
	\caption{$E=2$}
	\label{}
\end{subfigure}
	\begin{subfigure}[b]{0.24\textwidth}
	\centering
	\includegraphics[width=\textwidth]{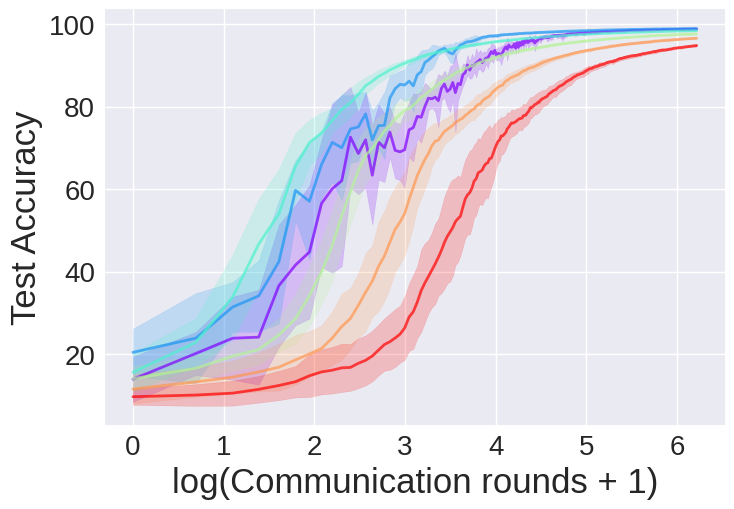}
	\caption{$E=5$}
	\label{}
\end{subfigure}
\begin{subfigure}[b]{0.24\textwidth}
	\centering
	\includegraphics[width=\textwidth]{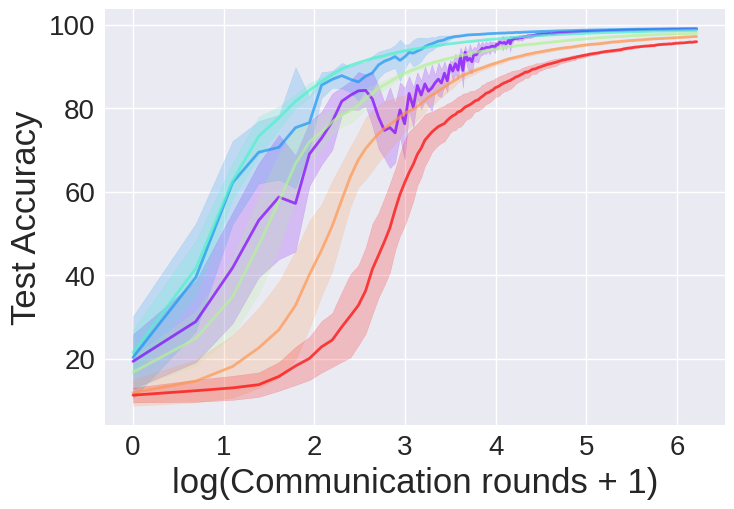}
	\caption{$E=10$}
	\label{}
\end{subfigure}
	\begin{subfigure}[b]{0.24\textwidth}
	\centering
	\includegraphics[width=\textwidth]{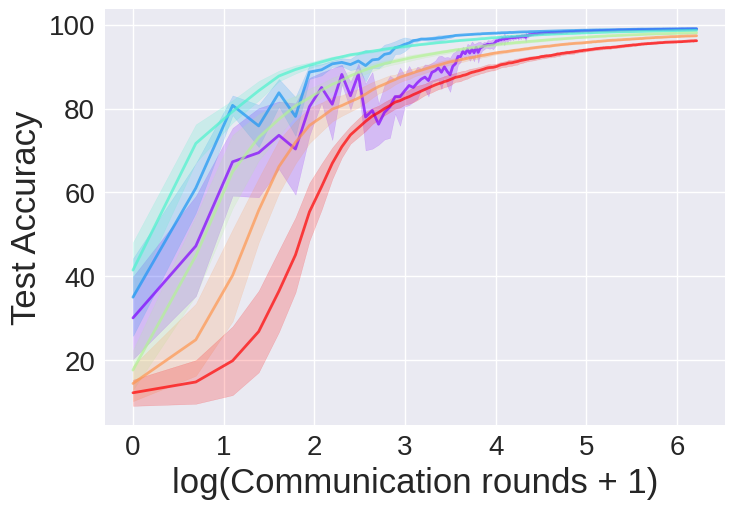}
	\caption{$E=20$}
	\label{}
\end{subfigure}
\caption{EMNIST: $1$-SignFedAvg under different noise scales and different numbers of local steps}
\label{fig:emnist_1sign}
\end{figure}

\begin{figure}[htbp]
\begin{minipage}[t]{1\linewidth}
\centering
\includegraphics[width= 4.5 in ]{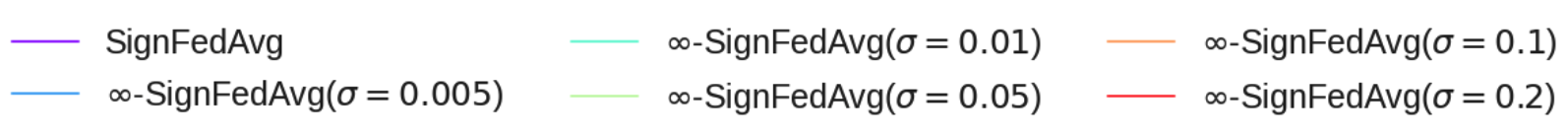}
\end{minipage}
	\centering
	\begin{subfigure}[b]{0.24\textwidth}
	\centering
	\includegraphics[width=\textwidth]{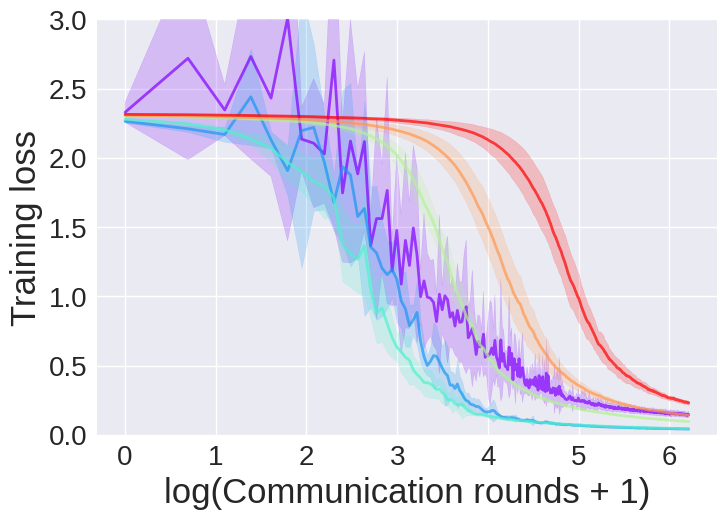}
	\caption{$E=2$}
	\label{}
\end{subfigure}
	\begin{subfigure}[b]{0.24\textwidth}
	\centering
	\includegraphics[width=\textwidth]{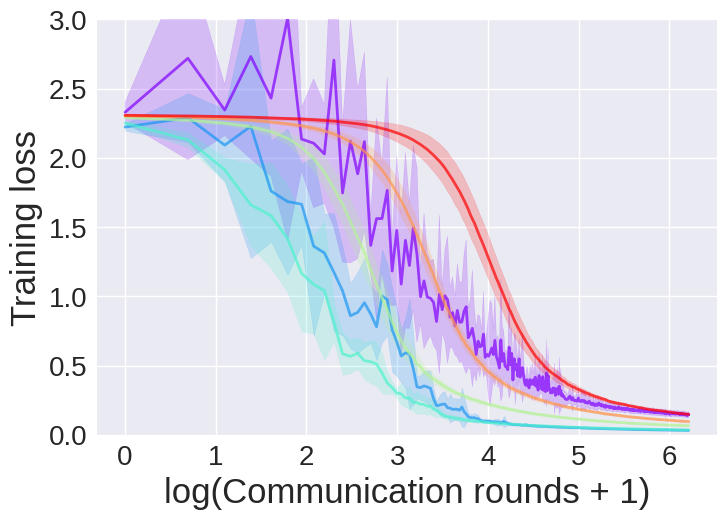}
	\caption{$E=5$}
	\label{}
\end{subfigure}
\begin{subfigure}[b]{0.24\textwidth}
	\centering
	\includegraphics[width=\textwidth]{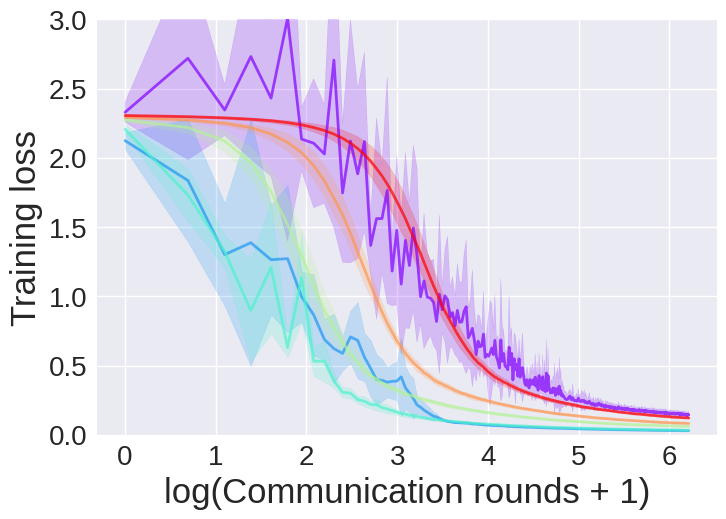}
	\caption{$E=10$}
	\label{}
\end{subfigure}
	\begin{subfigure}[b]{0.24\textwidth}
	\centering
	\includegraphics[width=\textwidth]{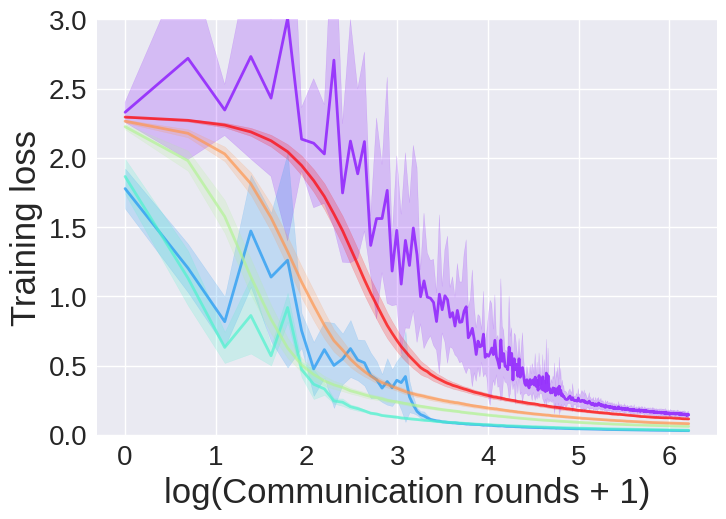}
	\caption{$E=20$}
	\label{}
\end{subfigure}
\begin{subfigure}[b]{0.24\textwidth}
	\centering
	\includegraphics[width=\textwidth]{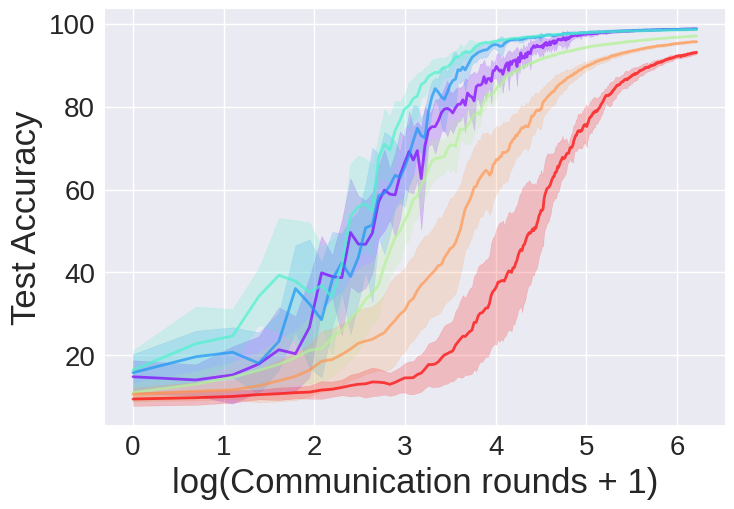}
	\caption{$E=2$}
	\label{}
\end{subfigure}
	\begin{subfigure}[b]{0.24\textwidth}
	\centering
	\includegraphics[width=\textwidth]{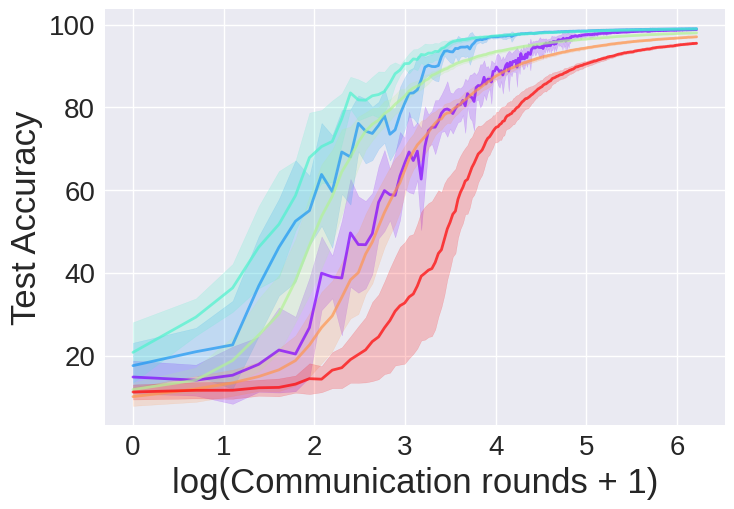}
	\caption{$E=5$}
	\label{}
\end{subfigure}
\begin{subfigure}[b]{0.24\textwidth}
	\centering
	\includegraphics[width=\textwidth]{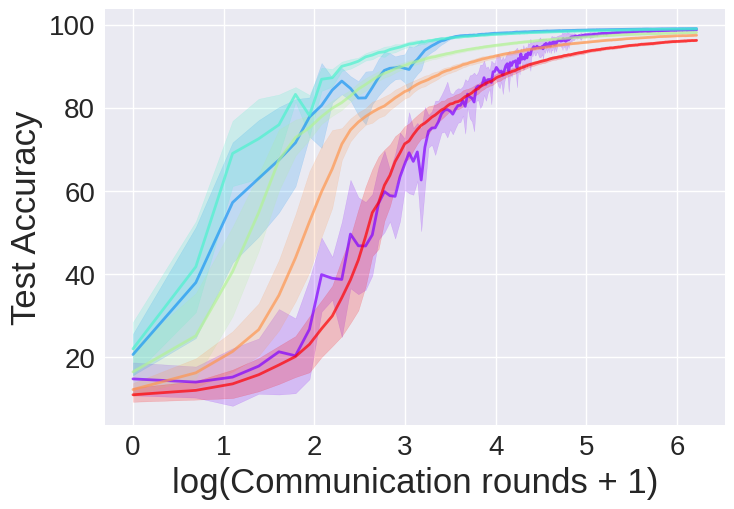}
	\caption{$E=10$}
	\label{}
\end{subfigure}
	\begin{subfigure}[b]{0.24\textwidth}
	\centering
	\includegraphics[width=\textwidth]{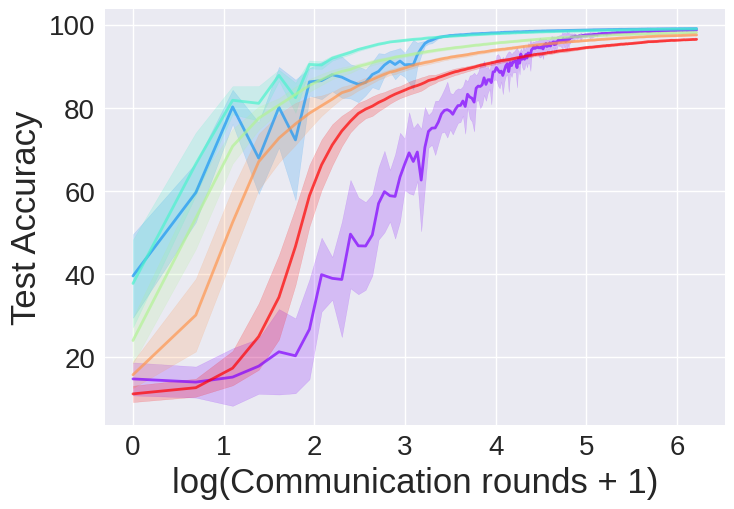}
	\caption{$E=20$}
	\label{}
\end{subfigure}
\caption{EMNIST: $\infty$-SignFedAvg under different noise scales and different numbers of local steps}
\label{fig:emnist_infsign}
\end{figure}

 \textbf{EMNIST:} For the experiment on EMNIST, we fixed the client stepsize as 0.05. We tuned the server stepsize, noise scales via grid search: $[1,0.5,0.1,0.05,0.01,0.005]$ for stepsize, $[0,0.005,0.02,0.05,0.01,0.03,0.05,0.1,0.2]$ for noise scale. The comparison between $1$-SignFedAvg and $\infty$-SignFedAvg on EMNIST is shown in Figure \ref{fig:exp3-mnist_comp}. The used hyperparameter in the top row of Figure \ref{fig:exp3-cifar} and \ref{fig:exp3-mnist_comp} are summarized in Table \ref{table:hp_EMNIST}.  We also visualize the performance of $1$-SignFedAvg and $\infty$-SignFedAvg under various noise scales and local steps in Figure \ref{fig:emnist_1sign} and Figure \ref{fig:emnist_infsign}.
 
\begin{figure}[htbp]
\begin{minipage}[t]{1\linewidth}
\centering
\includegraphics[width= 4.5 in ]{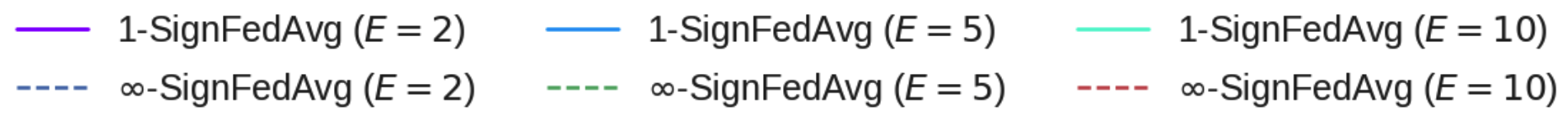}
\end{minipage}
	\centering
\begin{subfigure}[b]{0.32\textwidth}
	\centering
	\includegraphics[width=\textwidth]{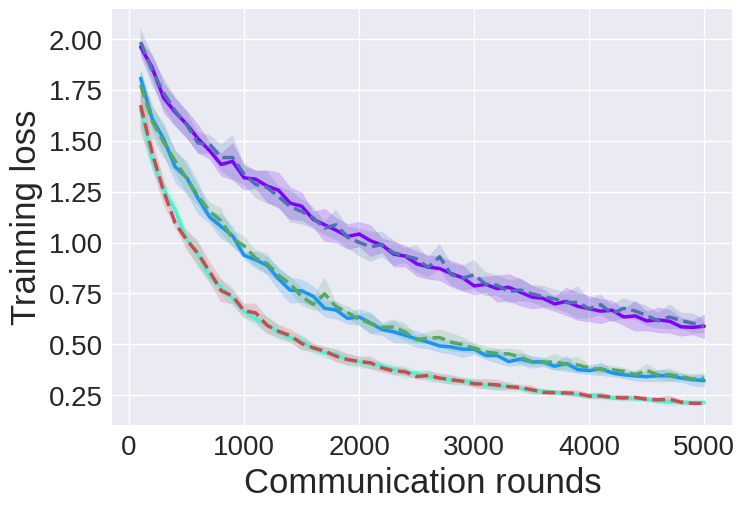}
	\caption{Training Loss}
	\label{}
\end{subfigure}
	\begin{subfigure}[b]{0.32\textwidth}
	\centering
	\includegraphics[width=\textwidth]{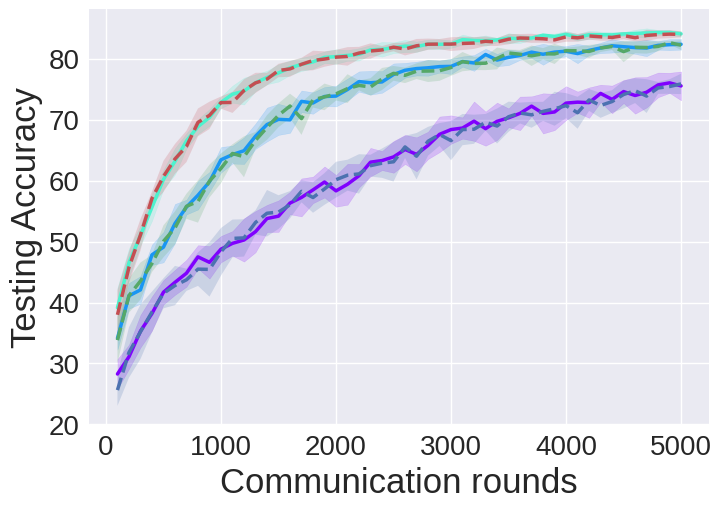}
	\caption{Test Accuracy}
	\label{}
\end{subfigure}
\begin{subfigure}[b]{0.32\textwidth}
	\centering
	\includegraphics[width=\textwidth]{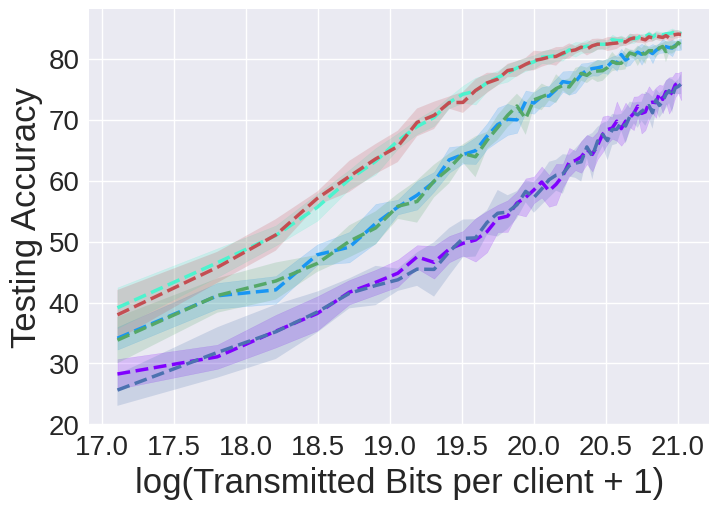}
	\caption{Test Accuracy w.r.t bits}
	\label{}
\end{subfigure}
\caption{Performance of  $1$-SignFedAvg and $\infty$-SignFedAvg on CIFAR-10 dataset. }
\label{fig:exp3-cifar_comp}
\end{figure}

\begin{figure}[htbp]
\begin{minipage}[t]{1\linewidth}
\centering
\includegraphics[width= 4.5 in ]{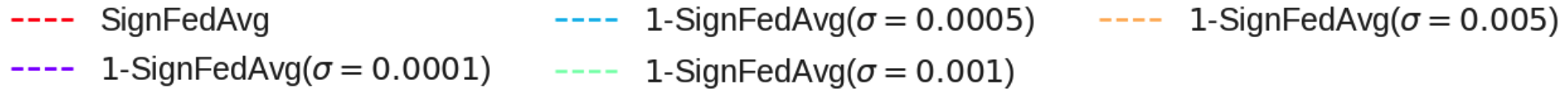}
\end{minipage}
	\centering
	\begin{subfigure}[b]{0.32\textwidth}
	\centering
	\includegraphics[width=\textwidth]{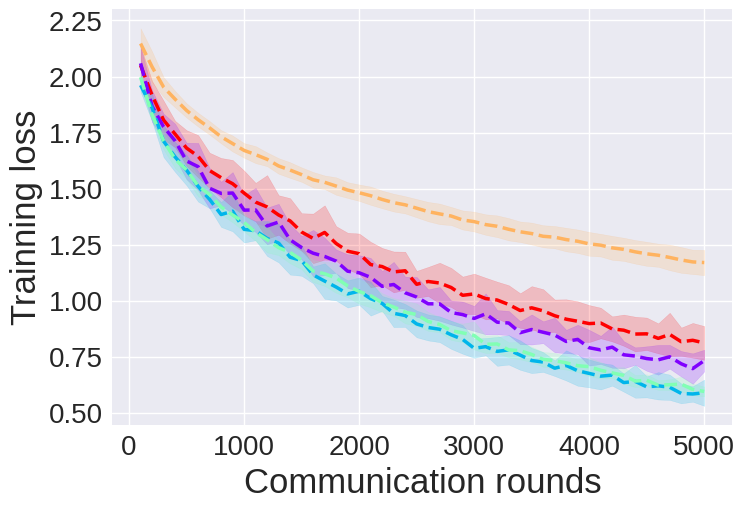}
	\caption{$E=2$}
	\label{}
\end{subfigure}
	\begin{subfigure}[b]{0.32\textwidth}
	\centering
	\includegraphics[width=\textwidth]{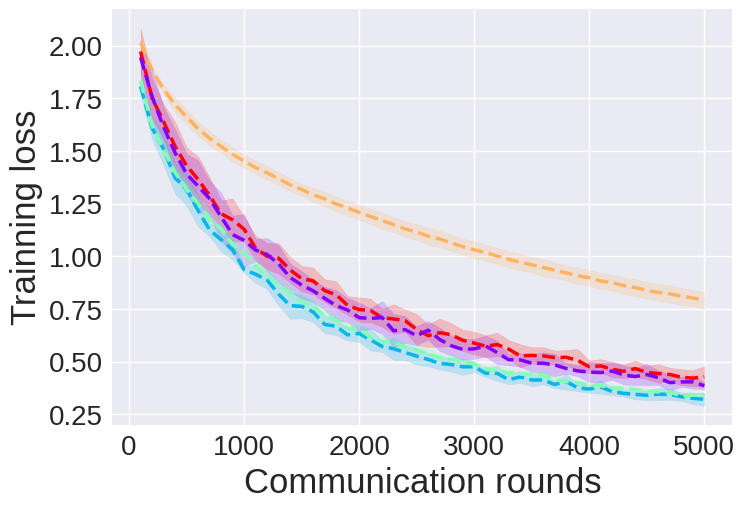}
	\caption{$E=5$}
	\label{}
\end{subfigure}
\begin{subfigure}[b]{0.32\textwidth}
	\centering
	\includegraphics[width=\textwidth]{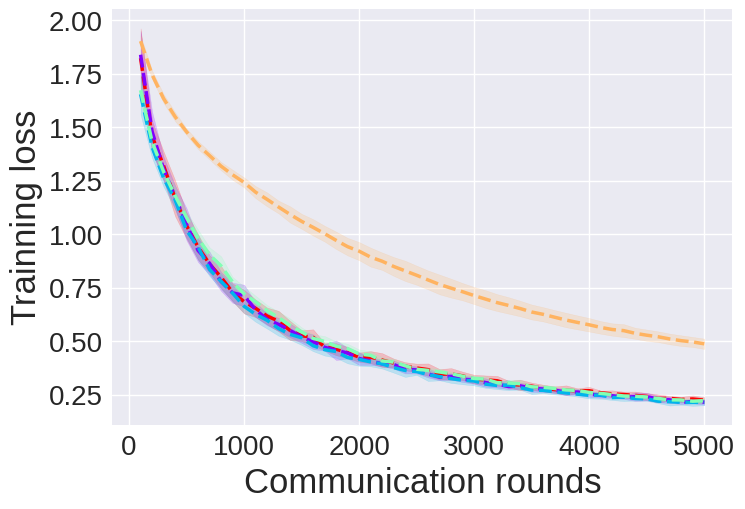}
	\caption{$E=10$}
	\label{}
\end{subfigure}
\begin{subfigure}[b]{0.32\textwidth}
	\centering
	\includegraphics[width=\textwidth]{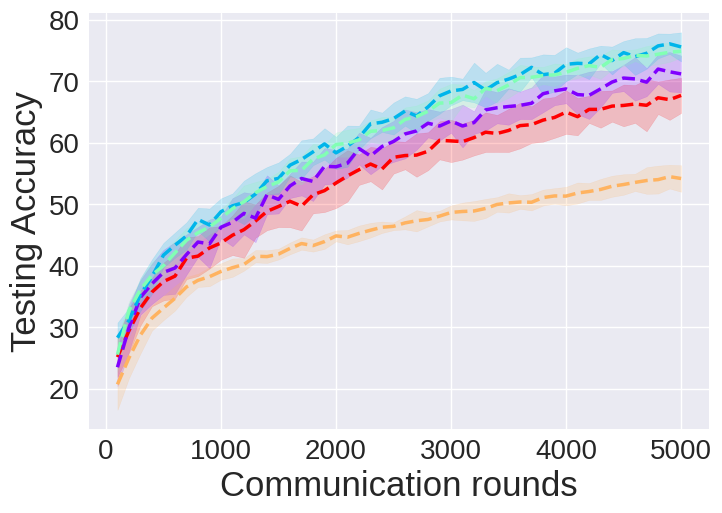}
	\caption{$E=2$}
	\label{}
\end{subfigure}
	\begin{subfigure}[b]{0.32\textwidth}
	\centering
	\includegraphics[width=\textwidth]{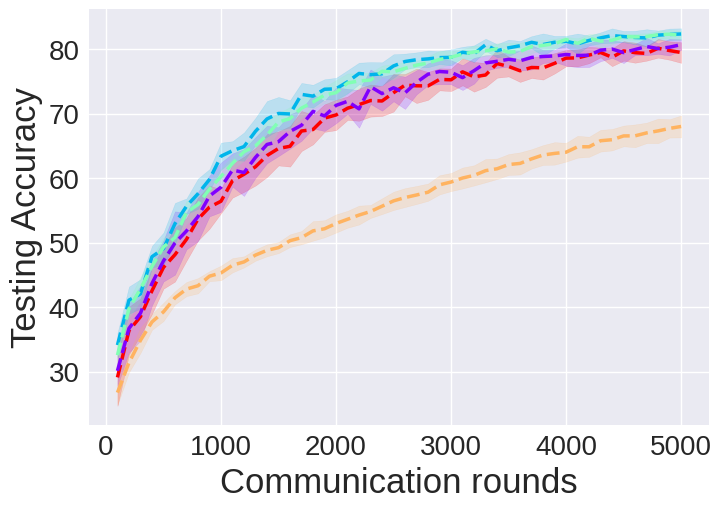}
	\caption{$E=5$}
	\label{}
\end{subfigure}
\begin{subfigure}[b]{0.32\textwidth}
	\centering
	\includegraphics[width=\textwidth]{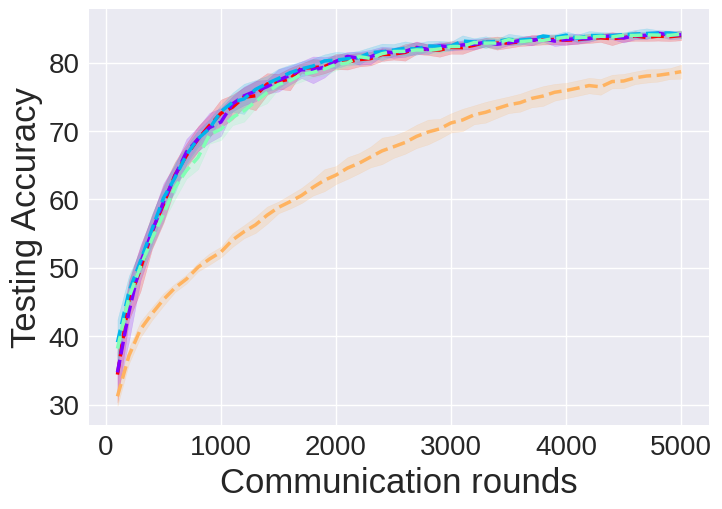}
	\caption{$E=10$}
	\label{}
\end{subfigure}
\caption{CIFAR-10: $1$-SignFedAvg under different noise scales and different numbers of local steps}
\label{fig:cifar_1sign}
\end{figure}

\begin{figure}[htbp]
\begin{minipage}[t]{1\linewidth}
\centering
\includegraphics[width= 4.5 in ]{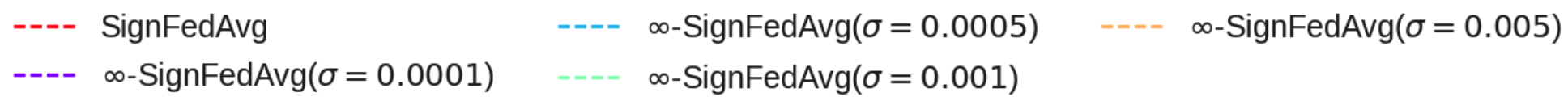}
\end{minipage}
	\centering
	\begin{subfigure}[b]{0.32\textwidth}
	\centering
	\includegraphics[width=\textwidth]{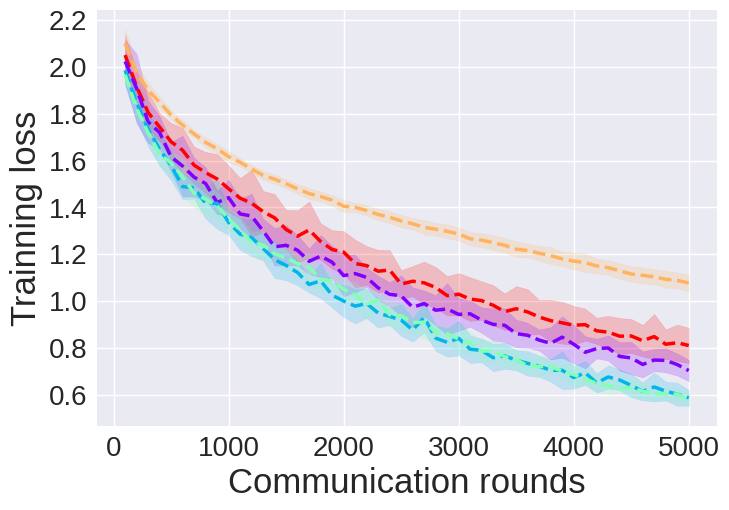}
	\caption{$E=2$}
	\label{}
\end{subfigure}
	\begin{subfigure}[b]{0.32\textwidth}
	\centering
	\includegraphics[width=\textwidth]{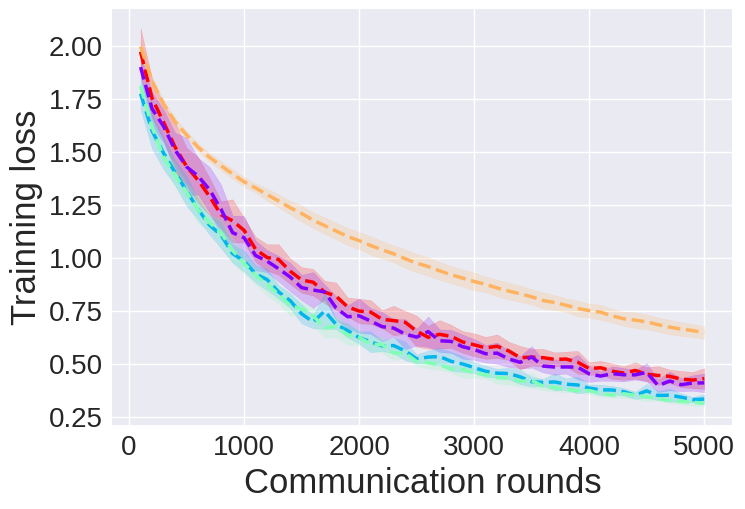}
	\caption{$E=5$}
	\label{}
\end{subfigure}
\begin{subfigure}[b]{0.32\textwidth}
	\centering
	\includegraphics[width=\textwidth]{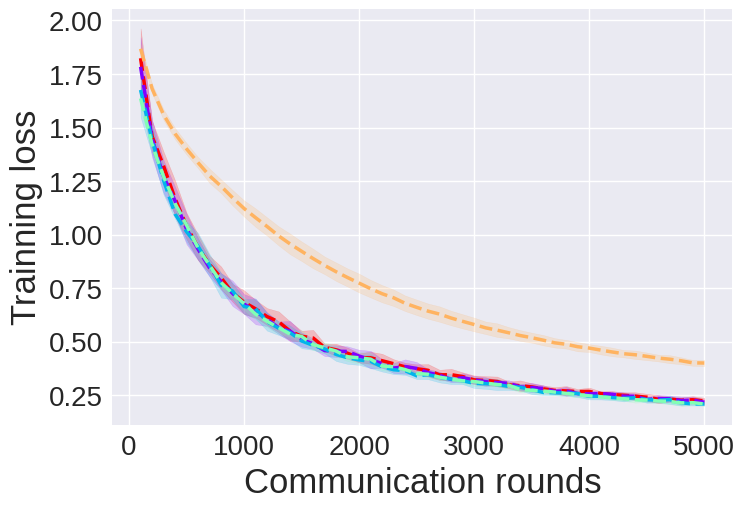}
	\caption{$E=10$}
	\label{}
\end{subfigure}
\begin{subfigure}[b]{0.32\textwidth}
	\centering
	\includegraphics[width=\textwidth]{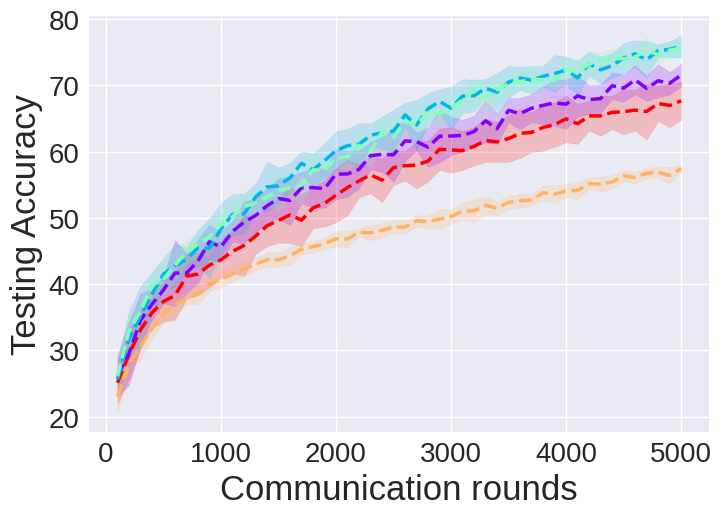}
	\caption{$E=2$}
	\label{}
\end{subfigure}
	\begin{subfigure}[b]{0.32\textwidth}
	\centering
	\includegraphics[width=\textwidth]{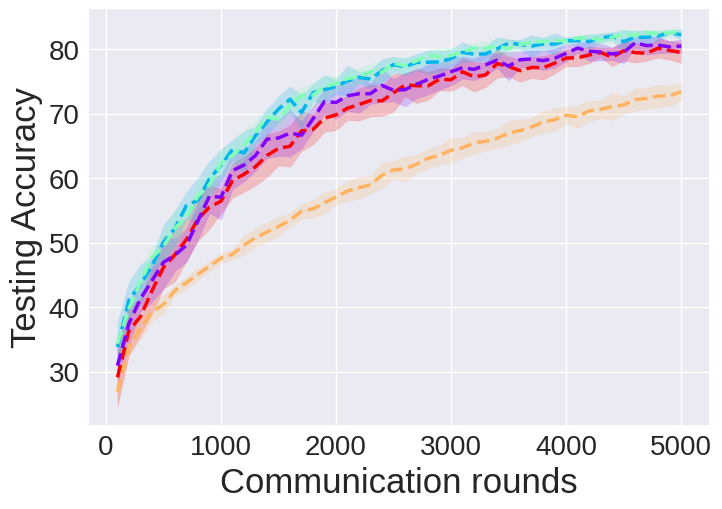}
	\caption{$E=5$}
	\label{}
\end{subfigure}
\begin{subfigure}[b]{0.32\textwidth}
	\centering
	\includegraphics[width=\textwidth]{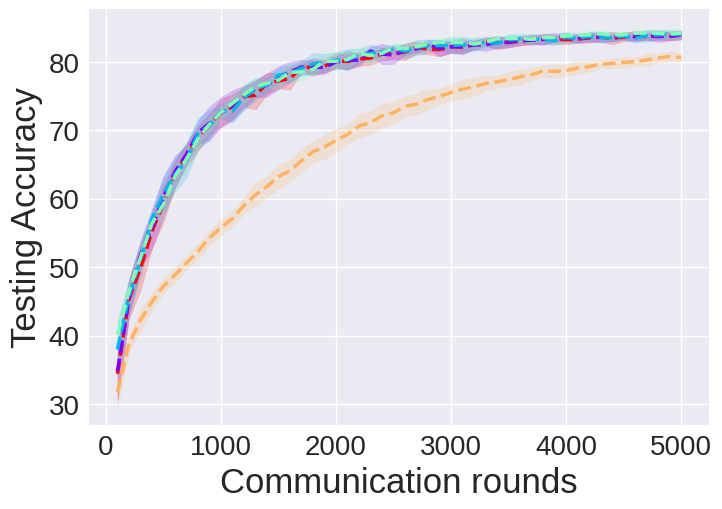}
	\caption{$E=10$}
	\label{}
\end{subfigure}
\caption{CIFAR-10: $\infty$-SignFedAvg under different noise scales and different numbers of local steps}
\label{fig:cifar_infsign}
\end{figure}

  \textbf{CIFAR-10:} For the experiment on CIFAR-10, we fixed the client stepsize as 0.1. We tuned the server stepsize, noise scales via grid search: $[10^0,10^{-0.5},10^{-1},10^{-1.5},10^{-2},10^{-2.5},10^{-3}]$ for the stepsize, $[0,0.0001,0.0005,0.001,0.005]$ for the noise scale. The comparison between $1$-SignFedAvg and $\infty$-SignFedAvg on CIFAR-10 is displayed in Figure \ref{fig:exp3-cifar_comp}. The used hyperparameter in the Figure \ref{fig:exp3-cifar} and \ref{fig:exp3-cifar_comp} are summarized in Table \ref{table:hp_cifar}. We also visualize the performance of $1$-SignFedAvg and $\infty$-SignFedAvg under various noise scales and different numbers of local steps in Figure \ref{fig:cifar_1sign} and Figure \ref{fig:cifar_infsign}. An interesting phenomeNon-in Figure \ref{fig:cifar_1sign} amd Figure \ref{fig:cifar_infsign} is that the more local steps are, the less impact the additive noise has on the convergence performance.
 
 \begin{table}[htpb]
	\centering
	\begin{tabular}{|c| c| c|} 
	 \hline
	 \makecell{\bfseries Algorithm} &{\bfseries Server stepsize}&   {\bfseries Noise scale} \\ 
	 \hline\hline
	 $1$-SignFedAvg & 0.03& 0.01\\ 
	 \hline
	  $\infty$-SignFedAvg & 0.03&0.01 \\ 
	 \hline
	 	 SignFedAvg & 0.03&0 \\ 
	 \hline
	\end{tabular}

	\caption{Hyperparameters for tested Algorithms on EMNIST.}
	\label{table:hp_EMNIST}
	\end{table}

% \begin{figure}[htbp]
% 	\centering
% 	\begin{subfigure}[b]{0.24\textwidth}
% 	\centering
% 	\includegraphics[width=\textwidth]{EMNIST_acc_noise_2.png}
% 	\caption{$E=2$}
% 	\label{}
% \end{subfigure}
% 	\begin{subfigure}[b]{0.24\textwidth}
% 	\centering
% 	\includegraphics[width=\textwidth]{EMNIST_acc_noise_5.png}
% 	\caption{$E=5$}
% 	\label{}
% \end{subfigure}
% \begin{subfigure}[b]{0.24\textwidth}
% 	\centering
% 	\includegraphics[width=\textwidth]{EMNIST_acc_noise_10.png}
% 	\caption{$E=10$}
% 	\label{}
% \end{subfigure}
% 	\begin{subfigure}[b]{0.24\textwidth}
% 	\centering
% 	\includegraphics[width=\textwidth]{EMNIST_acc_noise_20.png}
% 	\caption{$E=20$}
% 	\label{}
% \end{subfigure}
% \end{figure}

 \begin{table}[htpb]
	\centering
	\begin{tabular}{|c| c| c|} 
	 \hline
	 \makecell{\bfseries Algorithm} &{\bfseries Server stepsize}&   {\bfseries Noise scale} \\ 
	 \hline\hline
	 $1$-SignFedAvg & 0.0032& 0.0005\\ 
	 \hline
	  $\infty$-SignFedAvg & 0.0032&0.0005 \\ 
	 \hline
	 SignFedAvg & 0.0032&0 \\ 
	 \hline
	\end{tabular}
	\caption{Hyperparameters for tested Algorithms on CIFAR-10.}
	\label{table:hp_cifar}
	\end{table}

{

\subsubsection{Details for the experiment in Section \ref{sec:plateau}}
\label{app:plateau}

For the experiment results shown in Figure \ref{fig:exp-plateau}, except for the noise scale, both 1-SignSGD/1-SignFedAvg and  1-SignSGD-plateau/1-SignFedAvg-plateau used the same hyperparameters found in previous experiments. 
In Table \ref{table:hp_ada}, we show the hyperparameters 
of the Plateau criterion for the adaptive noise scale, which are chosen by a few rounds of trial and error. Besides, we also show the corresponding test accuracy in Figure \ref{fig:exp-plateau-test}, and how the noise scale evolves over communication rounds in Figure \ref{fig:exp-plateau-noise}.

 \begin{table}[htpb]
	\centering
	\begin{tabular}{|c| c| c|c|c|} 
	 \hline
	 \makecell{\bfseries Dataset} & $\sigma_{\text{init}}$&   $\sigma_{\text{bound}}$ & $\kappa$ &   $\beta$\\ 
	 \hline\hline
	 Non-i.i.d. MNIST & 0.01& 0.5& 30& 1.5\\ 
	 \hline
	 EMNIST & 0.0001&0.1& 10& 2 \\ 
	 \hline
	 CIFAR-10 & 0.001 &0.1& 200& 1.5 \\ 
	 \hline
	\end{tabular}
	\caption{{Hyperparameters of Plateau criterion for three different datasets.}}
	\label{table:hp_ada}
	\end{table}

\begin{figure}[htbp]
	\centering
\begin{subfigure}[b]{0.32\textwidth}
	\centering
	\includegraphics[width=\textwidth]{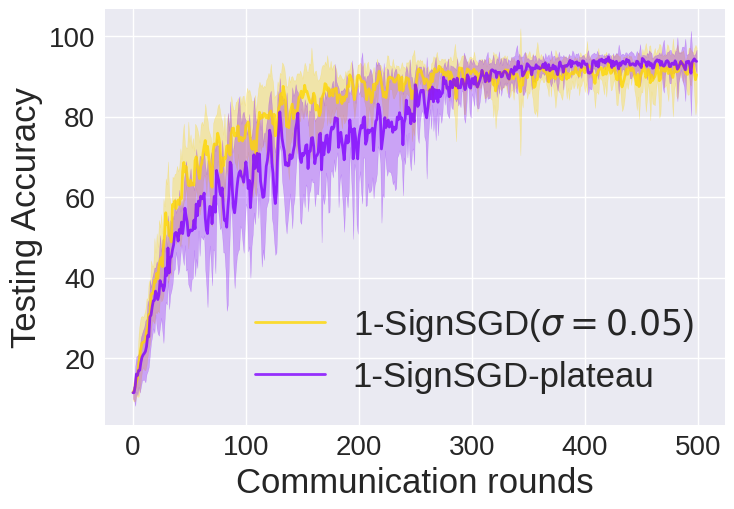}
	\caption{Non-i.i.d. MNIST}
	\label{}
\end{subfigure}
	\begin{subfigure}[b]{0.32\textwidth}
	\centering
	\includegraphics[width=\textwidth]{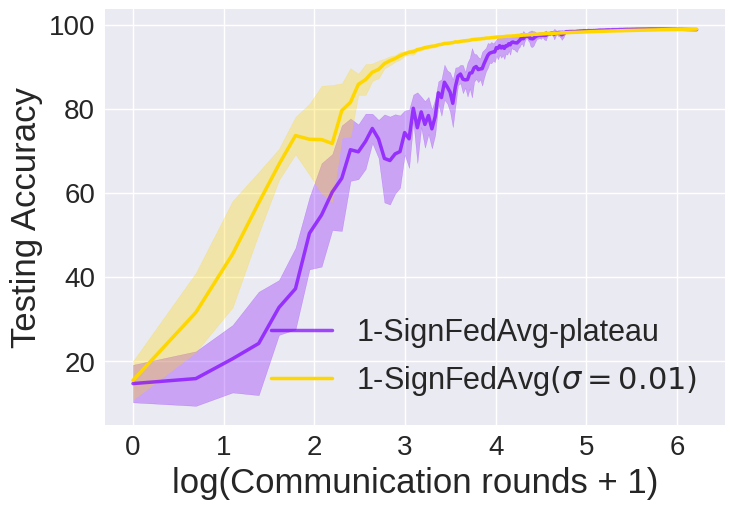}
	\caption{EMNIST}
	\label{}
\end{subfigure}
\begin{subfigure}[b]{0.32\textwidth}
	\centering
	\includegraphics[width=\textwidth]{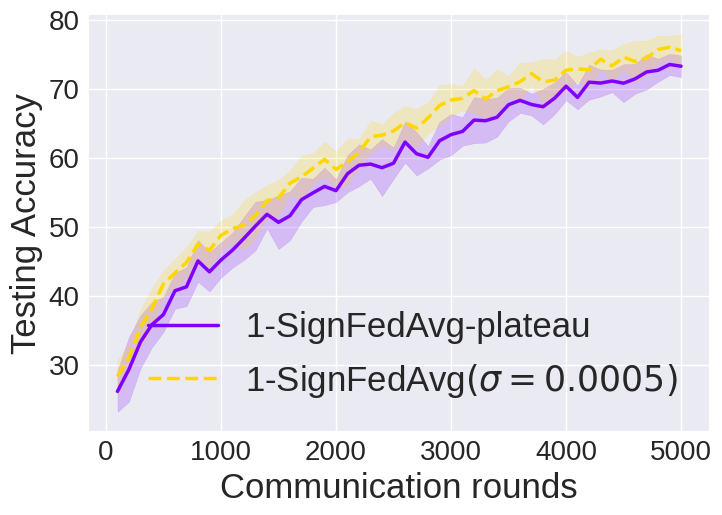}
	\caption{CIFAR-10}
	\label{}
\end{subfigure}
\caption{{The corresponding test accuracy to Figure \ref{fig:exp-plateau}. }}
\label{fig:exp-plateau-test}
\end{figure}

\begin{figure}[htbp]
	\centering
\begin{subfigure}[b]{0.32\textwidth}
	\centering
	\includegraphics[width=\textwidth]{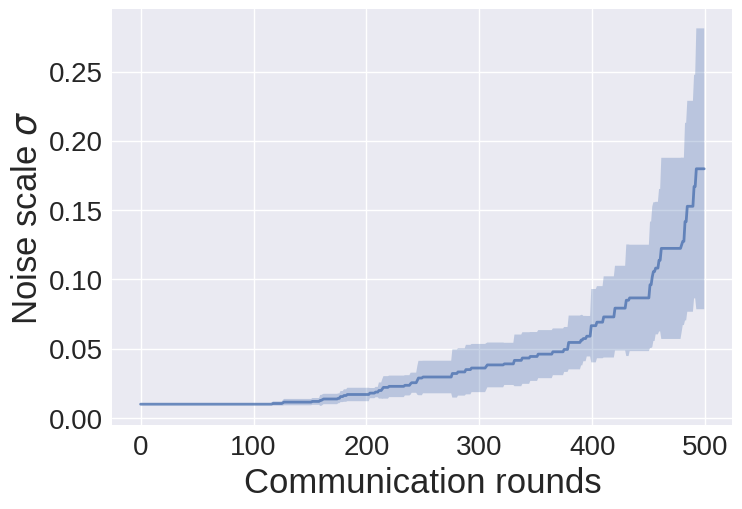}
	\caption{Non-i.i.d. MNIST}
	\label{}
\end{subfigure}
	\begin{subfigure}[b]{0.32\textwidth}
	\centering
	\includegraphics[width=\textwidth]{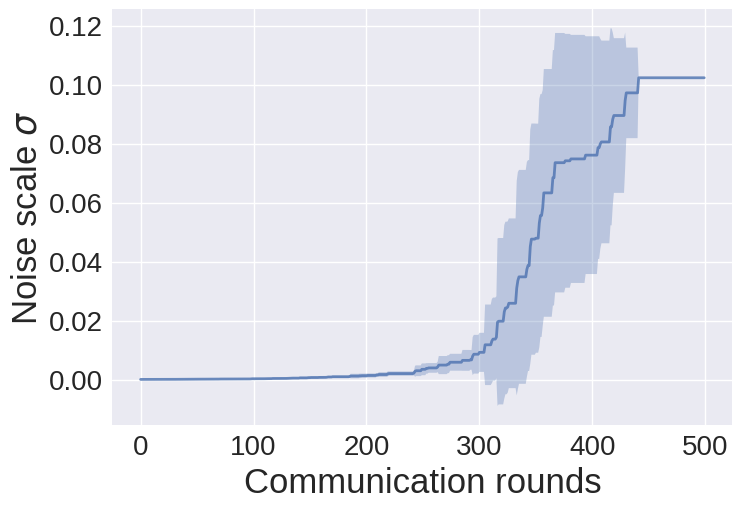}
	\caption{EMNIST}
	\label{}
\end{subfigure}
\begin{subfigure}[b]{0.32\textwidth}
	\centering
	\includegraphics[width=\textwidth]{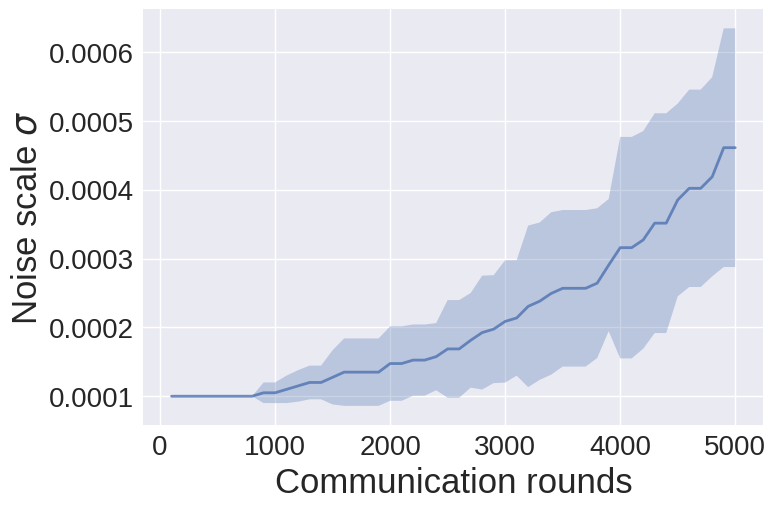}
	\caption{CIFAR-10}
	\label{}
\end{subfigure}
\caption{{The corresponding trends of noise scale to Figure \ref{fig:exp-plateau}.} }
\label{fig:exp-plateau-noise}
\end{figure}

\subsection{Comparison with unbiased stochastic quantization method}
\label{app:QSGD}

In this part, we compare our Algorithm \ref{alg:SignFedAvg} to the QSGD \cite{alistarh2017qsgd} along with its extension to FedAvg, i.e., FedPAQ \cite{reisizadeh2020fedpaq}. As we have shown that $z$-SignSGD/$z$-SignFedAvg with the Gaussian noise and uniform noise behave very closely, here we only consider $1$-SignSGD/$1$-SignFedAvg for comparison. We use the unbiased quantizer in \eqref{p:quantizer} for both QSGD and FedPAQ.

We can see that the quantization level $s$ plays as a key role in the performance and communication efficiency of QSGD and FedPAQ. In a rough sense, $s$ also represents the number of bits needed to transmit for a single coordinate. Thus, we will compare our algorithms to them with different choices of $s$. We remark that, even in the most extreme case, i.e., $s=1$, it still needs three alphabets $-1,1,0$ for communication, while sign-based method only uses $-1$ and $1$.

\textbf{Setting. }Again, we consider the three different datasets used in Section \ref{sec:exp2} and \ref{sec:exp2}. Specifically, we compare the $1$-SignSGD with QSGD on the non-i.i.d. MNIST dataset, and compare $1$-SignFedAvg with FedPAQ on EMNIST and CIFAR-10. For all the algorithms, the client's stepsize and batchsize are set to the same values used in Section \ref{sec:exp2} and \ref{sec:exp2}. For $1$-SignSGD/$1$-SignFedAvg, we reuse the previously found optimal hyperparameters. For QSGD, we tune the server stepsize via grid search on $[0.1,0.05,0.01,0.005]$. For FedPAQ, we tune the server stepsize via grid search on $[1,0.5,0.1,0.05,0.01,0.005]$. The chosen server stepsizes for QSGD and FedPAQ under three datesets are presented in Table \ref{table:hp_qsgd}.

\begin{table}[htpb]
	\centering
	\begin{tabular}{|c| c| c|c|} 
	 \hline
	 \makecell{\bfseries Algorithm} & \makecell{\bfseries Non-i.i.d. MNIST}&   \makecell{\bfseries EMNIST} & \makecell{\bfseries CIFAR-10}\\ 
	 \hline\hline
	 QSGD($s=1$) & 0.01& \diag{.1cm}{1.5cm}{}{} & \diag{.1cm}{1.5cm}{}{}\\ 
	 \hline
	 QSGD($s=2$)& 0.05&\diag{.1cm}{1.5cm}{}{}& \diag{.1cm}{1.5cm}{}{}\\ 
	 \hline
	 QSGD($s=4$)& 0.05&\diag{.1cm}{1.5cm}{}{}& \diag{.1cm}{1.5cm}{}{} \\ 
	 \hline
	 FedPAQ($s=1$) & \diag{.1cm}{2.7cm}{}{}& 1& 1\\ 
	 \hline
	 FedPAQ($s=2$)& \diag{.1cm}{2.7cm}{}{}&1& 1\\ 
	 \hline
	 FedPAQ($s=4$)& \diag{.1cm}{2.7cm}{}{}&1& 1 \\ 
	 \hline
	 FedPAQ($s=8$) & \diag{.1cm}{2.7cm}{}{}& 1& 1\\ 
	 \hline
	\end{tabular}
	\caption{{The chosen server stepsizes for tested QSGD and FedPAQ on three datasets.}}
	\label{table:hp_qsgd}
	\end{table}

\textbf{Results. } From Figure \ref{fig:exp-qsgd}, we can see that, our proposed sign-based compressor is consistently superior to the unbiased stochastic quantization method in low precision region (1 bit to 8 bits), except the only case that QSGD with $s=4$ is slightly better than our $1$-SignSGD on the non-i.i.d MNIST dataset. These results again, as \cite{bernstein2018signsgd,karimireddy2019error} did, show that the biased compressor, or more specifically the sign-based compressor, can be a strong competitor to those unbiased quantizer due to reduced variance. Our contribution in this work is to provide a generic framework that bridges the unbiased compressor and the biased one, which allows one to conveniently seek an optimal trade-off between the compression bias and variance.

\begin{figure}[b]
	\centering
	\begin{subfigure}[b]{0.32\textwidth}
	\centering
	\includegraphics[width=\textwidth]{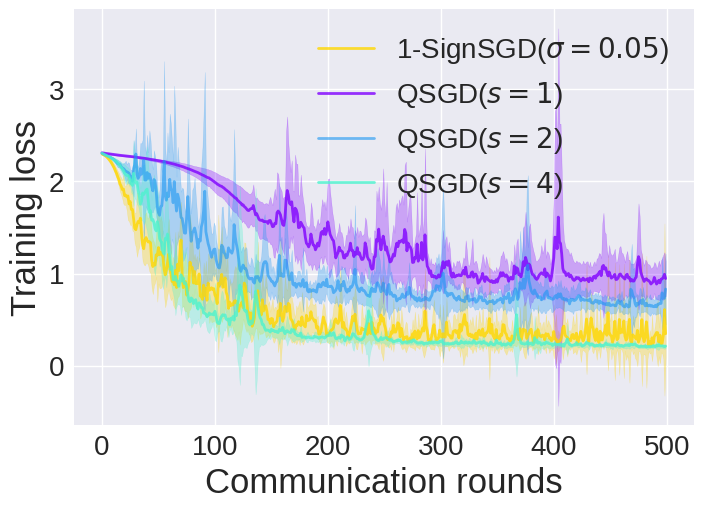}
	\caption{Non-i.i.d MNIST}
	\label{}
\end{subfigure}
	\begin{subfigure}[b]{0.32\textwidth}
	\centering
	\includegraphics[width=\textwidth]{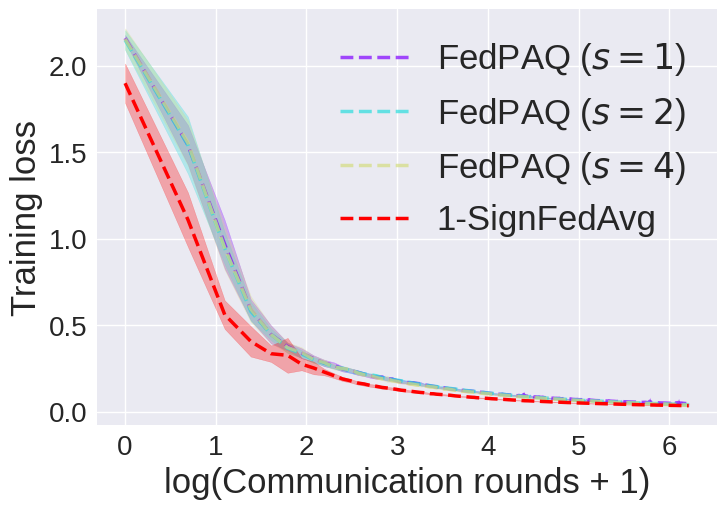}
	\caption{EMNIST}
	\label{}
\end{subfigure}
\begin{subfigure}[b]{0.32\textwidth}
	\centering
	\includegraphics[width=\textwidth]{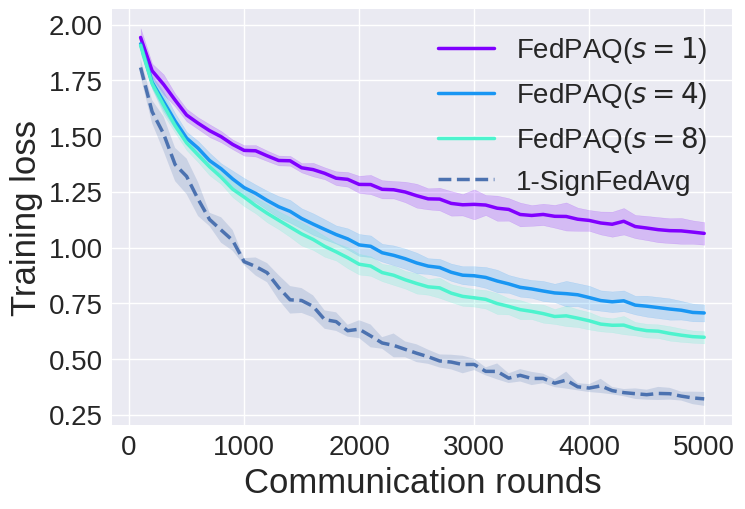}
	\caption{CIFAR-10}
	\label{}
\end{subfigure}
\begin{subfigure}[b]{0.32\textwidth}
	\centering
	\includegraphics[width=\textwidth]{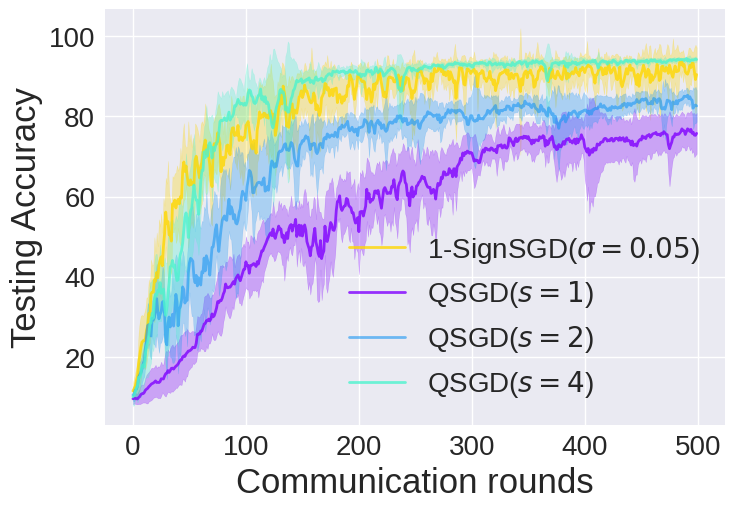}
	\caption{Non-i.i.d MNIST}
	\label{}
\end{subfigure}
	\begin{subfigure}[b]{0.32\textwidth}
	\centering
	\includegraphics[width=\textwidth]{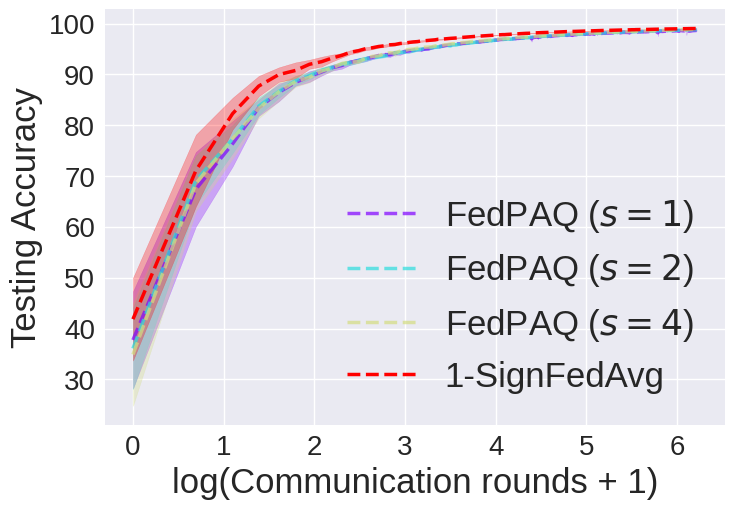}
	\caption{EMNIST}
	\label{}
\end{subfigure}
\begin{subfigure}[b]{0.32\textwidth}
	\centering
	\includegraphics[width=\textwidth]{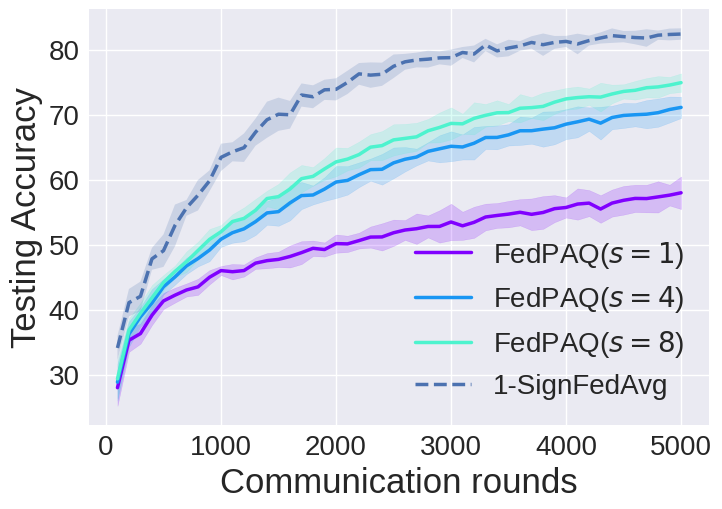}
	\caption{CIFAR-10}
	\label{}
\end{subfigure}
\caption{{Comparison of $1$-SignSGD/$1$-SignFedAvg with QSGD/FedPAQ on three datasets.}}
\label{fig:exp-qsgd}
\end{figure}

\subsection{Differential Private Federated Learning on EMNIST}
\label{app:DP}

Let us first review the definition of DP.
\begin{definition}[Approximate DP \cite{dwork2014algorithmic}]
\label{def:dp}
	A randomized algorithm $M$ that takes as input a dataset consisting of individuals is $(\varepsilon,\delta)$-differentially private if for any pair of datasets $S$,$S'$ that differ in the record of a single individual, and for any event $E$,
\begin{align}
	\label{p:DP}
	\mathbb{P}[M(S) \in E] \leqslant \mathrm{e}^{\varepsilon} \mathbb{P}\left[M\left(S^{\prime}\right) \in E\right]+\delta.
\end{align}
\end{definition}

The value $\varepsilon$ is regarded as the privacy budget, and the smaller it is the stronger privacy the algorithm provides. The quantity $\delta$ is usually set to $\frac{1}{n}$. The most popular mechanism to achieve DP is the Gaussian mechanism \cite{dwork2014algorithmic}. Specifically, similar to \cite{agarwal2021skellam,kairouz2021distributed}, here we consider client-level DP guarantee for Federated Learning, i.e, we regard each client as a single data point in Definition \ref{def:dp}. Besides, we also adopt the local version of DP gurantee, i.e., each dataset in Definition \ref{def:dp} contains only one data point. Such DP guarantee do not assume that the server is trustworthy and hence is commonly used in practice \cite{agarwal2021skellam,kairouz2021distributed}.  For more details on DP and its application in FL, we refer readers to  \cite{dwork2014algorithmic,mironov2017renyi,abadi2016deep,geyer2017differentially}. 

Here we describe the differential private version of Algorithm \ref{alg:SignFedAvg}, which we term DP-SignFedAvg (Algorithm \ref{alg:DP-SignFedAvg}).  The only difference between DP-SignFedAvg and $z$-SignFedAvg is that $z=1$ is chosen (Gaussian noise), and the norm of local gradients is clipped before perturbing it by the noise and applying the sign compression.  To obtain the client-level privacy guarantee, we adopt the privacy accounting method in \cite{mironov2019r}.

\label{app:DP-SignFedAvg}
\begin{algorithm}[H]
	\begin{algorithmic}[1]
	  \REQUIRE Total communication rounds $T$, Number of local steps $E$, Number of clients $n$, Client sampling ratio $q$, Clients stepsize $\gamma$, Server stepsize $\eta$, Noise coefficient $\sigma$, Norm clipping coefficient $C$.
	  \STATE Initialize $x_0$ and for $i=1,...,n.$  
	\FOR{$t=1$ to $T$}
    
    \STATE Sample a set of  clients $\Sc$ with size $qn$ for current round.
	\STATE \textbf{On Clients:}
	\FOR{$i$ in $\Sc$}
	\STATE $x_{t-1,0}^i=x_{t-1}$
	\FOR{$s=1$ to $E$}
	\STATE $g_{t-1,s}^{i}=g_i(x_{t-1,s-1}^i)$, where $g_i(\cdot)$ is the minibatch gradient oracle of the $i$-th client.
	\STATE $x_{t-1,s}^i =x_{t-1,s-1}^i - \gamma g_{t-1,s}^{i}$.
	\ENDFOR
	\STATE $\Delta_{t-1}^{i} = \text{Sign}\left(\frac{x_{t-1}-x_{t-1,E}^i}{\max\{1,\|x_{t-1}-x_{t-1,E}^i\|/C\}}+\Nc(0,\sigma^2C^2 I)\right)$.
	\STATE Send $\Delta_{t-1}^{i}$ to the server.
	\ENDFOR

	\STATE \textbf{On Server:}

      \STATE  $x_t = x_{t-1}- \eta \frac{1}{n}\sum_{i=1}^n  \Delta_{t-1}^{i}.$ 
      \STATE Broadcast $x_t$ to clients. 
	
	\ENDFOR
	\RETURN $x_T$. 
	\end{algorithmic}
	\caption{DP-SignFedAvg}
	\label{alg:DP-SignFedAvg}
	\end{algorithm}
	
Now we investigate the empirical performance of the DP-SignFedAvg on EMNIST, and compared it with the uncompressed DP-FedAvg used in \cite{agarwal2021skellam,kairouz2021distributed}.

% and compare the results to existing works \cite{agarwal2018cpsgd,agarwal2021skellam,kairouz2021distributed}.

{\bfseries Settings. }  
 We followed a setting similar to \cite{kairouz2021distributed} for the experiment on EMNIST. We adopted the client-level differential privacy, i.e., to treat each client as a single data point, and perturbed the local gradients before sending them to server. We also used the technique of privacy amplification by client sub-sampling in \cite{kairouz2021distributed,geyer2017differentially}.  For both DP-FedAvg and DP-SignFedAvg, the same CNN in Section \ref{sec:exp2} was used, and the maximum norm for clipping was set to 0.01. We sampled 100 clients at each communication round and ran both algorithms for 500 communication rounds.
Similar to \cite{kairouz2021distributed}, we run the experiments under the privacy budgets $\varepsilon=[1,2,4,6,8,10]$. In Table \ref{table:hp_EMNIST-DP}, we provide the hyperparameter for DP-FedAvg and DP-SignFedAvg for all levels of privacy budgets. Unlike previous experiments, the noise scales used in this experiment were determined by the privacy budget and the privacy accounting method in \cite{mironov2019r}.

\begin{table}[htpb]
	\centering
	\begin{tabular}{|c| c| c|c|} 
	 \hline
	 \makecell{\bfseries Privacy budget} &{\bfseries $\eta$ for DP-FedAvg}& {\bfseries $\eta$ for DP-SignFedAvg}  &  {\bfseries Noise scale}\\ 
	 \hline\hline
	 1.0029& 1 & 0.03 & 2.77\\ \hline
	 2.0171 & 2 & 0.05 & 1.57\\ \hline
	 4.0459 & 5 & 0.05 & 1.02\\ \hline
	 6.0135 & 5 & 0.05 & 0.845\\ \hline
	8.0336 & 5 & 0.05& 0.75 \\ \hline
	 9.9996 & 5 & 0.05 & 0.685\\ \hline
	\end{tabular}

	\caption{Hyperparameters for DP Algorithms on EMNIST.}
	\label{table:hp_EMNIST-DP}
	\end{table}

{\bfseries Results. }
%For the experiment on EMNIST, we set the total communication rounds to 500. 
It can be seen from Figure \ref{fig:exp4} that DP-SignFedAvg is only slightly inferior to the uncompressed DP-FedAvg for various levels of privacy budget. It is worthy to note that the work \cite{kairouz2021distributed} conducted a similar experiment and showed that the compressed DP-FedAvg with 12 bits for each gradient coordinate can be far worse than the uncompressed DP-FedAvg. It is a strong contrast to our DP-SignFedAvg which uses only 1 bit for each coordinate.

%A similar experiment in previous work \cite{kairouz2021distributed} shows that the compressed DP-FedAvg with 12 bits for each coordinate can be significantly worse than the uncompressed DP-FedAvg. As a comparision, our proposed DP-SignFedAvg only use 1 bit per-coordinate, and as  we can see from Figure \ref{fig:exp4},  DP-SignFedAvg is only slightly inferior to the uncompressed DP-FedAvg under multiple levels of privacy budget. 

\begin{figure}[htbp]
	\centering
	\begin{subfigure}[b]{0.45\textwidth}
	\centering
	\includegraphics[width=\textwidth]{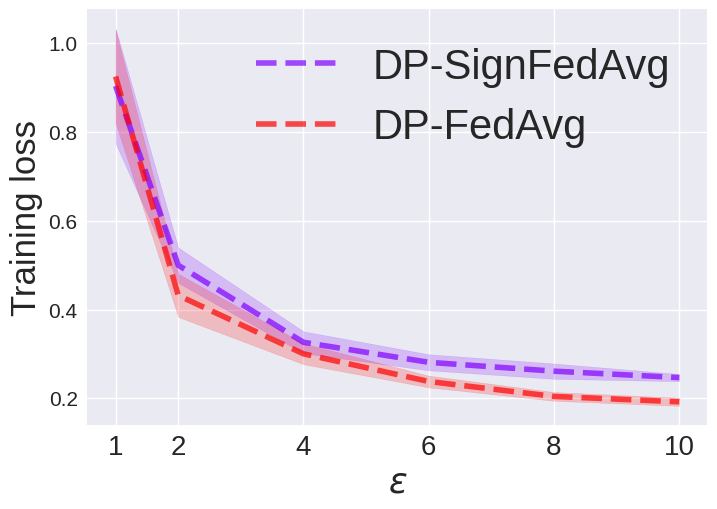}
	\caption{Training Loss}
	\label{}
\end{subfigure}
	\begin{subfigure}[b]{0.45\textwidth}
	\centering
	\includegraphics[width=\textwidth]{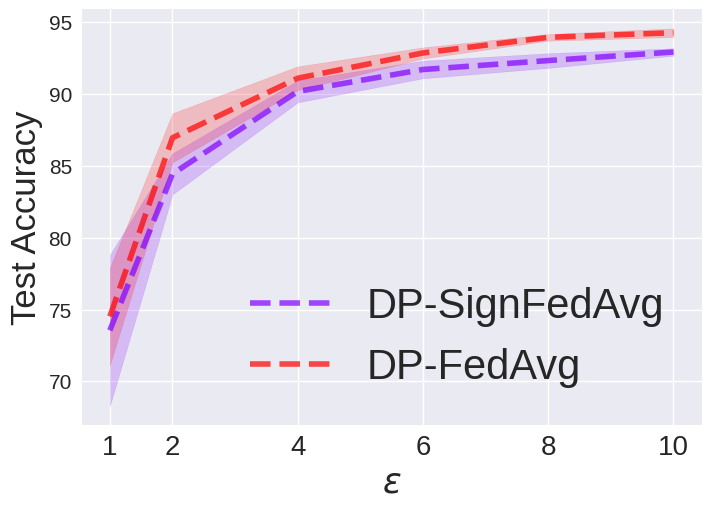}
	\caption{Test Accuracy}
	\label{}
\end{subfigure}
\caption{Performance of DP-SignFedAvg and DP-FedAvg}
\label{fig:exp4}
\end{figure}}

\end{document}